\RequirePackage{fix-cm}
\documentclass[twocolumn]{svjour3}          % twocolumn
\smartqed  % flush right qed marks, e.g. at end of proof
\usepackage[utf8]{inputenc}
\usepackage[T1]{fontenc}
\usepackage{graphicx}
\usepackage{hyperref}
\usepackage{caption}
\usepackage{subcaption}
\usepackage{booktabs}
\usepackage{tabularx}
\usepackage{multirow}
\usepackage{gensymb}
\usepackage{amsmath}
\usepackage{xcolor}
\usepackage{amssymb}

\newcommand{\hec}[1]{{#1}}
\newcommand{\rez}[1]{#1}

\newcommand\scalemath[2]{\scalebox{#1}{\mbox{\ensuremath{\displaystyle #2}}}}

\begin{document}

\title{HeRo 2.0: A Low-Cost Robot for Swarm Robotics Research%\thanks{Grants or other notes
% HeRo 2.0: An open-source, 18 USD, small robot for swarm robotics research
}
% \subtitle{Do you have a subtitle?\\ If so, write it here}


\author{\centering Paulo Rezeck \and 
        H\'ector Azp\'urua \and 
        Maur\'icio F. S. Corr\^ea\and 
        Luiz Chaimowicz
}


\institute{Paulo Rezeck and Hector Azp\'urua \at
            Graduate Program in Computer Science, \\
            Computer Vision and Robotics Laboratory,\\
            Department of Computer Science, \\
            Universidade Federal de Minas Gerais, Brazil. \\
            \email{\{rezeck, hector.azpurua\}@dcc.ufmg.br}
            \and
            Hector Azp\'urua is also with \at
            Instituto Tecnol\'ogico Vale,\\
            Ouro Preto, MG, Brazil.\\
            \email{hector.azpurua@itv.org}
            \and
            Maur\'icio F. S. Corr\^ea \at
            %Electrical Engineer, \\
            Computer Vision and Robotics Laboratory,\\
            Department of Computer Science, \\
            Universidade Federal de Minas Gerais, Brazil. \\
            \email{mauricio.ferrari@dcc.ufmg.br}
            \and
            Luiz Chaimowicz \at
            %Professor of Computer Science, \\
            Computer Vision and Robotics Laboratory,\\
            Department of Computer Science, \\
            Universidade Federal de Minas Gerais, Brazil. \\
            \email{chaimo@dcc.ufmg.br}           %  \\
        %   \and
    %
}

\date{Received: date / Accepted: date}

\maketitle

\begin{abstract}
\hec{The current state of electronic component miniaturization coupled with the increasing efficiency in hardware and software allow the development of smaller and compact robotic systems.} The convenience of using these small, simple, yet capable robots has gathered the research community's attention towards practical applications of swarm robotics. This paper presents the design of a novel platform for swarm robotics applications that is low cost, easy to assemble using off-the-shelf components, and deeply integrated with the most used robotic framework available today: ROS (Robot Operating System). The robotic platform is entirely open, composed of a 3D printed body and open-source software. We describe its architecture, present its main features, and evaluate its functionalities executing experiments using a couple of robots. Results demonstrate that the proposed mobile robot is very effective given its small size and reduced cost, being suitable for swarm robotics research and education.

 % mais algum???
%Mobile Robot, Low-Cost, Education.
\end{abstract}

\section{Introduction}
\label{sec:introduction}
%% Introduction %%

Robotic swarms are potentially becoming  well suited for a wide range of real-world problems with a high societal and economic impact. The requirement of distributed and decentralized processing relying only on local information brings several practical advantages over other robotic systems allowing scalability, resiliency, and adaptability. \hec{This aspect further leverages the use of swarm robots in agriculture, the mining industry, warehouse management, and robotics education.}

\hec{In spite of the increasing application potential of real-world robot swarms, several challenges are still open, ranging from efficient processing and communication to robust locomotion and sensing.} In addition, one of the main challenges in employing  swarm-based solutions \hec{in the real world} is the development of capable yet affordable robotic platforms.

Moreover, despite the existence of off-the-shelf solutions and some open software and hardware efforts,  the cost of the platforms and the logistics make it difficult for many researchers or educators to acquire or reproduce most of them. In order to alleviate such issues, the robotic platform presented in this work takes advantage of the recent technological advancements to use mass-produced components that are smaller, affordable, and long-term available. In addition, the design and assembly process follow new trends, such as the \textit{Maker Movement} and \textit{Do It Yourself}, which allow others to reproduce and customize the robot \hec{using additive manufacturing}.

% ***** LUIZ
In this sense, we assume the following requirements as an objective to build a swarm-capable robotic platform:

\begin{itemize}
  \item \textbf{Affordability}: Robots  should be as inexpensive as possible since most swarm teams may have tens or hundreds of robots;
  
  \item \textbf{Small and yet capable}: Robots should be small and equipped with some form of sensing capability to allow interaction with their environment; Also, they should have a long-term power autonomy since the swarm may need to operate long enough for the collective behavior to emerge;
  
  \item \textbf{Reliability}: Robots should be highly fault-tolerant;
  
  \item \textbf{Scalability}: They should be able to successfully perform different tasks even when the number of robots increases. In this sense, communication capabilities should support a large number of robots;
  
  \item \textbf{Easily reproducible}: Robots must be easily assembled and must not use hard-to-acquire \hec{or extremely hard-to-assemble} components;
  
  \item \textbf{Easily programmable}: Robots should be easily programmable and compatible with modern robotic frameworks \hec{and development pipelines}.
\end{itemize}

Satisfying these conditions in a single design is a difficult challenge. The design choices concerning one requirement, such as size, produce additional constraints to others, such as sensing and powering. \hec{Consequently, the design process should simultaneously take all of these constraints and find convenient design solutions for multi-purpose applications.}

\begin{figure}[t]
	\centering
	% \fbox{ }
	% trim={<left> <lower> <right> <upper>}
	\includegraphics[width=0.46\textwidth]{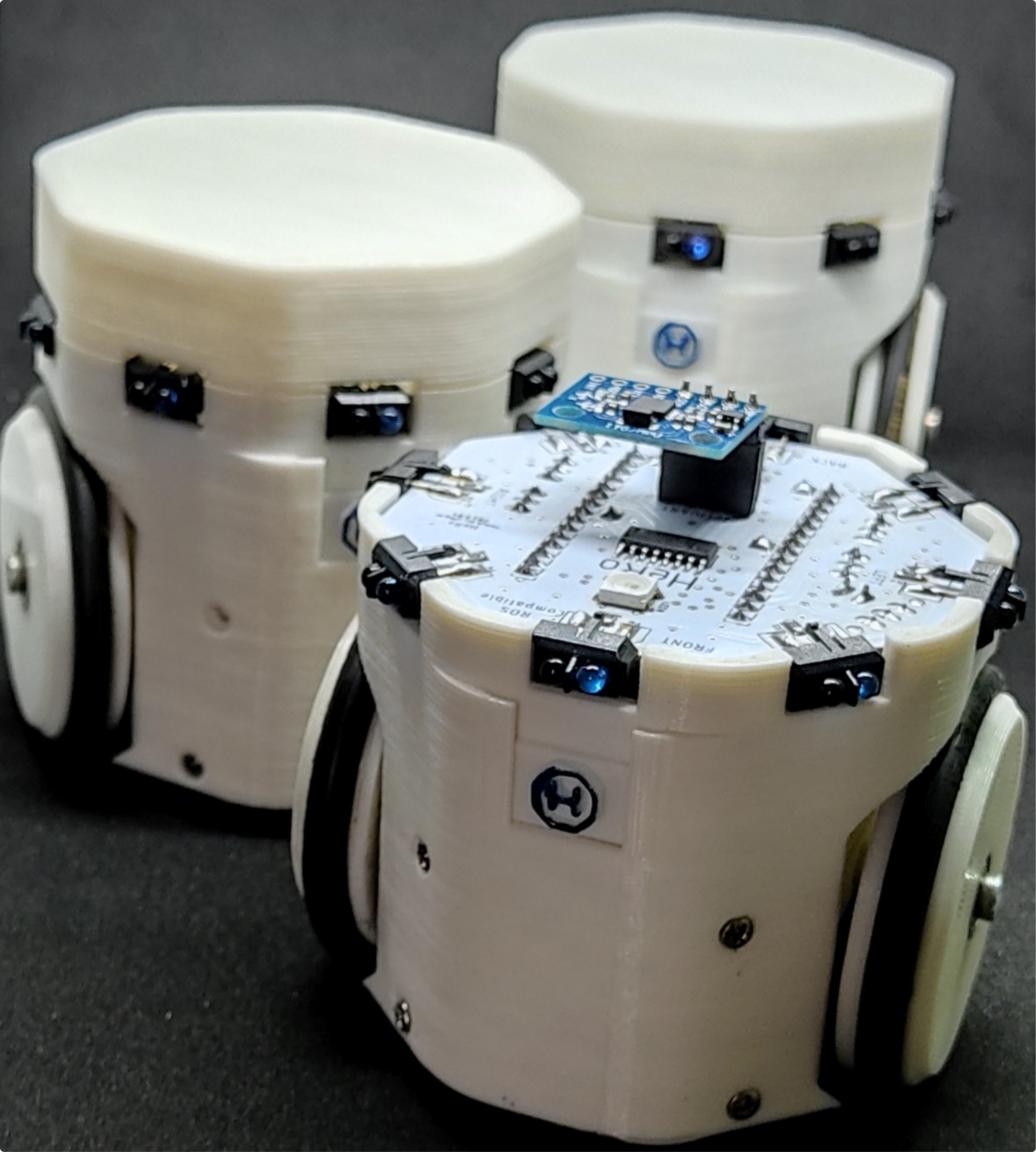}
	\caption{\hec{Design of the proposed open swarm robotic platform. The body of the robots was designed for and fabricated using additive manufacturing.}}
	\label{fig:hero_cover}
\end{figure}

Assuming the above requirements for a capable swarm robotic device, we present the design of HeRo: a significant low-cost robot (18 USD) composed of a 3D printed body and off-the-shelf components (Fig.~\ref{fig:hero_cover}). The robot is entirely open-source and carries a diverse set of sensors that makes it suitable for a wide range of swarm applications and education efforts. The platform is deeply integrated with the highly popular Robot Operating System (ROS) framework for quick prototyping, allowing remote and local robot control using standard programming interfaces. In order to facilitate the development of swarm algorithms, we also provide a simulated robot model with a realistic test environment in Gazebo. 
% Acho que não é necessária essa info aqui -- Luiz
%The proposed development ecosystem also allows updating the internal robot firmware using Wi-Fi over the air (FOTA) technology. 
This work is an evolution upon the first, simpler, version of the HeRo platform presented in~\cite{rezeck2017hero}.

%\todo{add that this is the updated version of the initial LARS paper}

% ***** LUIZ
\hec{The real-world sensorial and locomotion performance of the HeRo platform was evaluated in comparison to other popular commercial robot platforms. Some of the metrics for comparison are odometry accuracy, range, and quality of IR sensors (when used as range sensors), power consumption and autonomy, communication robustness, and scalability potential. Further, we also evaluated the HeRo performance in some cooperative tasks such as flocking, transportation, and mapping. Besides being relatively small in size, results show that the HeRo platform is significantly capable, making it cost-effective and suitable for swarm applications.}

\rez{The remainder of this paper is structured as follows. A review of the literature of small robot platforms and systems is presented in Section~\ref{sec:related_work}. The mechanical and electrical design, as well as the software and communication architecture, are presented in Section~\ref{sec:3_1_mecha_elect_desing} and Section~\ref{sec:3_2_soft_comm_arch}, respectively. The robot's performance with respect to a set of metrics is evaluated experimentally in Section~\ref{sec:experiments}. In Section~\ref{sec:casestudy}, we present the use of HeRo in some swarm applications. Finally, Section~\ref{sec:conclusion} brings the conclusions and directions for future work.}

\section{Related Work}
\label{sec:related_work}

\hec{Swarm robots have some elementary features that differentiate them from other types of platforms, such as simplicity, the capacity to scale and cooperate, size, and communication capabilities, among others~\cite{olaronke2020systematic}. Especially, some critical aspects for a swarm robot are footprint and cost, as those two aspects will facilitate the scalability of a real-world swarm system. Modern robotic systems also leverage the ecosystem and modularity of the Robot Operating System (ROS)~\cite{quigley2009ros} to improve the development environment and allow realistic simulations.}

\hec{In this sense, a wide range of small and relatively simple robots have been proposed for swarm applications. 
%A general trend of those robotics platforms relies upon the commercial nature of the project. 
Most of these platforms are open or have open-source parts, while only some of them are closed source or only available commercially. In this section, we present the most prevalent and relevant platforms for general swarm experimentation and highlight the most important pros and cons of each one.} We divide them according to their locomotion mechanisms and restrict this comparison to small robots (less than $10$~cm), which are generally more suitable for swarm robotics. Table~\ref{table:comparison} presents a summary of this comparison. 
%Due to many platforms present in the literature and trade-off issues between capability and size, we limited this comparison to small robots (less than $10$~cm). 
% Table~\ref{table:comparison} summarizes a comparison among them, and in the following, we present more details separating them by their motion mechanisms. 

{
\newcommand*\rot[1]{\hbox to1em{\hss\rotatebox[origin=br]{-50}{#1}}}
\begin{table*}[t]
	\centering
	\caption{Comparison of popular swarm robotics platforms.}
	\label{table:comparison}
	\resizebox{\textwidth}{!}{
	%\begin{tabular}{lcc lllllll}
	\begin{tabular}{lcc cccccccc}
% 		\toprule
		\textbf{Robot}  & 
		\rot{\textbf{Cost (USD)}}   & 
		\rot{\textbf{Size (cm)}}    & 
		\rot{\textbf{Motion/Speed (cm/s)}}  & 
		\rot{\textbf{Communication}} & 
		\rot{\textbf{Autonomy (h)}}     & 
		\rot{\textbf{Builtin Sensors}} & 
		\rot{\textbf{Open Source}} & 
		\rot{\textbf{ROS Enabled}} & 
		\rot{\textbf{FW Programming}} \\
		\midrule
		Kilobot~\cite{rubenstein2014kilobot} & $100/14^*$  & $3.3$ & vibration, $1$ & IR & $3-24$ & proximity, light & \checkmark & - & OTA \\
		Droplets~\cite{klingner2014stick} & $100$  & $4.4$ & vibration, $1$ & IR & $\infty$ & distance, light, bearing & \checkmark & - & wired \\
		
		\midrule
		Khepera I~\cite{mondada1994mobile} & N/A  & $5.5$ & wheel, $100$ & IR/RF & $1-$ & distance, light, encoder & - & - & wired \\
		Alice~\cite{caprari2003design} & N/A  & $2.2$ & wheel, $40$ & IR/RF & $1-10$ & distance, light & - & - & wired \\
		AMiR~\cite{arvin2009development} & $85^*$ & $7.5$ & wheel, $10$ & IR & $2$ & distance, light, bearing & - & - & wired \\
		E-puck~\cite{mondada2009puck} & $975$  & $7.0$ & wheel, $13$ & Wi-Fi/Bluetooth & $1-3$ & distance, light, camera, mic, imu & - & \checkmark & OTA \\
		Jasmine~\cite{kernbach2011swarmrobot} & $120^*$  & $3.0$ & wheel, $30$ & IR & $1-2$ & distance, light, bearing & \checkmark & - & wired \\
		GRITSBots~\cite{pickem2015gritsbot} & $50^*$  & $3.0$ & wheel, $25$ & RF & $1-10$ & distance, bearing, imu & \checkmark & - & OTA \\
		Zooids~\cite{le2016zooids} & $50^*$  & $2.6$ & wheel, $44$ & RF & $1-2$ & touch sensor & \checkmark & - & wired \\
		mROBerTO~\cite{kim2016mroberto} & $60^*$  & $1.6$ & wheel, $15$ & Bluetooth  & $1-6$ & distance, light, imu, camera & \checkmark & - & OTA \\
		WsBot~\cite{limeira2019wsbot} & $17^*$  & $3$ & wheel, $3.5$ & Wi-Fi  & $4$ & - & - & \checkmark & wired \\
		MicroMVP~\cite{yu2017portable} & $90^*$  & $8.0$ & wheel, $25$ & Zigbee & $1-2$ & - & \checkmark & \checkmark & wired \\
		Cellulo~\cite{ozgur2017cellulo} & $140^*$  & $7.5$ & wheel, $18$ & Bluetooth & $2$ & touch, visual odometry & - & - & OTA \\
		Colias IV~\cite{hu2018colias} & $100^*$  & $4.0$ & wheel, $35$ & Bluetooth & $1-3$ & distance, light, camera, mic, imu & \checkmark & - & wired \\
		Mona~\cite{arvin2018mona} & $120$ & $6.5$ & wheel, $15$ & RF & $\infty$ & distance, light, encoder & \checkmark & \checkmark & wired \\
		
		\midrule
		\textbf{HeRo} & $\mathbf{18^*}$ & $\mathbf{6.7}$ & \textbf{wheel,} $\mathbf{25}$ & \textbf{Wi-Fi} & $\mathbf{3-9}$ & \textbf{distance, light, encoder, imu} & \textbf{\checkmark} & \textbf{\checkmark} & \textbf{OTA} \\
		\bottomrule
		& *parts only \\
	\end{tabular}}
\end{table*}
}

\subsection{Vibration-based platforms}
Recently, robots using vibration-based motion mechanisms have become more common. In general, such mechanism can be easily coupled to the robot, but requires an extra effort in the robot's motion control algorithms. In addition, it requires a smooth experimentation surface and may have a relatively slow movement. 
Moreover, there is no real form of odometry, making it challenging to move precisely over long distances or perform for a long time if this information is necessary.

% Kilobot
The \textbf{Kilobot}~\cite{rubenstein2014kilobot}, developed at \textit{Harvard University} - USA, is one of the most popular swarm robots. It is an open-source platform
%, and it is the winner of the AFRON robot design challenge 
with parts costing only $14$~USD. But it is also produced and distributed as a commercial product for $100$~USD. The robot has an ATmega328 ($8$-bit at $16$~MHz) microcontroller and is equipped with \rez{an ambient light sensor on top} and an IR sensor on the bottom used for proximity readings and communication. The robot has an alternative moving principle based on two vibration motors, reducing cost and size, but it also limits the robot's maximum speed up to $1$~cm/s. An overhead controller \rez{device} is used to communicate via IR with all robots enabling remote control and uploading the robot's firmware over the air. Even with a relatively high commercial cost and limited sensing, research groups were able to successfully carry out experiments with up to $1000$~robots~\cite{slavkov2018morphogenesis}, showing that Kilobot may be an interesting platform for swarm applications. 

% Droplets 
The \textbf{Droplet}~\cite{klingner2014stick,farrow2014miniature} is another vibration-based small robot developed at the \textit{Correll Robotics Lab} at the \textit{University of Colorado Boulder}, USA. Despite being slightly larger than Kilobot, this robot features improvements in the mechanism of locomotion and sensing. The robots carry six IR sensors for proximity, bearing, and robot-to-robot communication. For locomotion, it uses three vibration motors to allow omnidirectional control of the robot, which is very convenient given its low speed ($1$~cm/s). An Xmega128a3u ($16$-bit at $32$~Mhz) microcontroller is also an improvement over Kilobot, allowing control, data processing, and general-purpose computation. In addition, the Droplet can also perform continuous experiment runs due to a powered floor equipped with alternating positive charge and ground stripes. Besides powering, this feature is also suitable for data transmission, enabling programming an entire swarm directly via the floor. The commercial cost of this robot is similar to the Kilobot ($100$~USD), but it still requires a powered floor mechanism to power the robots.

\subsection{Wheel-based platforms}
Although vibration-based locomotion does not require any complex mechanism to actuate the robot, such approach proves unsuitable for precise movements over long distances, mainly due to their nonlinear behavior and excessive slippage towards undesired directions. On the other hand, wheel based systems are more practical to control and efficient given that the torque generated by the motor acts directly and roughly linearly on the wheel. Below we list some wheel-based robotic platforms.

% Kephera I
\textbf{Khepera}~\cite{mondada1994mobile,mondada1999development} is one of the early small robots developed in the mid-1990s at the \textit{LAMI laboratory at \'Ecole Polytechnique F\'ed\'erale de Lausanne} (EPFL), Switzerland. The original version of Khepera is a small ($5.5$~cm) differential wheeled mobile robot that has been used by researchers of several universities for different applications. Two DC brushed servo motors with incremental encoders actuate and control the robot's wheels to reach up to $100~cm/s$. A Motorola 68331 ($32$-bits at $16$~MHz) microcontroller running $\mu$KOS RTOS serves as the robot's main processor, enabling motion control, sensing, and communication. In addition to eight IR sensors used to estimate distance and ambient light, the robot also allows extra modules that expand its functionality. Some examples are gripper-like manipulation, vision, and robot-to-robot communication modules. Over the years, several versions of Khepera have been developed, improving unit processing, locomotion, and sensing but requiring an increase in size. The latest version, the Khepera IV~\cite{soares2016khepera}, is still a differential wheeled mobile robot with a diameter of $14$~cm. This robot houses twelve IR sensors, five ultrasound sensors, two microphones, encoders, an inertial measurement unit (IMU), and a camera. The main processing unit is a Gumstix embedded computer running GNU/Linux, and Bluetooth allows robot-robot communication or communication with a remote server. Its commercial version retails for $3180$~USD.

% Alice
\textbf{Alice}~\cite{caprari2003design} is another small robot developed for swarm applications at the \textit{Autonomous Systems Lab at \'Ecole Polytechnique F\'ed\'erale de Lausanne} (EPFL), Switzerland. Alice is a two-wheeled differential drive robot made of a light plastic chassis with PCB on top. The robot has a small footprint of $2.2$~cm and uses two high-efficiency swatch motors for locomotion reaching up to $40$~cm/s. A low-power PIC16F877 ($8$-bits at $4$~MHz) microcontroller controls the robot and executes other applications. Alice has various built-in sensory modules such as $4$~IR sensors mounted around the robot for obstacle detection and short-range robot-to-robot communication. An IR receiving on top allows the robot to receive external commands, and a radio frequency (RF) module is used for remote communication. In addition, the robot supports different expansion modules, such as a gripper module and a linear camera. The first design of Alice used two watch batteries allowing the robot to operate for up to $10$ hours. Further evolution of the platform allows the use of solar panels. 

% AMiR
\textbf{AMiR}~\cite{arvin2009development} is a two-wheeled differential drive small robot developed at the \textit{University Putra}, Malaysia. It is an open platform, and the components required to assemble it cost about $85$~USD. The robot's footprint is $7.5$~cm, and two micro DC internal gear motors actuated the robot with a maximum speed of $10$~cm/s. An ATmega168 ($8$-bits $8$~MHz) microcontroller is used as the main processor to control all functions such as communication, trajectory, and perception, among others. The robot carries $6$~IR sensors enabling proximity and bearing estimation and also short-range robot-to-robot communication. The robot uses a $3.7$~V $200$~mAh lithium battery allowing it to operate up to $2$~hours. In addition to the physical platform, AMiR has been successfully simulated in Player/Stage and has been used by several researchers and robotics educators~\cite{arvin2011imitation,arvin2014comparison}.

% E-puck
The \textbf{E-puck}~\cite{mondada2009puck} is one of the most successful small-size commercial robots. Initially designed for education it has also been used for swarm robotics research. The E-puck is a two-wheeled differential drive robot, and its retail cost is about $975$~USD. The robot has a small footprint of $7.0$~cm and uses two planetary-geared step motors for actuation, reaching up to $10$~cm/s. The latest version of the e-puck is powered by an STM32F4 ($32$-bits at $180$~MHz) microcontroller, and an Espressif ESP32 is used as Wi-Fi/Bluetooth module. The robot hosts various built-in sensors, including microphone arrays, proximity sensors, a $640\times480$ pixels camera, and an inertial motion unit. The robot can be programmed through a serial BUS or Bluetooth interface, and Wi-Fi is used for communication. In addition, it can be extended with other sensing modules, such as bearing and an omnidirectional camera module, and even processing modules using Raspberry Pi. The robot uses a $3.7$~V $1200$~mAh lithium battery allowing it to operate up to $3$~hours. A growing user community provides software, documentation, and discussion groups favoring the platform's integration with various simulators and robotics frameworks, such as Gazebo and ROS. Despite its benefits, the commercial version of the basic E-puck is quite expensive, making it not affordable for large swarms.

% Jasmine
\textbf{Jasmine}~\cite{kernbach2011swarmrobot} is another widely used two-wheeled differential drive small robot. Developed at the \textit{University of Stuttgart}, Germany, its parts cost about $120$~USD. The Jasmine robot has a small footprint of $3$~cm and uses two micro DC internal gear motors to actuate the wheels, reaching a maximum speed of $30$~cm/s. The third version of the robot is equipped with an ATmega168 ($8$-bits at $20$~MHz) microcontroller, and uses $6$~IR sensors for proximity and bearing estimation, light measurements, and communication with other robots. The robot also has LEDs on top, allowing status monitoring or debugging. In addition, many customized boards can extend the robot's capabilities, including improved sensing and connectivity. The current version of Jasmine uses a $3.7$~V $250$~mAh lithium battery that has enough capacity for running time up $2$ hours. Moreover, the robots can autonomously recharge the battery by touching a pair of metal contacts (power and ground) attached to the wall for convenience. Thus, the robot detects when its battery needs to be recharged and moves autonomously to the dock without human intervention.

% GRITSBot
\textbf{GRITSBot}~\cite{pickem2015gritsbot} is a small robot developed at \textit{Georgia Institute of Technology}, USA. GRITSBot is part of \textit{Robotarium}, a project to make multi-agent experiments more accessible to the research community, opening up a showcase testbed to the general public~\cite{pickem2017robotarium,wilson2020robotarium}. The GRITSBot is another wheeled differential-drive robot composed of three modular layers that house five functional robot blocks. The motor layer is responsible for controlling the two stepper motors and odometry estimation. The mainboard houses an Atmega328 ($8$-bit at $16$~MHz) microcontroller, the wireless communication module, the battery charging circuit, and the power supply. A Nordic nRF24L01 microchip serves as a low-power consumption communication module operating at $2.4$~GHz. This module enables robot-to-robot communication, over-the-air firmware reprogramming, and remote control from a server. The sensor layer includes six infrared distance sensors, an accelerometer, and a gyroscope. A 400 mAh LiPo battery supplies the robot allowing long-time power autonomy up to five hours. The robots can also move autonomously to a power source and automatically recharge the battery conveniently. %These low-power transceivers were chosen over Wi-Fi due to their reduced power consumption. However, the drawback of these low-power transceivers is their lower data rate, which is limited to $2$~Mbit/s. Also, these modules operate at the same Wi-Fi frequency, making them susceptible to interferences when used in indoor environments.

% Zooids
\textbf{Zooid}~\cite{le2016zooids} is a small robot platform designed for swarm applications available at an approximate cost of $50$~USD. This robot is an open-source open-hardware platform created as a joint work between the Shape Lab at \textit{Stanford University} (USA) and the Aviz team at \textit{Inria} (France). 
The motors are mounted in a non-collinear fashion, allowing a small footprint of only $2.6$~cm.
Even though the motors do not rotate around the same axis, the robot has a similar net force and moment as a robot with colinear motors. An STM32F4 ($32$-bit at $48$~MHz) microcontroller manages the overall logic computation and communicates wirelessly with the main master computer using an nrf24L01 $2.4$~GHz radio chip. In addition, the robot is equipped with touch sensors for tactile swarm applications and some on-top photodiodes used for localization. A projector-based tracking system is used for robot position tracking.  This device projects a sequence of gray-coded patterns onto a flat surface, enabling the robots to use their photodiodes to decode the gray code into position and orientation. Unlike classical camera-based systems, this projector-based tracking system does not add any latency from networking for the local feedback control on each robot, making position control more stable. However, this localization system costs approximately $700$~USD and archives similar resolution compared to overhead-camera localization systems.

% mROBerTO
The \textbf{mROBerTO}~\cite{kim2016mroberto} is a small footprint ($1.6$~cm) robot developed at\textit{University of Toronto}, Canada. Despite the small footprint, the robot features several built-in sensors such as distance, ambient light, IMU, and camera, making it interesting for swarm applications. In addition, the robot supports extensions such as a module with 8 IR sensors for obstacle detection. The robot's mainboard is a Nordic nRF51422 microchip composed of an ARM Cortex-M0 ($32$-bits at $16$~MHz) with built-in Bluetooth Smart and ANT+ capability. The nRF51422 board supports over-the-air programming, saving time when setting up several of robots simultaneously. Regarding the robot actuation mechanism, the first version of mROBerTO did not require wheels and used the motor shafts directly in contact with the floor surface to move the robot. Despite being a compact actuation mechanism, allowing the robot to reach speeds of up to $15$~cm/s, it requires a smooth contact surface for proper robot control. In more recent versions, the authors improved the actuation mechanism to utilize small stepper motors with wheels~\cite{eshaghi2020mroberto}.

% WsBot
The \textbf{WsBot}~\cite{limeira2019wsbot} is another small footprint ($3.3$~cm) robot developed at \textit{Universidade Tecnol\'ogica Federal do Paran\'a}, Brazil. This robot has a design similar to mROBerTO but at an assembly cost of only $17$~USD. In addition to being extremely inexpensive, one may easily assembly this robot using only off-the-shelf parts. The WsBot is envisioned for demonstrations of applications in Industry 4.0, so the robot has a built-in wireless charging system allowing automatic battery recharges for continuous operation. Two micro DC motors with small wheels drive the robot allowing it to reach a speed of up to $3.5$~cm/s. An Espressif ESP8266 ($32$-bits $160$~MHz) microchip, with built-in Wi-Fi, enables remote control of the robot using a server executing ROS. The robot does not feature any built-in sensor or extension boards. Instead, a global localization system based on an overhead camera and fiducial markers is used for robot close-loop control. 

% MicroMVP
\textbf{MicroMVP}~\cite{yu2017portable} is a small robot developed at MIT, USA. It has an open-source design utilizing 3D printing technology, and it is also extremely simple and easy to assemble. MicroMVP was designed to use only off-the-shelf components, and it is composed of an ATmega32U4 ($8$-bit at $16$~MHz) microcontroller with built-in xBee support and two geared motors. As WsBot, MicroMVP does not provide built-in sensors reducing its applicability as a swarm-capable robot. It also uses an overhead camera and fiducial markers attached on top of the robot to localize them, serving as loop-close control. However, MicroMVP uses more expensive components reaching an assembling cost of $90$~USD.

% Cellulo
\textbf{Cellulo}~\cite{ozgur2017cellulo} is one of the world's first tactile small robot platforms developed at \textit{\'Ecole Polytechnique Federale de Lausanne} (EPFL), Switzerland. It combines autonomous capabilities with haptic-enabled multi-user tactile interaction allowing research on rehabilitation, gaming, and human-computer interaction. The robots are designed to be small, sturdy, low-cost, and simple to operate. The current Cellulo robot is equipped with a self-localization system based on an activity sheet and a downward-facing camera, holonomic motion, six capacitive touch buttons, Bluetooth communication, and a low-cost PIC32MZ ($32$-bit at $200$~MHz) microcontroller. The localization system~\cite{hostettler2016real} enables the user to estimate the global pose of many robots and also is robust against kidnapping and occlusions (usually due to user manipulation). %In addition, the robot moves holonomic using a permanent-magnet assisted omnidirectional ball drive~\cite{ozgur2016permanent}. Holonomic motion is a significant improvement over differential drive motion (commonly found in a swarm-like robot), as it allows the robot to start moving in any direction. Despite these improvements, the localization system is not straightforward to deploy and store away and also is not robust enough against external contaminants such as dust and wheel marks on the ground. In addition, the rubber shards from the ball wheels quickly accumulate in the bearing and may decrease the performance drastically.

% Colias
\textbf{Colias} is a novel alternative to AMiR developed at the \textit{University of Lincoln}, UK for swarm robotic applications. Colias sensor unit is based on extension boards to achieve better modularity. In this way, each part has different features and functions that can work independently. The mainboard uses an ATmega168 ($8$-bit at $8$~MHz) microcontroller to control the motors and power management. This board houses IR sensors that provide proximity measurements used for obstacle detection. The motion is produced by two differential-driven wheels reaching a maximum speed of $35$~cm/s. The new Colias IV~\cite{hu2018colias} is additionally powered with a high-level ARM Cortex M4 microcontroller running at $180$~MHz, two digital microphones, one $9$-axis motion sensor, and a tiny VGA camera to enable visual tasks. A Bluetooth extension module enables Colias IV to communicate with a remote host device such as a laptop or a smartphone, receiving motion commands or sending sensor data. %Besides the hardware, Colias comes with a set of basic software libraries for sensor reading and motion control. However, programming still requires a direct connection with the user computer making experimentation hardly scalable.

% Mona
\textbf{Mona}~\cite{arvin2018mona} is a small open-source robot built as a customized design of Colias. It has also been designed as a modular platform, allowing additional modules, such as wireless communication or a vision board. Mona is mainly designed to investigate the feasibility of the proposed Perpetual Robotic Swarm~\cite{arvin2018perpetual}. The robot is specially designed to use an inductive charging approach and several additional functions such as a radio frequency (RF) transceiver and battery level monitoring module. This perpetual charging interface allows for large-scale, long-term autonomy robotics research. Mona has also been developed to be compatible with several standard programming environments, and it has been successfully used for both education and research at the \textit{University of Manchester}, UK. The robot has been produced as a low-cost platform for robotic education and swarm research in collaboration with a commercial partner. It retails at $120$~USD per robot, and it remains fully open-source (hardware and software).

\subsection{Proposed platform: {\bf HeRo}}
% HeRo
%Figure~\ref{fig:heroreal} shows the current version of \textbf{HeRo}. 
In this paper, we present the project and implementation of a novel small robot for swarm robotic applications. \hec{The current proposal is an evolution upon the first, simpler version of the HeRo platform which was briefly presented in~\cite{rezeck2017hero}. A summary of the characteristics of all HeRo versions is described in Table~\ref{table:hero_versions}.} In this improved version, the mainboard uses an Espressif ESP8266 ($32$-bits $160$~MHz) microcontroller to perform the motors' control and acquire and process sensor data. This microcontroller houses a built-in Wi-Fi module, allowing the robots to communicate among themselves robustly and reliably, using TCP/IP protocols. The locomotion system consists of using two differential-driven wheels reaching a maximum speed of $25$~cm/s. The board houses a set of sensors such as 8 IR sensors that provide light and proximity measurements for obstacle detection, an inertial motion unit for improved odometry and general use, and two rotary encoders for localization and motion control. The mainboard is also modular, allowing the user to attach several other components such as a camera, motors, displays, and transistors for communication or localization. To facilitate programming, HeRo supports FOTA (Firmware Over-The-Air) using a Wi-Fi interface. Such technology allows the users to upload their codes on many robots remotely. Moreover, HeRo is also a ROS-compatible robot and communicates using a TCP/IP connection with a remote computer executing ROS. Since the robot's autonomy is an important factor considering the time and number of experiments, HeRo provides a long-time autonomy using a powerful Li-Po battery. The main contributions of the proposed platform compared to its counterparts are its balance between low cost and capacity, simplicity in terms of assembly, and seamless integration with ROS allowing easy programming. 

\begin{table*}[!th]
    \centering
    \caption{Characteristics of the different HeRo versions.}
    \label{table:hero_versions}
    \resizebox{\textwidth}{!}{
        \begin{tabular}{llll}
            \toprule
            & HeRo v0.1 & HeRo v1.0 & \textbf{HeRo v2.0} \\
            \midrule
            Board & Arduino Nano & ESPressif ESP8266 - ESP12 & ESPressif ESP8266 - ESP12 \\
            MCU & Atmel Atmega328 8-bit @ 16 MHz & Tensilica LX106 32-bit @ 80/160 MHz & Tensilica LX106 32-bit @ 80/160 MHz \\
            Communication & RF nrf24l01 $2.4$~Ghz & Wi-Fi 802.11bgn & Wi-Fi 802.11bgn \\
            Actuation & Servo Motors & Servo Motors & Servo Motors \\
            Footprint & $10$~cm & $8$~cm & $6.7$~cm \\
            Sensors & None & 3 x infrared sensors & 8 x infrared sensors, encoders and IMU \\
            Battery & $3.7$~V $1000$~mAh Li-Po & $3.7$~V $1000$~mAh Li-Po & $3.7$~V $1800$~mAh Li-Po \\
            Cost* & $9$~USD  & $14$~USD & $18$~USD \\
            \bottomrule
            * parts only &
        \end{tabular}
    }
    % }\footnote{HeRo v2.0: \url{HeRo v2.0}
\end{table*}

\section{Mechanical and Electrical Design}
\label{sec:3_1_mecha_elect_desing}
% HeRo Desing

\rez{This section presents the mechanical and electrical design of our swarm robot. All decisions considered the maximum use of commercially available components for ease of production and assembly and minimum possible price without sacrificing the processing power and sensing capabilities.} \hec{Therefore, in the following, we present the best-suited system for HeRo after evaluating multiple microcontroller boards, wireless technologies, sensors, actuators, and model designs for additive manufacturing.}

%%%%%%%%%%%%%%%%%%%%%%%%%%%%%%%%%%%%%%%%%%%%%%%%%%%%%%%%%%%%%%%%
\subsection{Mechanical Design}
%%%%%%%%%%%%%%%%%%%%%%%%%%%%%%%%%%%%%%%%%%%%%%%%%%%%%%%%%%%%%%%%
\hec{One of the primary steps in developing a mobile robot is modeling its mechanics.} The process defines the kinematic model of the robot, its actuation mechanism, and also its structural design.

% \textcolor{red}{and measurement}

\subsubsection{Kinematic Model} 
After reviewing the literature, we observed that most robots proposed for swarm robotics are based on the differential-driven model. In a nutshell, a differential wheeled robot is a mobile robot whose movement is based on two separately driven wheels placed on either side of the robot body. The robot changes its direction by varying the relative speeds of its \hec{wheels, and therefore it does not require an additional steering motor. We also decided to implement such model since it is very suitable} for designing a small, low-cost robot that requires good maneuverability and speed using a simple actuation mechanism. 

\subsubsection{Actuators and Encoders} 
\label{sec:actuators}
A convenient and affordable way to actuate the wheels of a differential robot is the use of geared DC motors. Besides actuating the robot, these motors enable the use of encoders for computing odometry, which is important for localization and  closed-loop motion control.

Although it is intuitive to use geared DC motors with encoders, such components can significantly increase the cost of the robot. Thinking of a low-cost solution, we decided to use small continuous servo motors to actuate the robot's wheels. Such motor is similar to geared DC motors with an h-bridge component, allowing motor speed control. %However, the reduced cost and compact size of some commercially available servo motors make them an interesting solution for our robot. 

Besides having reasonable precision for speed control, continuous SG90 servo motors contain a built-in microchip that controls the motor speed and direction using only pulse-width modulation (PWM) signals. In addition, this motor has a significant torque of $1.8$~kgf/cm, which allows us to use a $5$~cm diameter wheel to reach a maximum linear speed of $25$~cm/s with $0.3$~kgf/cm of torque without losing traction. 

Instead of directly connecting the wheel to the motor shaft, we attach the wheel to the robot chassis and use a gear mechanism (1:1) to transmit torque from the motor to the wheel -- this further reduces backlash and wheel misalignment that impact the encoder readings.

\begin{figure}[!t]
  \centering
  \includegraphics[width=0.95\linewidth]{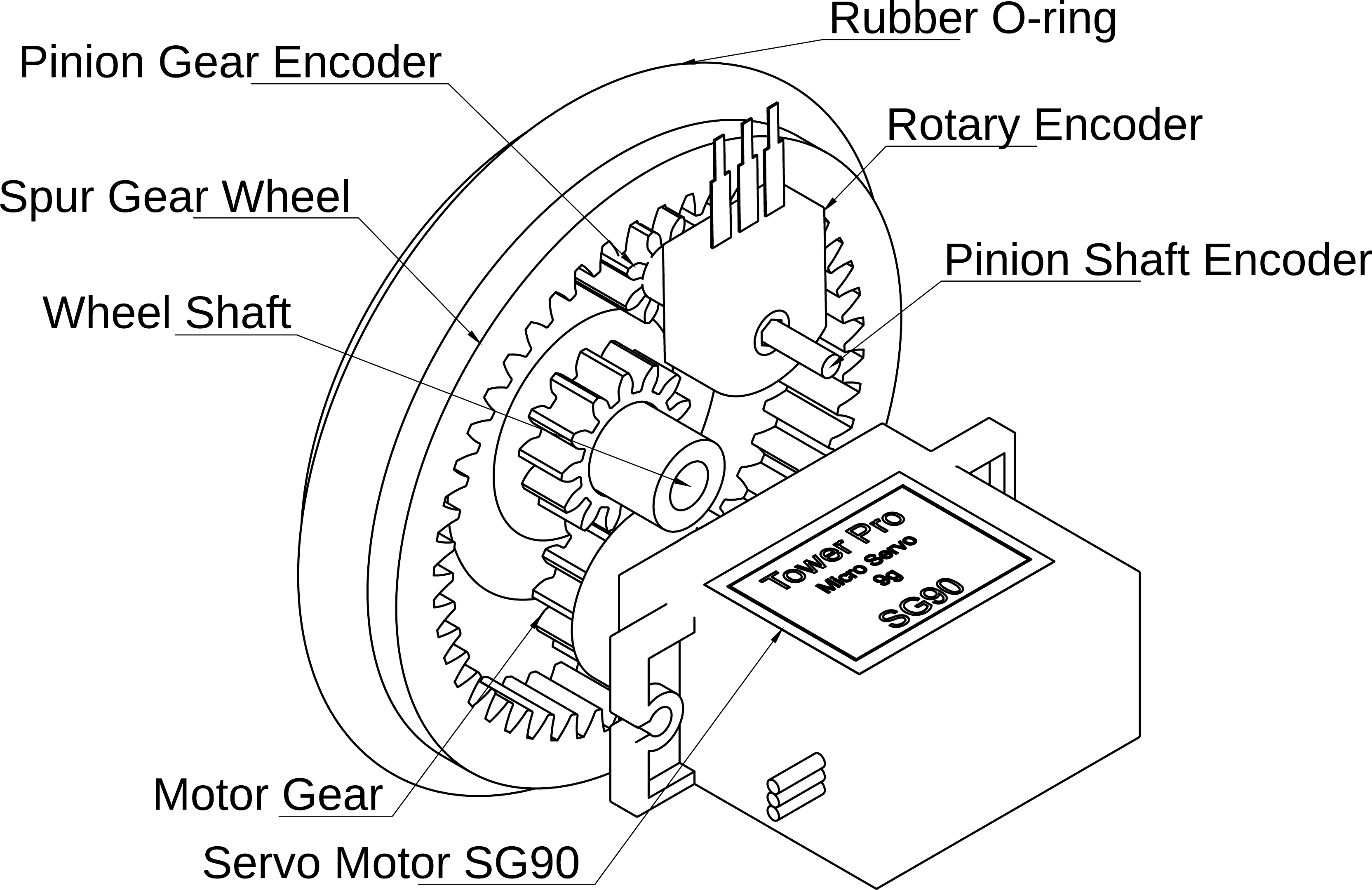}
  \caption{Torque transmission mechanism from the motor to wheel and wheel to the rotary encoder. 
  %\todo{aumentar tamanho do texto dentro da figura}
  %\todo{daria para separar levemente as palavras das linhas? para ficar similares a como estão na Fig 3}
  }
  \label{fig:herochassis}
\end{figure}

Since such motor does not have a built-in encoder, we took advantage of larger wheels to design a mechanical transmission system (1:6) between the wheel and a mechanical rotary encoder. By considering the low cost, availability, and compact form, we selected Kailh rotary encoders. This encoder is widely used on mouse devices as a step counter for the scroll button, and the simplest models, like the ones we use, can count $48$~steps per cycle. However, performing the wheel-encoder transmission (1:288) increases the wheel position measurement to $1.25\degree$ degrees of resolution, which means that the robot detects a wheel step of $0.54$~mm when it is moving. Fig.~\ref{fig:herochassis} illustrates the motor-wheel and the wheel-encoder transmission system. The motor, rotary encoder, and wheel shaft are fixed to the robot chassis, and the other parts are moving.

\subsubsection{Structural Design}
\label{sec:structuraldesign}
After defining the actuation mechanisms, we proceed with designing the robot's chassis. To facilitate assembly and further extensions, we design the robot's chassis to be modular and 3D-printable so one can easily print it using a conventional 3D printer. Overall, the robot structure comprises four main parts: the motor and board chassis, cover, and the e-Hat module. 

The \textbf{motor chassis} supports both motors and the wheels shaft -- where the wheels are attached. As the robot has two actuated wheels, it has only two contact points on the ground. To better adjust the balance and alignment of the robot, we created two screwable castor wheels. These castor wheels are attached to the motor chassis and allow us to fine-tune the robot's balance. On top of the motor chassis, we attach the \textbf{board chassis} that holds the encoders, battery, and the main processing board.

Considering further extensions, we take advantage of a modular chassis to coin the concept of \textbf{e-Hat}. Such part is attached on top of the robot and works as a shield extending the robot's sensorial or acting capabilities. For instance, we developed an e-Hat with an IMU sensor. Other components, such as a camera, sonar, actuator, or even a UWB transceiver for indoor localization, can also be used.

\begin{figure}[!t]
  \centering
  \includegraphics[width=0.99\linewidth]{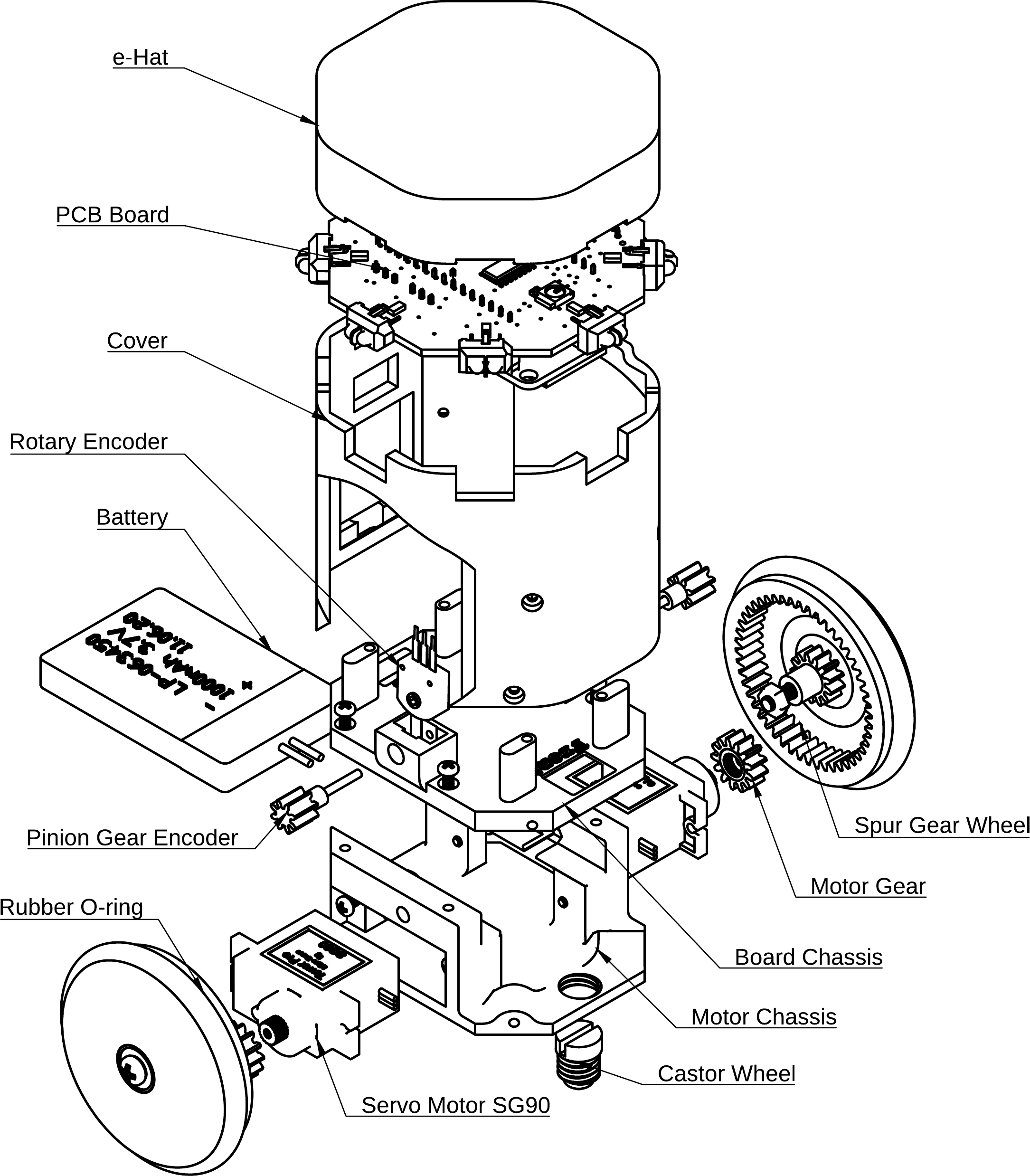}
  \caption{\hec{An expanded view of the robot's components and body parts.}}
  \label{fig:heroexpanded}
\end{figure}

Finally, we design a \textbf{cover} part to prevent dust accumulation inside the robot, protecting the main processing board and gears. In addition to protection, this part also enhances the robot's visual aesthetic. Fig.~\ref{fig:heroexpanded} shows an expanded view of the robot's design. An interactive CAD visualization is available at A360 platform\footnote{Robot Design CAD: \url{https://a360.co/3lWHiv0}}.

%%%%%%%%%%%%%%%%%%%%%%%%%%%%%%%%%%%%%%%%%%%%%%%%%%%%%%%%%%%%%%%%
\subsection{Electrical Design}
\label{sec:electricaldesign}
%%%%%%%%%%%%%%%%%%%%%%%%%%%%%%%%%%%%%%%%%%%%%%%%%%%%%%%%%%%%%%%%
% In parallel to the mechanical design \rez{process}, there is the electrical design of the robot. At this point, we define the electronic components built in the robot, such as processing unit, sensing, and power management.
\rez{In addition to the robot's mechanical design, we also present its electrical design. This process defines the electronic components built in the robot, such as processing unit, sensing, and power management.}

\subsubsection{Microcontroller}
One of the major decisions concerning the robot's electrical design is the selection of an appropriate microcontroller. This component defines the robot's computational capacity and the number and variety of components we can use.

After considering many alternatives, we select the Espressif ESP8266 as the main processing unit. This microcontroller is \hec{remarkably} inexpensive given its excellent processing power (32-bit 160 MHz) with 4 MB of memory. Also, it has a built-in Wi-Fi microchip that provides a fast IEEE 802.11 connection with a full TCP/IP stack. So, the robots can communicate among them or with a remote computer using robust and scalable protocols. Moreover, it is efficient, easy to program, and widespread in Maker communities, \hec{allowing others to easily develop customized modules for the robot.}

% ATMEGA microcontrollers have been one of the top alternatives in the development of small, low-cost robots. Such microcontrollers are relatively inexpensive, efficient, easy to program, and widespread in Maker communities. However, this series of microcontrollers does not have much processing power, making it difficult to embed several components in the robot. So recent alternatives are the STM32 family and Raspberry Pi Pico boards that feature $32$-bits ARM Cortex processors. Both microcontrollers are extremely powerful, given their compact size, but they are also more expensive. 

% After considering many alternatives, we select the Espressif ESP8266 microcontroller. Besides being extremely inexpensive, it has excellent processing power ($32$-bit $160$~MHz) with a built-in Wi-Fi microchip that provides a fast IEEE $802.11$ connection with a full TCP/IP stack. So, the robots can communicate among them or with a remote computer using robust and scalable protocols.

\subsubsection{Sensing}
%Once the microcontroller is set, we define the sensors and components usually required in swarm robots. 
% like most other swarm platforms in the literature.

A important requirement for a small robot used in swarm experiments consists of measuring distances to neighboring robots and obstacles. In HeRo, we chose to use infrared sensors for this due to their size and cost. 
We arrange eight IR transmitters and receivers (TCRT5000) around the circumference of the robot with $45\degree$ increments. \hec{We selected such sensor because it is cost-effective and has a reasonable resolution and range.}

\hec{Although it is not common using this type of sensor for long distances (> 10 cm), we found a strategy to increase its range without a considerable decrease in accuracy by exploring the sensor saturation time.} In fact, the manufacturer defines the sensor's maximum range\footnote{TCRT5000: \url{www.vishay.com/docs/83760/tcrt5000.pdf}} to only $2$~cm. If the sensor is continuously powered, its sensitivity degrades due to some electrical saturation phenomenon. However, if we turn on the sensor for only a few microseconds, take the reading, and then turn it off, we can increase the detection range to up to $20$~cm. 

To control the power of the sensor, we use a MOSFET component. Due to the limited number of ADC pins on the microcontroller -- it only has one pin with 10-bits of resolution -- we have to include an 8-channel analog multiplexer enabling the microcontroller to read all the eight IR sensors. This setup allows us to precisely measure distances, avoiding environment light interference since we can measure it by using only the IR receivers. 

\hec{As mentioned earlier, the robot has two pairs of rotary encoders coupled to the wheels by a drive mechanism. An encoder is a sensor that generates digital signals in response to the motion, providing information about position, velocity, and direction. As the typical mouse wheel internally works as a precise encoder, we took advantage of this inexpensive component (less than USD $0.10$) and used it as a robotic sensor. 
%The mechanical encoder is highly comparable with other low-cost encoders in resolution (e.g., optical and magnetic encoders). 
It comprises a conductive disc and three contacts that generate two square waves in quadrature when the encoder shaft rotates, enabling counting $48$ pulses per shaft revolution and also identifying the turn direction.}% \rez{Thus, this mechanical encoder is highly comparable with other low-cost encoders in resolution (e.g., optical and magnetic encoders).}}

In addition to the encoders, the robot also houses two WS2812b RGBA LED indicators used for status monitoring or debugging. These addressable LEDs have an IC built right into the LED, allowing communication via a one-wire interface (it uses one digital pin to control multiple LEDs in series). We can also control the brightness, and the color of each LED individually, which allows us to produce unique and complex effects for status in a simple way.%\footnote{Video of LEDs indicators: \url{https://youtu.be/Ra7JX8Awkc0}}.

\subsubsection{E-Hat}
\begin{figure}[t]
    \begin{subfigure}{.49\textwidth}
      \centering
      \includegraphics[width=0.46\linewidth]{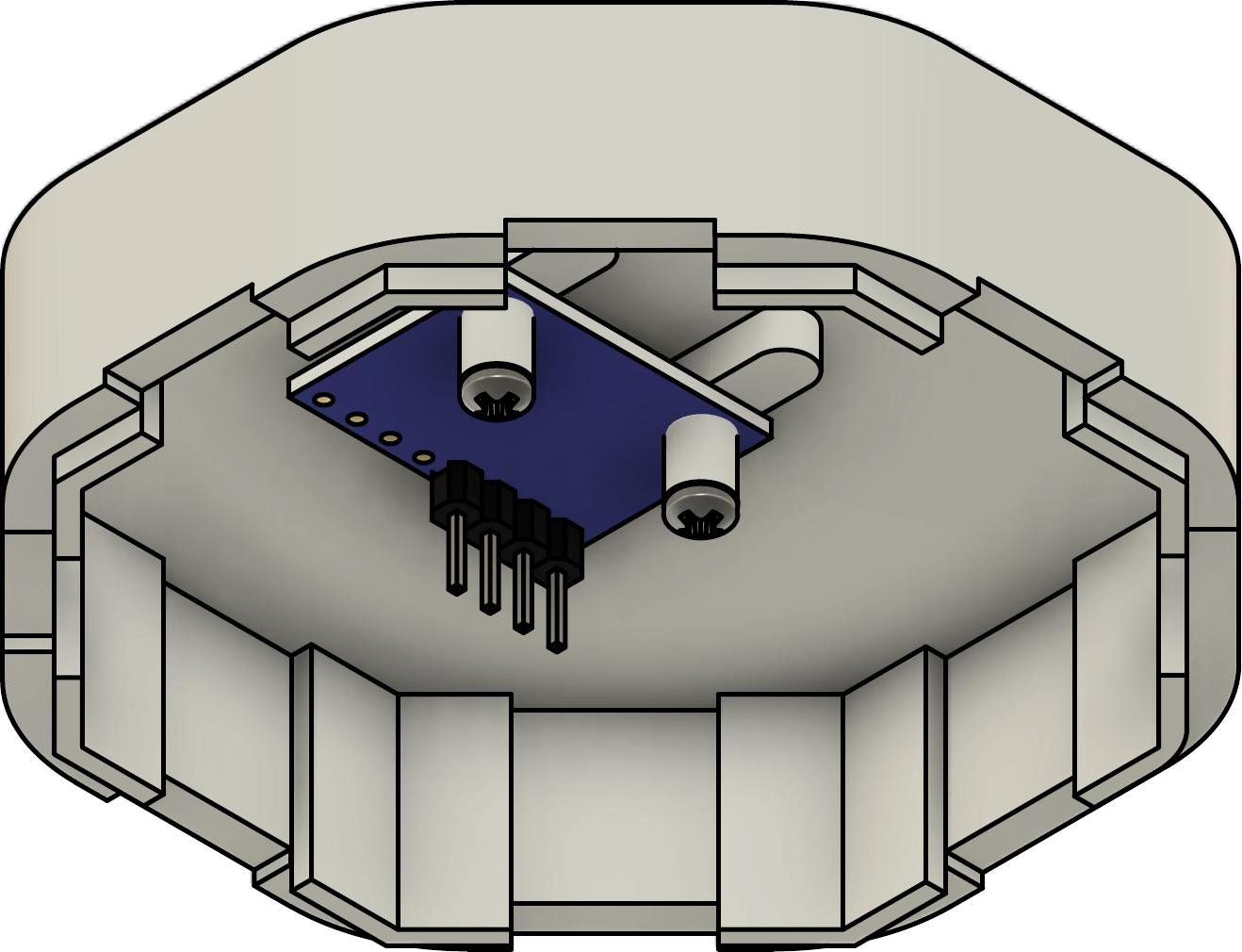}
      \caption{\hec{IMU e-Hat.}}
      \label{fig:ehatimu}
    \end{subfigure}
    \begin{subfigure}{.49\textwidth}
      \centering
      \includegraphics[width=0.92\linewidth]{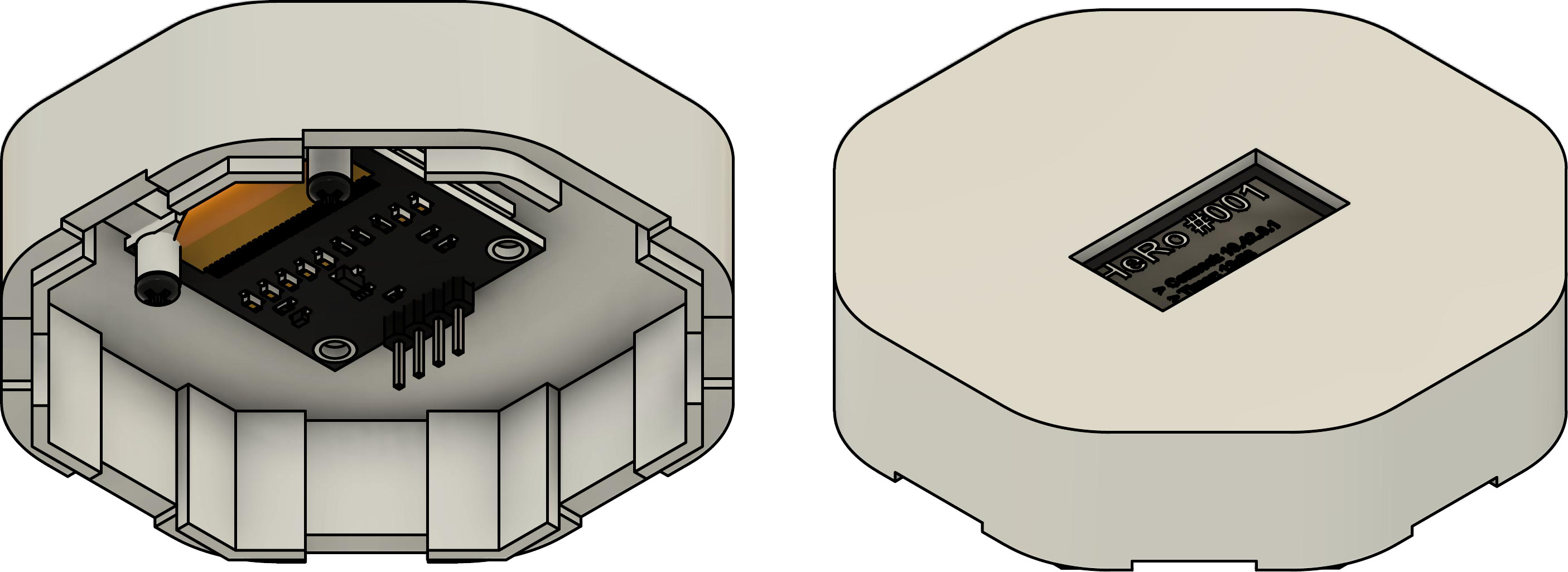}
      \caption{\hec{Display e-Hat.}}
      \label{fig:ehatdisplay}
    \end{subfigure}
\caption{\hec{Example of multiple e-Hat versions for the HeRo platform.}}
\label{fig:ehats}
\end{figure}

Besides the built-in features, the robot functionality can be extended using e-Hats. This robot module works as a shield, allowing users to create/customize specific modules for their applications. This module is coupled to a 4-pin bus on top of the robot configured to use I2C or UART protocols. In addition to communication, the bus also provides 5V (800 mA) to power the module. In this paper, we developed two e-Hats for demonstration and experimentation (see Fig.~\ref{fig:ehats}). The first one consists of an e-Hat IMU composed of an MPU6050 sensor with gyroscope and accelerometer that can be fused into the velocity and position estimation to account for odometry errors, such as the slip produced by the wheels. The second one is an e-Hat display that can be used either as a user interface or as part of a location system based on camera and fiducial markers.

\subsubsection{Power Supply}

Since the robot's autonomy is an important factor considering the time and number of experiments, HeRo uses a 3.7 V 1800 mAh Li-Po battery. The battery voltage is regulated by a MT3608 DC-DC \hec{step-up module, managing the board power supply to $5$~V.} These components enable the robot to perform up to 3 hours of experiments, considering the continuous use of all components. The motors are directly powered by a step-up power module avoiding any voltage drop impacting the robot's speed. In addition to this module, we also use a TP4056 module to recharge the battery using a USB cable.

\subsubsection{Assembly}
\begin{figure*}[!t]
  \centering
  \includegraphics[width=0.9\linewidth]{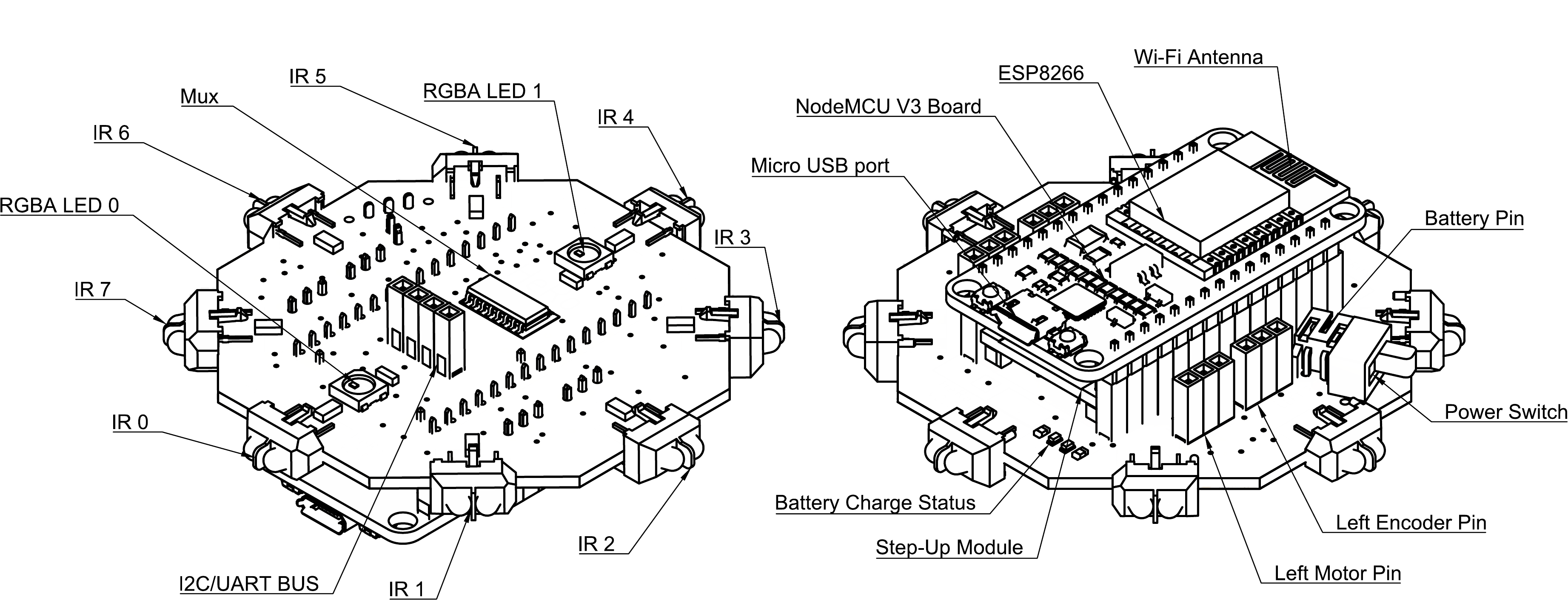}
  \caption{\hec{Overview of HeRo's PCB board.}}
  \label{fig:herodesign}
\end{figure*}

\hec{Because most of the robot's parts are off-the-shelf components, we decided to simplify the mounting and wrap them right on a PCB board. To increase reproducibility, we carefully design this PCB board so that even novice users can assemble it. As an alternative, the user can also assemble it in several specific PCB manufacturers, which nowadays attend at a highly affordable cost. Fig.~\ref{fig:herodesign} shows the proposed front and back views PCB board's design.}

%The figure shows the PCB board's design and, on the side, a PCB board purchased from one of these manufactures and assembled in our laboratory.

\subsection{Part Costs}
\rez{After defining the mechanical and electrical components of the robot and its assembly process, we can estimate its cost.} Table~\ref{table:parts_cost} gives a summary of the cost of the components used in HeRo. \hec{All part prices assume retail buying from standard part distributors on the Internet.} We expect this cost would be greatly reduced if parts are acquired in bulk directly from manufacturers.

\begin{table}[!t]
\centering
	\caption{Parts cost per robot \hec{unit}.}
	\label{table:parts_cost}
	\scalebox{0.42}{
	\resizebox{\textwidth}{!}{
\begin{tabular}{lll}
\toprule
\textbf{Parts} & \textbf{Quantity} & \textbf{Cost (USD)} \\
\midrule
Servo Motors SG-90 & 2 & 2.06 \\
Mouse Encoder 48 PPR & 2 & 0.10 \\
ESP8266 Nodemcu & 1 & 2.50 \\
Rubber O Ring 38mm & 2 & 0.10 \\
IR TCRT5000 & 8 & 0.68 \\
LED RGB WS2812b & 2 & 0.51 \\
IMU MPU6050 & 1 & 0.85 \\
LI-PO Battery 3.7 V 1800 mAh & 1 & 5.85 \\
PCB board and Components & 1 & 4.30 \\
3D Printer Parts (PLA) and Fastening & 1 & 1.50 \\
\midrule
\textbf{Total} &  & \textbf{18.72 USD}\\
\bottomrule
\end{tabular}}}
\end{table}

\begin{figure*}[!t]
  \centering
  \includegraphics[width=0.90\linewidth]{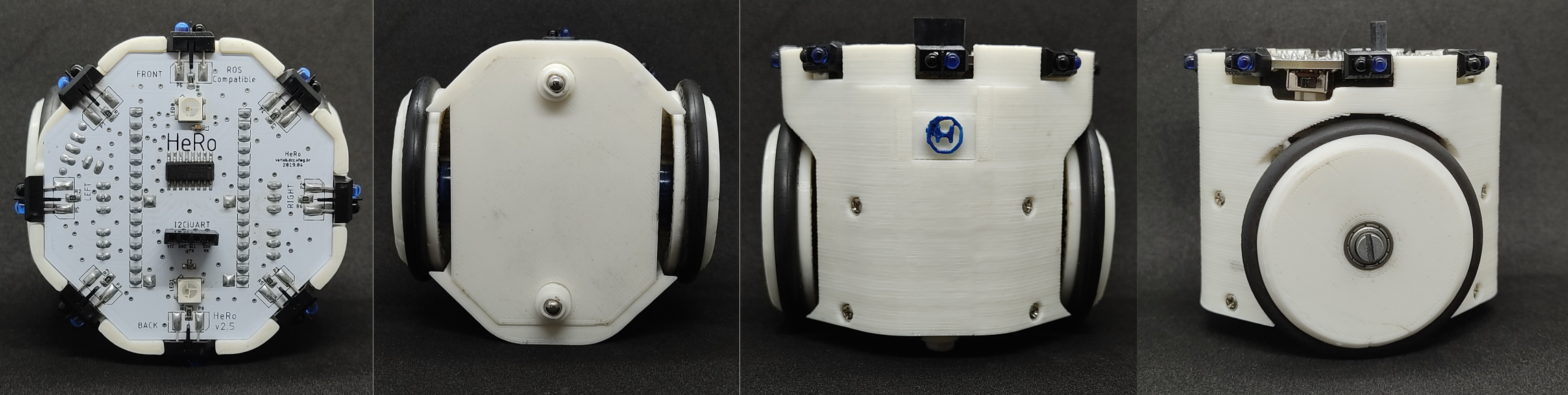}
  \caption{Top, bottom, front, and left views of \hec{the HeRo swarm platform.}}
  \label{fig:heroreal}
\end{figure*}

Finally, after carrying out the robot assemblies, we arrived at the result shown in the Fig.~\ref{fig:heroreal}.

\section{Software and Communication Architecture}
\label{sec:3_2_soft_comm_arch}
\rez{In addition to the physical robot, we also describe a computational framework to facilitate its use in swarm applications. In this section, we present a software and communication architecture that enables the programming of multiple robots. Moreover, we also present a simulated environment, useful in the early stages of application development.}

%%%%%%%%%%%%%%%%%%%%%%%%%%%%%%%%%%%%%%%%%%%%%%%%%%%%%%%%%%%%%%%%
\subsection{Software Architecture}
%%%%%%%%%%%%%%%%%%%%%%%%%%%%%%%%%%%%%%%%%%%%%%%%%%%%%%%%%%%%%%%%
% One of the most typical dilemmas in robot swarm systems is the convenience of promptly programming multiple robots. This work proposes a diversified architecture that enables the user to implement swarm applications promptly, efficiently, and elegantly. 

% The proposed software architecture allows the robots to be remotely programmed and executed remotely on a computer or locally on the robot. The first programming strategy follows the development strategy using middleware ROS, one of the most popular platforms for robotics. To make the robot ROS-Compatible we implement a firmware that interfaces the ROS by topics and services using ROSSerial protocol. Such an approach allows the users to create their applications in several programming languages ​​and control multiple robots remotely using TCP/IP networking. The second strategy follows the OTA (Over-The-Air) technology that allows the user to burn the robot's firmware via a Wi-Fi interface remotely.

% In the following, we detail the implementation of each module of our firmware as well as the communication process, programming framework, simulation and visualization. Figure \ref{fig:software_architecture} shows an overview of how these modules interact to each other as well as the user's applications. 

A typical robot swarm dilemma is how to program multiple robots \hec{quickly}, easily, and efficiently. A practice that has become very common in the experimentation stage is to use a master-slave architecture in which robots (slaves) remotely communicate with a computer (master) running the user application. This strategy is highly efficient as it does not require the user to burn the firmware every time he needs to change his application. Despite being convenient, remotely executing the application on a computer does not always capture effects that impact the application at the deployment level, e.g., local communication issues, low processing capacity, and other errors. Thus, strategies that use FOTA (Firmware Over-The-Air) technology enable users to remotely load their applications on the robot and run it directly on it.

In this work, we propose a flexible architecture to use any of the programming practices mentioned above. Our architecture is composed of a firmware compatible with ROS (Robot Operating System) and FOTA. In the following, we detail the software architecture.

\subsubsection{Firmware}

The firmware is one of the fundamental parts of the robot since it provides low-level control for the motors and access to the sensor's data. For HeRo, we chose to implement the firmware using the Arduino IDE. Such platform is easy to use and widespread in makers' communities, making the firmware easier to follow and modify. It also provides many libraries enabling us to control the microcontroller ports and handle actuators, sensors, and TCP sockets. 

\begin{figure*}[!t]
  \centering
  \includegraphics[width=0.6\linewidth]{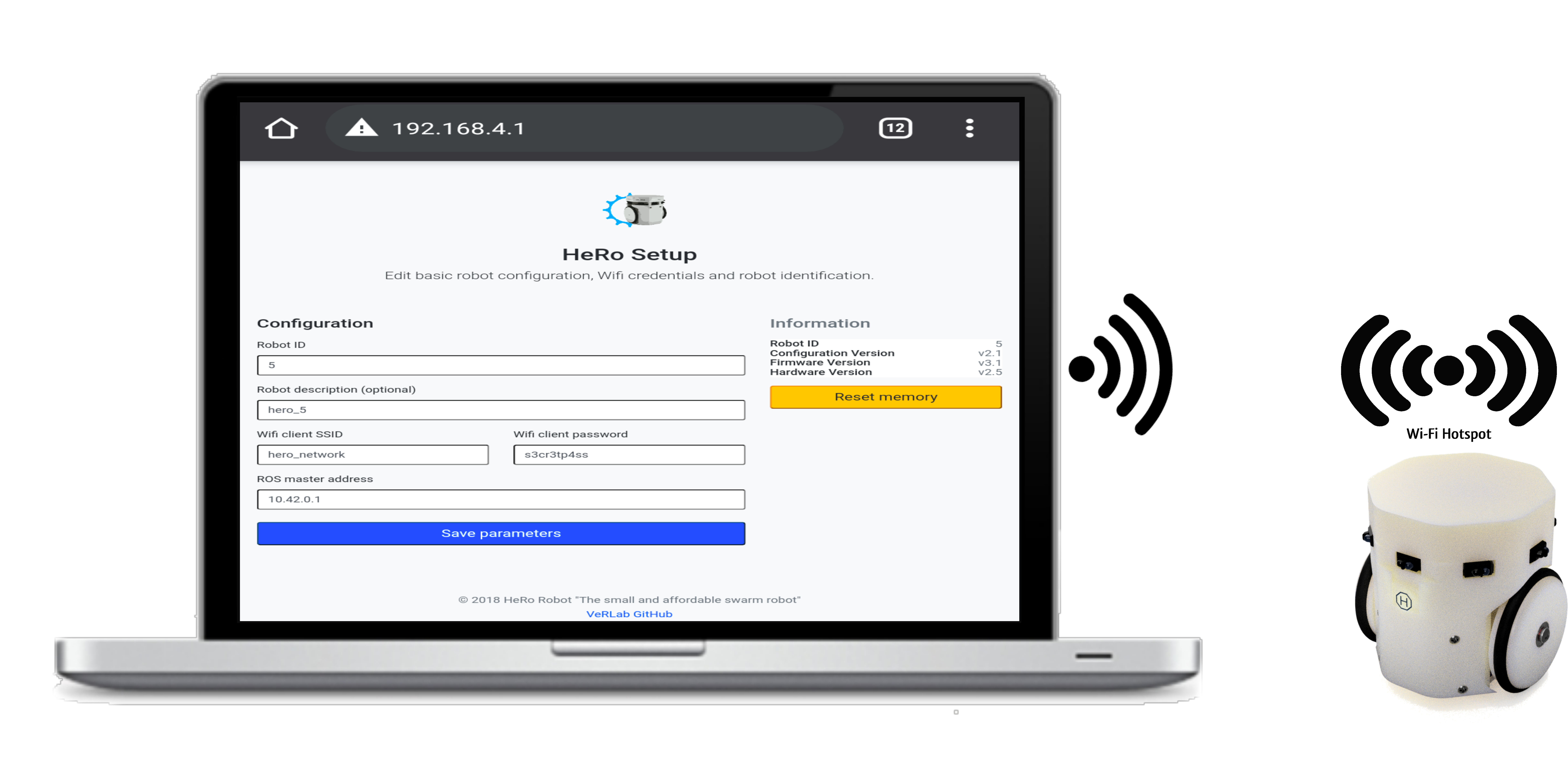}
  \caption{Configuring the robots to connect to a server running ROS only requires  the connection with an access point raised by the robot. Then, using a web configuration platform, robot connection parameters can be easily modified.}
  \label{fig:heroconfig}
\end{figure*}

The firmware is built on top of the rosserial framework allowing the robots to be compatible with the ROS middleware. Rosserial comprises different tools, including a protocol for wrapping standard ROS serialized messages and multiplexing multiple topics and services over network sockets. Such framework abstracts several communication concepts, allowing a compact and efficient implementation. In practice, the user only needs to configure a few communication parameters so that the robots can connect using TCP/IP networking with a remote computer running ROS. To facilitate this configuration process, we implemented a remote configuration mode for the robot using a web interface (Fig.~\ref{fig:heroconfig}). 
%To reach such a interface the users should connect to an access point created by the robot. 
\hec{To open this interface, the users should connect to an access point created by the robot. Once the robot is properly configured, it can connect to the user's network allowing the user to interact with the robot's features through topics and services. This entire process can be done in less than a minute.}

In addition to providing a master-slave communication architecture, the firmware is also composed of some basic modules that compute the robot's cinematic control, odometry, and sensor data. Next, we describe all \hec{these modules present in the firmware.}

\paragraph{\textbf{Sensors}}
In its basic form (without the e-Hat), the robot has eight infrared transceivers and two quadrature mechanical rotary encoders as sensors. The eight infrared sensors are mounted around the robot to provide a complete field-of-view of the environment.  These are connected to a 10-bits ADC port on the microcontroller through an 8-channel analog multiplexer allowing the estimation of the distance to an obstacle as well as the ambient light once we can control the infrared emitter. Obstacle detection and distance estimation use fundamental principles of electromagnetic radiation and its reflection. Mathematically, the reflected signal intensity measured with a sensor is modeled by~\cite{benet2002using}:

\begin{equation}
    \label{eq:infrared}
    s(x, \theta) = \frac{\alpha}{x^2} cos(\theta) + \beta,
\end{equation}

where $s(x, \theta)$ is the output value of the sensor, $x$ is the distance to the object, and $\theta$ is the angle of incidence with the surface; the model variable $\alpha$ includes several parameters such as the reflectivity coefficient, output power of the emitted IR light and the sensitivity of the sensor and it is estimated empirically; $\beta$ is the offset value of the amplifier and ambient light effect and it is measured regularly after performing the Equation~\ref{eq:infrared}. 

As mentioned, HeRo has two quadrature encoders attached to each wheel. A quadrature encoder, also known as an incremental rotary encoder, is commonly used to measure the speed and direction of a rotating shaft. The encoders' channels are connected to the microcontroller's interrupt pins. \hec{Each pulse calls an interrupt routine in the microcontroller, increasing an independent counter variable} to estimate how far each wheel has turned. To estimate the velocity of the wheel, we measure the frequency of the pulses. The output of the encoders is used as an input to a controller for closed-loop motion control and for localization.

\paragraph{\textbf{Odometry}}
Odometry is the most used method for determining the momentary position of a mobile robot. In most practical applications, odometry provides easily accessible real-time positioning information in-between periodic absolute position measurements. Several different types of sensors are commonly used for odometry. \hec{This work addresses odometry by placing encoders on each wheel} and counting how far each wheel has turned. Using these two measurements we can estimate how far the robot has moved forward and its heading. The distance traveled by the robot is computed by the average of how much each wheel has turned and is presented in Equation~\ref{eq:wheel_travel}. On the other hand, the heading of the robot is estimated (assuming insignificant wheel slip) from the difference of these displacements over the distance between the wheels and is presented in Equation~\ref{eq:wheel_turn}.
%\hec{Both equations are presented below:}
\begin{equation}
\label{eq:wheel_travel}
\Delta L =  \frac{\Delta\theta_r r_r + \Delta\theta_l r_l}{2},
\end{equation}
\begin{equation}
\label{eq:wheel_turn}
\Delta\theta =  \frac{\Delta\theta_r r_r - \Delta\theta_l r_l}{l},
\end{equation}
where $\Delta\theta_r$ and $\Delta\theta_l$ represents how much each encoder has turned in the time interval; $r_r$ and $r_l$ are the right and left wheel radius; and $l$ represents the distance between the wheels of the robot.

\hec{Once we have computed how far the robot has traveled and turned, we can integrate this information to estimate its current pose.} Considering the pose of the robot at time $t$ in a plane is given by the state vector $X(x(t), y(t), \theta(t))$, the next pose of the robot at time $t + 1$ is given by 
\begin{equation}
\scalemath{0.76}{
X(t+1) = 
\begin{bmatrix}
x(t+1) \\ 
y(t+1) \\ 
\theta(t+1)
\end{bmatrix}
=
\begin{bmatrix}
x(t) + \frac{\Delta L}{\Delta \theta} (sin(\theta(t) + \Delta\theta) - sin(\Delta \theta)\\ 
y(t) + \frac{\Delta L}{\Delta \theta} (cos(\theta(t) + \Delta\theta) - cos(\Delta \theta)\\ 
\theta(t) + \Delta\theta
\end{bmatrix}.
}
\end{equation}
However, this model does not explain the position increment when the robot performs straight forward movements. In this case,  we have $\Delta\theta = 0$ and this makes $\frac{\Delta L}{\Delta \theta}$ undefined. In order to approach this special case, we test this condition and if it occurs the following model is computed for the odometry
\begin{equation}
X(t+1) = \begin{bmatrix}
x(t+1) \\ 
y(t+1) \\ 
\theta(t+1)
\end{bmatrix}
=
\begin{bmatrix}
x(t) + \Delta L  cos(\theta(t))\\ 
y(t) + \Delta L  sin(\theta(t))\\ 
\theta(t)
\end{bmatrix}.
\end{equation}

\paragraph{\textbf{Motion Control}}

% Previously, we define the robot as a two-wheel differential-drive mobile robot composed of two servo motors each with a quadrature encoder. A differential wheeled robot is a mobile robot whose movement is based on two separately driven wheels placed on either side of the robot body. It can thus change its direction by varying the relative rate of rotation of its wheels and hence does not require an additional steering motor. To control the velocity of the mobile robot, the mathematical model of the robot kinematics and a suitable controller should be implemented. Mathematical modeling of the robot kinematics is one of the major parts of the control system in robot design. In short, this model takes the velocities of the actuators and transform them into the generalized coordinate vector. To control our robot, the high-level velocities with respect to the tangential velocity on each wheel is computed by the inverse kinematic model, 

%To control our robot, the high-level velocities with respect to the mobile robot are related to the tangential velocity on each wheel by the inverse kinematic model, 

\hec{Previously, we defined the robot as a two-wheel differential-drive mobile robot composed of two servo motors, each with a quadrature encoder. The robot can change its direction by varying the relative rotation rate of its wheels and hence does not require an additional steering motor. The inverse kinematic model computes the tangential velocity on each wheel given a set of high-level velocities using the following equations,}
\begin{equation}
\begin{bmatrix}
v_{l}(t) ~ v_{r}(t)
\end{bmatrix}^T 
=  \begin{bmatrix}
\frac{2v(t) - l\omega(t)}{2} ~ \frac{2v(t) + l\omega(t)}{2}
\end{bmatrix}^T,
\end{equation}
where $v_{l}(t)$ and $v_{r}(t)$ are the tangential velocities of the left and right wheels; $v(t)$ and $\omega(t)$ are the high-level linear and angular velocities of the mobile robot; and $l$ is the distance between the left and right wheels.

\begin{figure}[]
      \centering
      \includegraphics[width=0.95\linewidth]{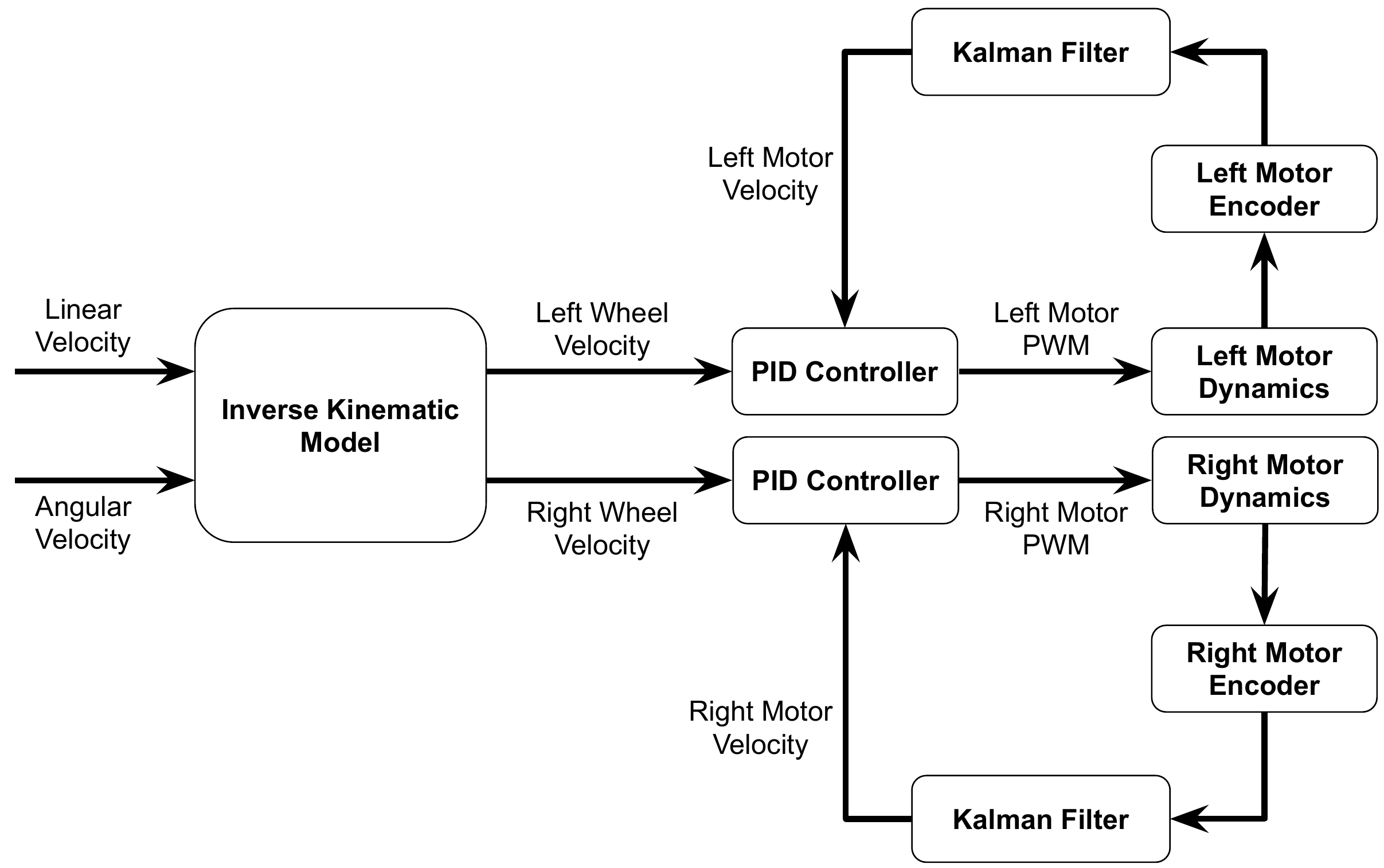}
      \caption{Diagram of the controller used for closed-loop velocity control.}
      \label{fig:overview_motion_control}
\end{figure}

Once computed the tangential velocities on each wheel, we need to control the motors \hec{so} that they maintain these velocities. The proportional-integral-derivative controller (PID controller) is the most common control algorithm used for this application. It \hec{can} correct the present error through proportional action, eliminate steady-state offsets through integral action, and better estimate the future trends through a derivative action. The mathematical model of a PID is defined by
\begin{equation}
u(t) = K_{p}e(t)+K_{i} \int _{0}^{t}e(t')\,dt' + K_{d}{\frac {de(t)}{dt}},
\end{equation}
where $u(t)$ is the control signal to each motor; $e(t)$ is the error regarding the tangential velocity of each wheel; and $K_{p}$, $K_{i}$ and $K_{d}$, all non-negative, denote the coefficients for the proportional, integral, and derivative terms, respectively. \hec{To estimate the error, we subtract the tangential velocity obtained by the inverse kinematic model from the tangential velocity estimated by the encoders. A Kalman} filter is used to reduce the reading noise estimated by the encoder. Figure \ref{fig:overview_motion_control} shows an overview of this controller.

%% Atualizar figura.... filtro de kalman

% PID tuning comprises of setting the best value of $K_p$, $K_i$ and $K_d$ of the PID so that the system performance can be increased. The proportional terms have the effect of reducing the rise time but never eliminate the steady-state error. The integral terms have the effect of eliminating this steady-state error, but can make the transient response worse. The derivative term has the impact of increasing the stability of the system, reducing the overshoot, and improving the transient response. After properly tuning the PID terms our controller is able to perform the velocity control on the robot's wheels. Figure \ref{fig:overview_motion_control} shows an overview of this controller.

\subsection{Communication Architecture}
%\todo{falar de west2018ros, trabalho que mostra que utilizar protocolo rosserial e escalavel para swarm...} \cite{west2018ros}

% fundir isso com o que foi falado antes...
In order to provide communication between a workstation and the robots, we implement HeRo as a \hec{ROS-compatible robot by connecting them using TCP/IP.}

The Robot Operating System~\cite{quigley2009ros} is an open-source, meta-operating system for robotic applications. It provides similar services expected from a typical operating system, including hardware abstraction, low-level device control, implementation of commonly-used functionality, message-passing between processes, and package management. It also provides tools and libraries for obtaining, building, writing, and running code across multiple platforms.

The communication is conducted by a publish/sub\-scribe model, where topics made up of predefined message structures can be communicated between multiple nodes (processes) in the network. These topics, for example, odometry, can be accessed by any node in the network, allowing for easy scalability of publishers and subscribers. In this way, robots can readily communicate with other robots in the network in a well-defined way. 

%However, most instances of swarm robots, including HeRo, are very limited to process native ROS instance. In %nearly all cases, researchers are unable to provide the functionalities of the swarm robot in ROS, since they are not able to locally run the ROS core -- a collection of process and programs that are pre-requisites of a ROS-based system.

\hec{However, most swarm robots, including HeRo, are unable to process a full-fledged native ROS instance given the restricted CPU resources. To integrate these functionalities to less powerful microcontrollers without a complete instance of ROS, we implemented the communication module over the rosserial protocol. Rosserial\footnote{Rosserial: \url{http://wiki.ros.org/rosserial}} is a protocol for wrapping standard ROS serialized messages and multiplexing multiple topics and services over a network socket.}

\begin{figure}[!t]
      \centering
      \includegraphics[width=0.96\linewidth]{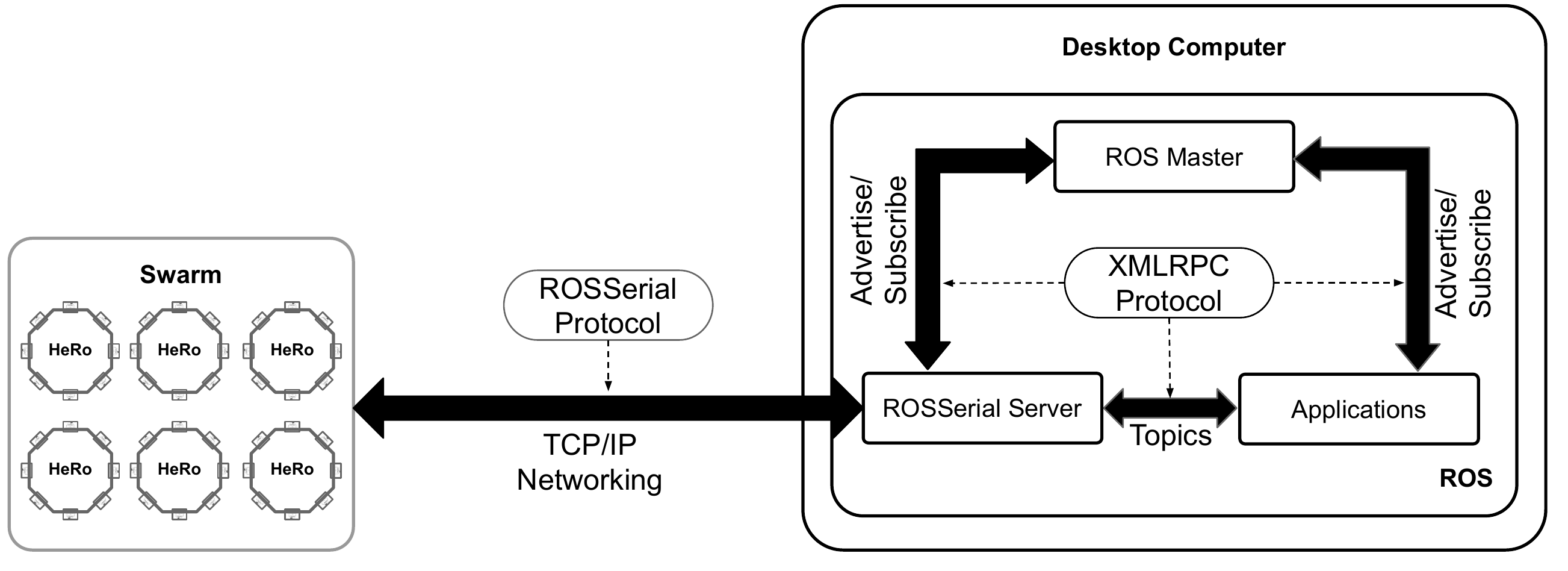}
      \caption{An overview of the communication process. The robot's microcontroller acts as a bridge to the sensors and actuators and then rosserial acts as another bridge from the microcontroller to ROS.}
      \label{fig:communication_architecture}
\end{figure}

In short, the rosserial nodes convert data from normal structured XMLRPC protocol handled by TCP natively in ROS to serialize data out to the microcontroller. This node also deserializes data from the microcontroller back into the correct message structures to be sent around the conventional ROS network. Figure \ref{fig:communication_architecture} shows an overview of this process. %The rosserial protocol has been validated as a reliable and scalable communication method for swarm systems~\cite{west2018ros}.

\hec{In theory, the network's bandwidth limits the number of connected robots: as more robots are added, more connections are made, taking up capacity. However, we did not observe any overhead communicating with multiple robots, even using a consumer-grade wireless network router. A typical network addresses 254 devices, but network techniques (e.g., subnets) allow increasing this limit as much as we need. A complete study showing the reliability and scalability of using this protocol for swarm robots is present in~\cite{west2018ros}.}

\subsection{Simulation and Visualization}
% melhorar essa parte, atualizar essa parte

%\hec{Although programming directly on a real robot can provide genuine feedback and is more impressive than simulations, simulators play an essential role in robotics research as tools for quick and efficient testing of new concepts, strategies, and algorithms. For this reason, we also provide a simulated model of HeRo.}

The execution of simulations plays an essential role in robotics research as a tool for quick and efficient testing of new concepts, strategies, and algorithms. Moreover, good visualization tools are very important during the experiments to better track and observe the robot execution. In this sense, we also developed a simulation model of HeRo that can be used together with Gazebo and RViz.

\subsubsection{Gazebo Simulator}
In order to make simulations with our robots in ROS, we decide to use Gazebo since it is fully integrated with ROS. Gazebo~\cite{koenig2004design} is a multi-robot simulator for complex indoor and outdoor environments. It is suitable for simulating a population of robots, sensors, and objects in a three-dimensional world. \hec{ROS and Gazebo use the 3D model of a robot or its parts, whether to simulate or visualize them, through the XML files, called Unified Robot Description Format (URDF). This file describes all the structures of the robot, such as its parts, joints, dimensions, and texture, among others.}

\begin{figure}[]
      \centering
      \includegraphics[width=0.98\linewidth]{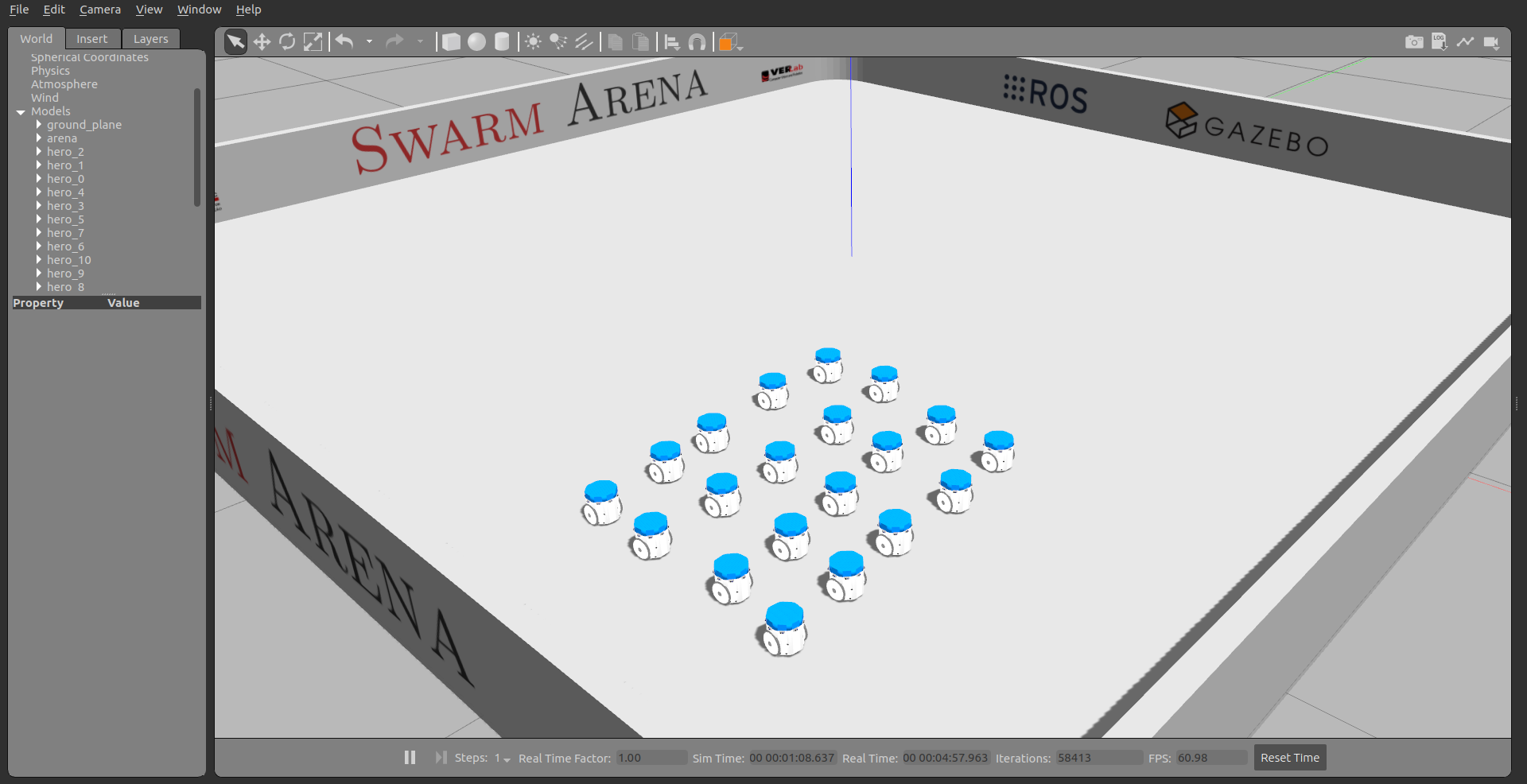}
      \caption{Multiple instances of HeRo being simulated in the Gazebo simulator.}
      \label{fig:hero_gazebo}
\end{figure}

After describing the robot using the URDF file, creating a simulated model using  the built-in plugins provided by the Gazebo is straightforward. However, this approach is inefficient when simulating multiple robots and requires a high computational cost. To reduce consumption and increase the number of simulated robots, we designed a compact plugin that implements all the robot's functionalities. In this way, we optimized the maximum processing performed by each simulated robot without overloading Gazebo's physics engine. Fig.~\ref{fig:hero_gazebo} shows multiples instances of HeRo being simulated in the Gazebo simulator.

\subsubsection{Robot Visualization Tool}
\begin{figure}[]
      \centering
      \includegraphics[width=0.99\linewidth]{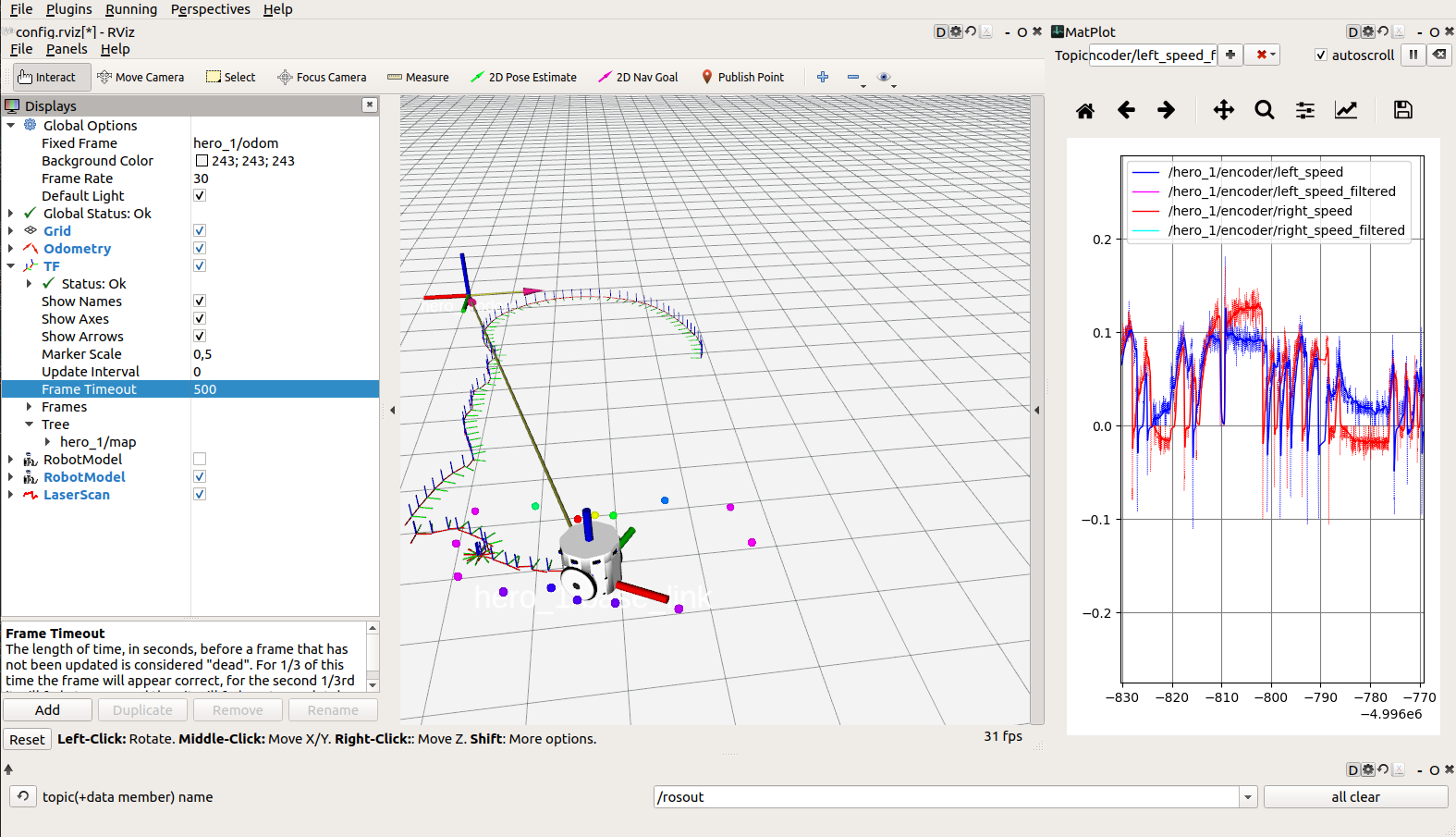}
      \caption{RViz showing a single HeRo robot. RViz is a 3D visualization tool for ROS allowing control and observe the current state of the robot.}
      \label{fig:hero_rviz}
\end{figure}

\rez{In addition to the simulation, it is also essential to have a robot visualization tool that shows the state of sensors and actuators during the experiment. In ROS, we can visualize the robot's state using the RViz visualization tool. This tool provides 3D visualization of the robot by loading the URDF file and can project sensors data obtained by the ROS topics such as odometry, laser, and IMU using plugins. Note that RViz is not a simulator but only a visualization tool. In this way, the robot visualized in this tool can be real or simulated depending only on who publishes the information. Fig.~\ref{fig:hero_rviz} shows an example of viewing a real robot in Rviz. In the image, we can see the 3D model of the robot overlaying a colored axis that indicates the robot's pose relative to an initial frame (colored axis in the background of the scene). The sequence of small axes indicated the temporal pose of the robot computed by the odometry. Colored spheres around the robot can move closer or further away from the robot and indicate the readings of the distance sensors. On the right, we can follow the linear velocity of each robot's wheel.}

% \hec{Once we have a 3D model and the robot's descriptions we generate the configuration file necessary for RViz. We associate the robot's parts, such as chassis and wheels, named links, through joints. Once this description file is generated, we can visualize the robot in ROS using the RViz visualization tool. This tool provides 3D visualization of the robot by loading the URDF file and associating its joints to the transform tree, which animates the model. In addition to displaying the robot model, RVIz can project sensors data such as} odometry, laser, and IMU using plugins. Note that RViz is not a simulator but only a visualization tool. In this way, the robot visualized in this tool can be real or simulated depending only on who publishes the information. Fig.~\ref{fig:hero_rviz} shows an example of viewing the robot in Rviz. Colored objects in the rectangular shape can move closer or further away from the robot, indicating the readings from the distance sensors. The grey shape on the floor represents part of a map built by the robot.

\subsection{Programming}
The communication architecture defined for our robot allows it to be programmed in two different ways: using the ROS framework or the OTA firmware technology.

In the first mode, we can program and run applications on a server, which communicates and controls each robot in a decentralized way. In other words, each algorithm is executed in a process on the server, and this process has access via Wi-Fi with its respective robot. This mode is very convenient and scalable in the early stages of testing with multiple robots. Furthermore, using the ROS framework for implementation, we have a series of tools and may use different programming languages.

\rez{On the other hand, processing algorithms remotely is not always suitable for robot swarm applications. In this case, the algorithm must run directly on the robot, maintaining the convenience of programming the robots simultaneously. In this programming mode, we use the OTA technology to burn the firmware in several robots using Wi-Fi. This process uses the Arduino IDE to implement and compile the application and then uses the command line to transmit the binary code for robots. Despite being convenient, this mode is limited in terms of the availability of high-level tools. In addition, it requires using a programming language compatible with the microcontroller, in this case, C/C++.}

% On the other hand, processing algorithms remotely is not always ideal for robot swarm applications. In this case, the algorithm must run directly on the robot, maintaining the convenience of robot programming. The robot also allows remote burning of the firmware using Wi-Fi. However, it is limited in terms of the availability of high-level tools and programming languages. In this module, one can use the Arduino IDE for programming the robots.

% firmware ros compatible
% controls & filter
% architecture
% communication
% web setup interface
% scalability
% rviz
% Gazebo

% \todo{\\
% - Colocar examples programacao ros e FOTA}
% ros examples
% arduino ide fota

%\section{Performance metrics and Experiments}
\section{Performance Evaluation}
\label{sec:experiments}
% Experiments
This section presents a series of experiments that evaluate our robot's performance as a capable swarm robot. Initially, we analyzed motion control and evaluate the robot's odometry in comparison to the {\em E-puck}, a popular commercial swarm robot. We also evaluate and discuss the performance and scalability of communication when using ROS and make an analysis of the robot's energy consumption when demanding different types of applications. 
%In the, we also present some case studies illustrating the application of the robot in typical swarm robotics tasks.

% Motion Control %
\subsection{Motion Control Analysis}
\rez{In this experiment, we evaluated the low-level control of the robot, which consists of ensuring that each wheel reaches the output speed. As previously detailed, the wheel speed control uses a PID controller. The feedback information consists of the current wheel speed, estimated by the encoder's readings and filtered by the Kalman filter.}

\begin{figure}[t]
  \begin{subfigure}{.5\textwidth}
    \centering
    \includegraphics[width=0.8\linewidth]{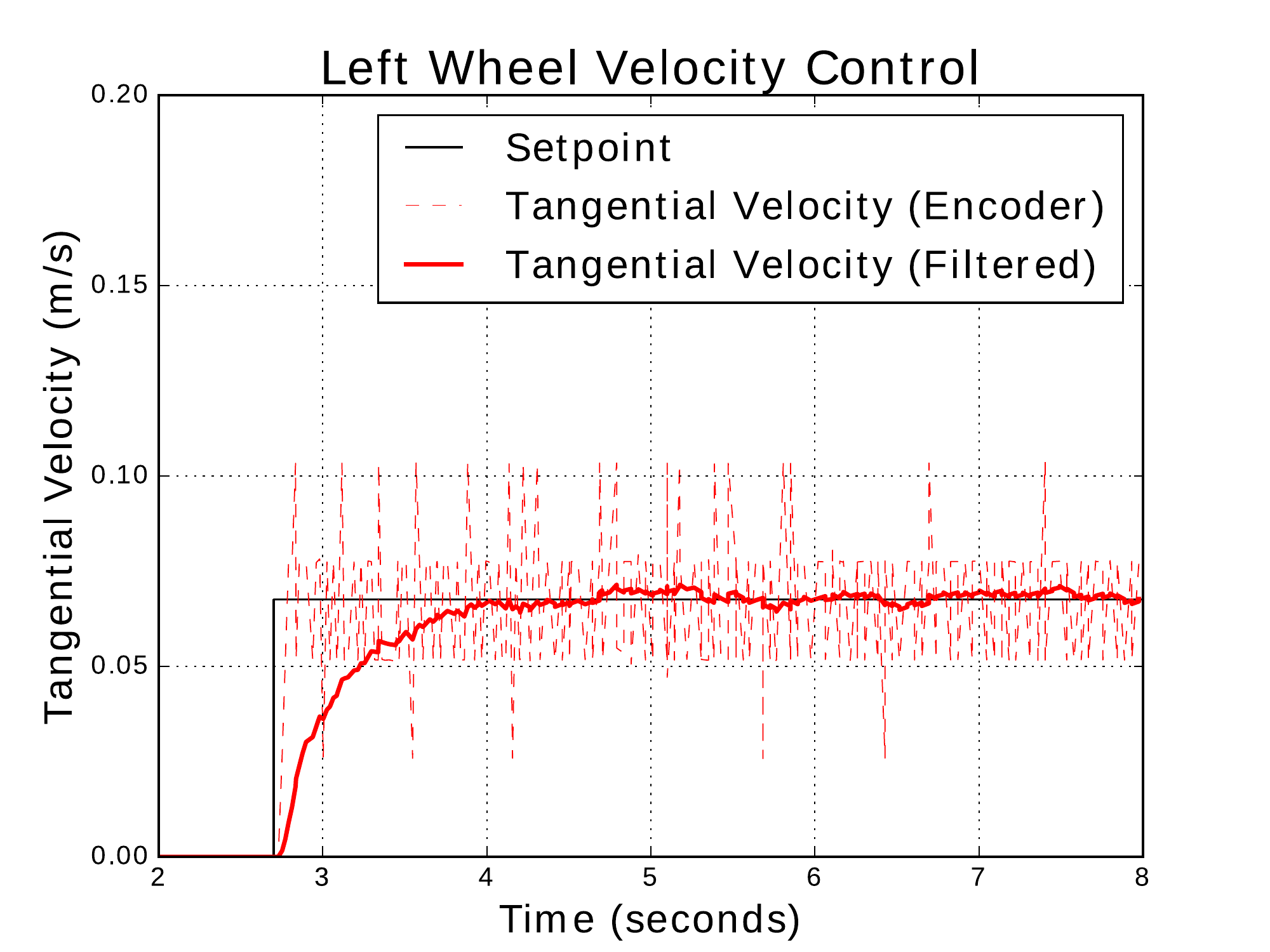}
    \caption{}
    \label{fig:motion_control_l}
  \end{subfigure}
  \begin{subfigure}{.5\textwidth}
    \centering
    \includegraphics[width=0.8\linewidth]{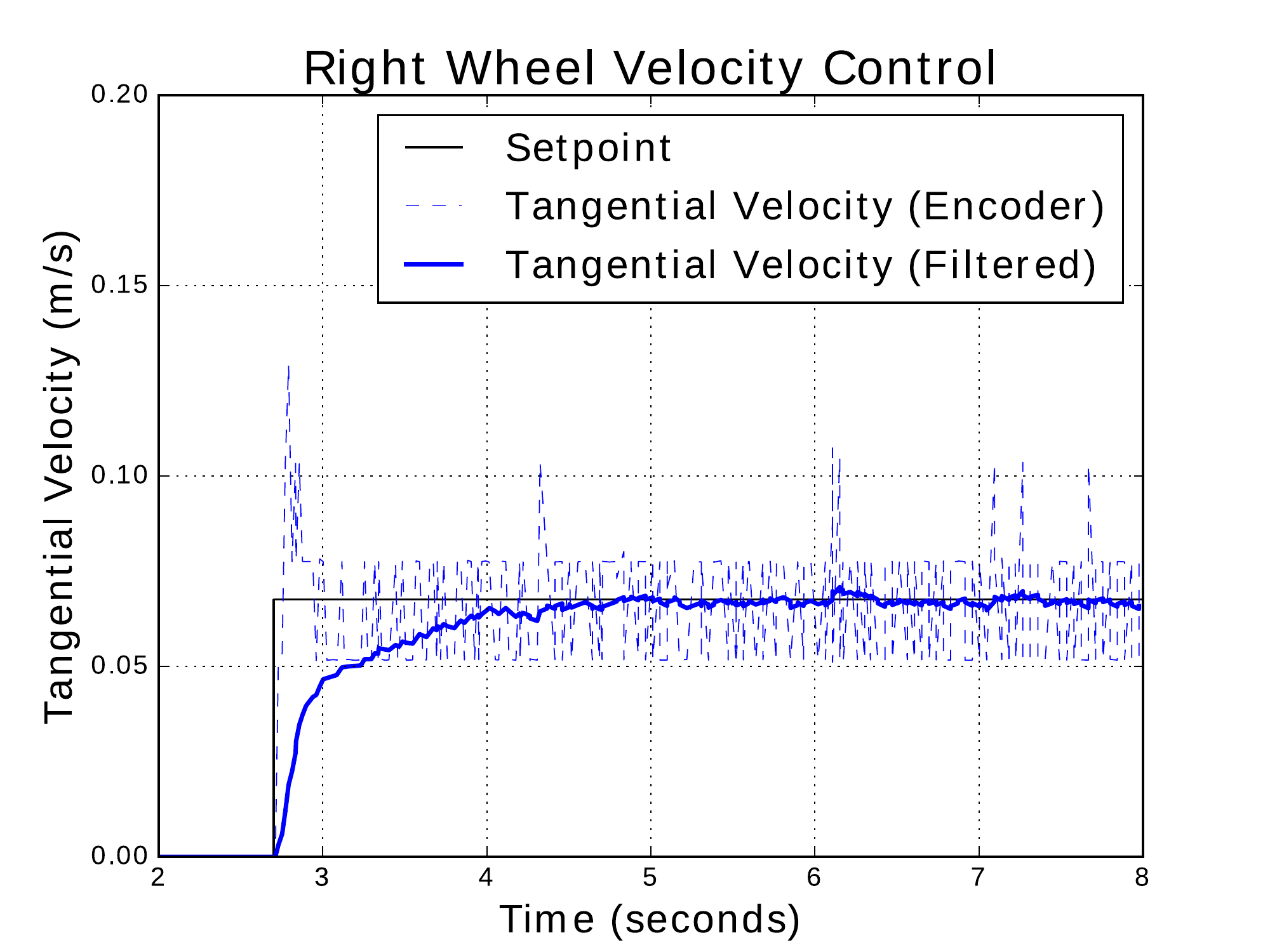}
    \caption{}
    \label{fig:motion_control_r}
  \end{subfigure}
  \caption{\rez{Analysis of the wheel speed control performed by the PID controller for the left (a) and right (b) wheels. The setpoint is set to $0.0675~m/s$ at $2.7~s$ and after approximately $1.3~s$ we estimate the mean speed of $0.0680 \pm 0.0015~m/s$ at the left wheel and $0.0667 \pm 0.0013~m/s$ at the right wheel.}}
  \label{fig:motion_control}
\end{figure}

\rez{One way to evaluate the controller's performance is to check its response time and residual ripple. Thus, we observe the controller's behavior when starting with the wheels halted, and then we set a desired tangential velocity (or setpoint) to $0.0675$~m/s in both wheels at time $2.7$~s. Fig.~\ref{fig:motion_control} shows the performance of each controller. }

\rez{As expected, we observe a similar response time for both wheels, reaching the setpoint after approximately $1.3$~s. Although we can make the system more responsive, we opted for a more conservative controller with no overshoot to avoid sudden movements making it difficult to control the robot. After the rise time, both wheels maintain the velocity with a maximum absolute error of $0.0012 \pm 0.0008~m/s$ for the left wheel and  $0.0013 \pm 0.0010~m/s$, which is remarkable considering the low-cost components used within the robot.}

% A dual PID controller implements the velocity control on each wheel of the robot. The encoders attached to each wheel estimates the tangential speed of the wheel, and a Kalman filter is used to reduce the reading noise. Fig.~\ref{fig:motion_control} shows the performance of each controller. 

% In this experiment, we set the desired tangential velocity (or setpoint) to $0.0675$~m/s in both wheels at $2.7$~s. As expected, we observe a similar response time for both wheels, reaching the setpoint after $1.3$~s. Although we can make the system more responsive, we opted for a more conservative controller with no overshoot to avoid sudden movements making it difficult to control the robot. After the rise time, both wheels maintain the velocity with a maximum absolute error of $0.0012 \pm 0.0008~m/s$ for the left wheel and  $0.0013 \pm 0.0010~m/s$, \hec{which is remarkable considering the low-cost components used within the robot.}

% Localization
\subsection{Localization}
In this experiment, we evaluate the odometry of our robot and compare the results with the one obtained by the E-puck~\cite{mondada2009puck}. To better analyze the capabilities of these robots, we implement the same odometry model and use the same experimental setup. 

This comparison is interesting because E-puck uses relatively expensive stepper motors against the inexpensive servo motor used in our robot. The E-puck computed its odometry by counting steps commanded to each motor, reaching a maximum resolution of 1024 steps per wheel revolution, \hec{which is more than the provided by} our encoders (288 steps per revolution). However, there is no feedback when the wheel steps, so it probably produces more false-positives counts.

To measure the pose estimation accuracy for both robots, we use the OptiTrack tracking system\footnote{OptiTrack: \url{http://optitrack.com/}} as a ground truth reference. The trajectory performed by both robots is a rectangular shape (1.3x1.1 m), delimited by four points. We control each robot to move to the four points consecutively until it completes three loops. Both robots traveled equal distances while maintaining the same velocity to keep the comparison as reliable as possible.

\begin{figure*}[!t]
  \begin{subfigure}{.32\textwidth}
    \centering
    \includegraphics[width=0.99\linewidth]{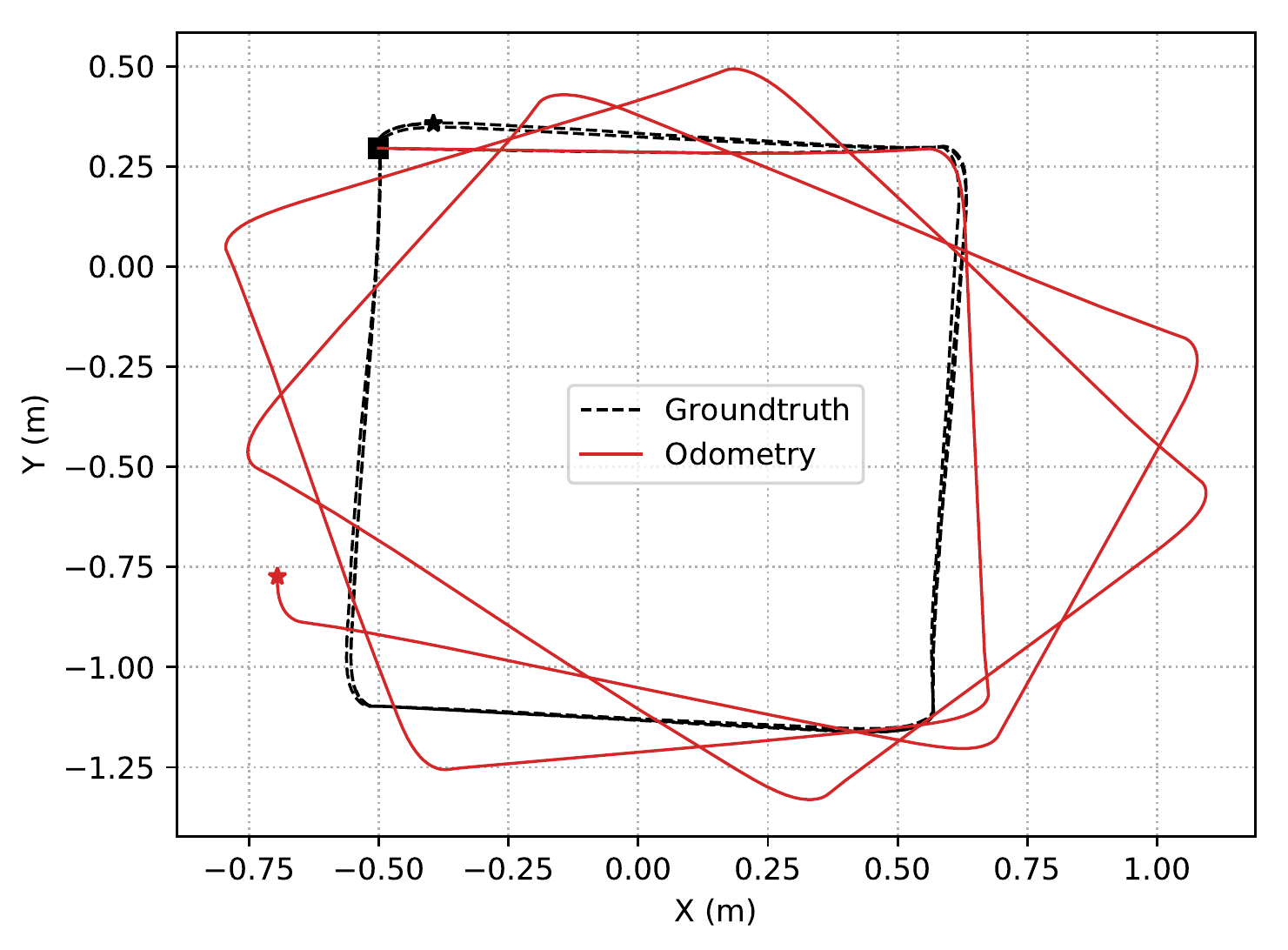}
    \caption{E-puck odometry.}
    \label{fig:epuckodom}
  \end{subfigure}
  \qquad
  \begin{subfigure}{.32\textwidth}
    \centering
    \includegraphics[width=0.99\linewidth]{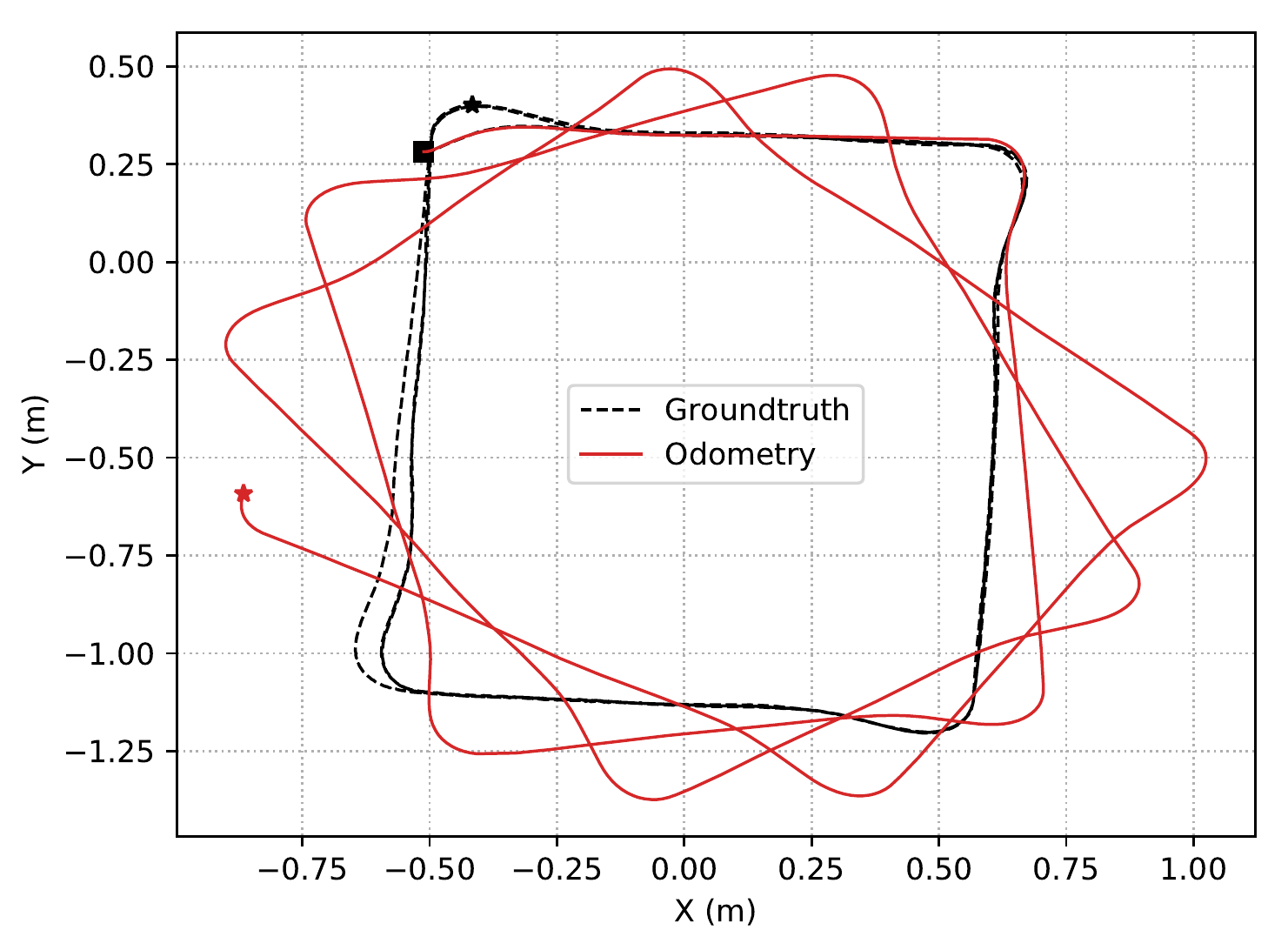}
    \caption{HeRo odometry.}
    \label{fig:heroodom}
  \end{subfigure}
  \qquad
  \begin{subfigure}{.32\textwidth}
    \centering
    \includegraphics[width=0.99\linewidth]{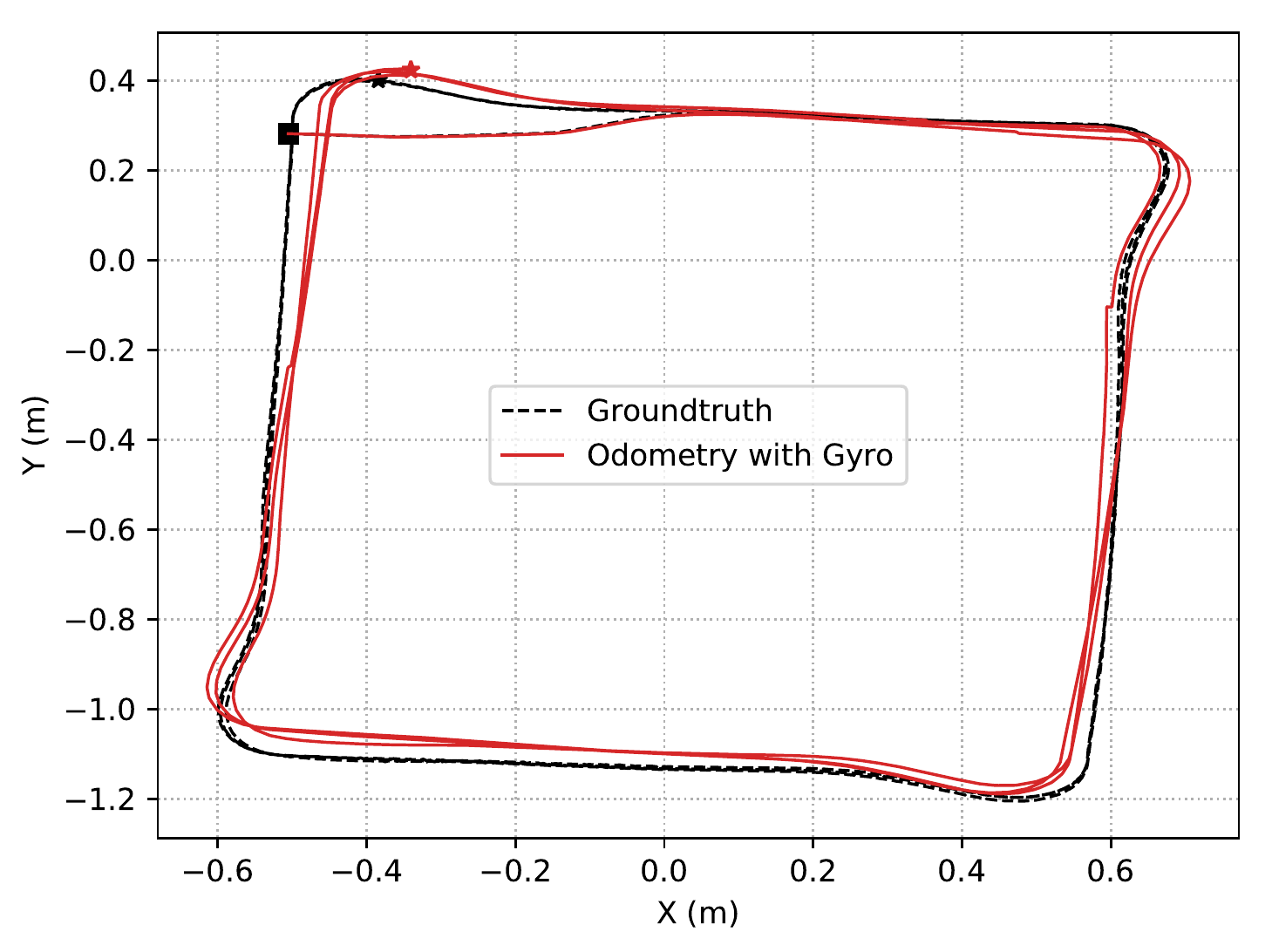}
    \caption{HeRo + IMU odometry.}
    \label{fig:heroimuodom}
  \end{subfigure}
  \caption{Trajectory performed by: (a) E-puck, (b) HeRo and (c) HeRo using e-Hat with inertial sensors.}
  %traveling a true distance of 328.45 cm but estimating 300.00 cm, a trajectory error of 8.66~\%.}
  \label{fig:odometry}
\end{figure*}

Besides comparing the odometry produced by both robots, we extended another HeRo \hec{with an IMU e-Hat.} Combining these two sensors with a Kalman filter improved our robot's orientation estimate and, consequently, improved the robot's odometry. Fig.~\ref{fig:odometry} shows the trajectories performed by (a) an E-puck, (b) a HeRo without e-hat, and (c) a HeRo using e-hat with a gyroscope and accelerometer. A video of this experiment is available on Youtube\footnote{Odometry Comparison: \url{https://youtu.be/9s6Fg20uOpc}}.

As observed, the HeRo's odometry is comparable to the E-puck's. Given that E-puck is one of the most robust and well-used robots for swarm experimentation, we believe that our robot also proves to be an attractive solution. Moreover, the components used by HeRo are highly affordable when compared to E-puck. Furthermore, using the module with inertial sensors improved the robot's orientation, making the localization more robust, allowing its use in other applications.

% Analyses of Infrared Distance Sensor - ok %
\subsection{Distance Sensor}
This experiment assesses the performance of the IR sensor concerning the distance estimation to a white obstacle. Before evaluating the performance of the distance sensor, we need to characterize the sensor (convert the analog signal to distance).

To convert the infrared sensor readings to distance, we first take the sensor readings using a 10-bits ADC input for various object distances, ranging from $0$ to $40$~cm, in one-centimeter intervals. To remove light interferences, we first read the IR sensor without activating the IR emitter and then turn the emitter on and take another reading. The difference between these two readings returns a more robust measurement of the effective light intensity reflected by the obstacle.

\begin{figure}[!t]
 \begin{subfigure}{.5\textwidth}
    \centering
    \includegraphics[width=.68\linewidth]{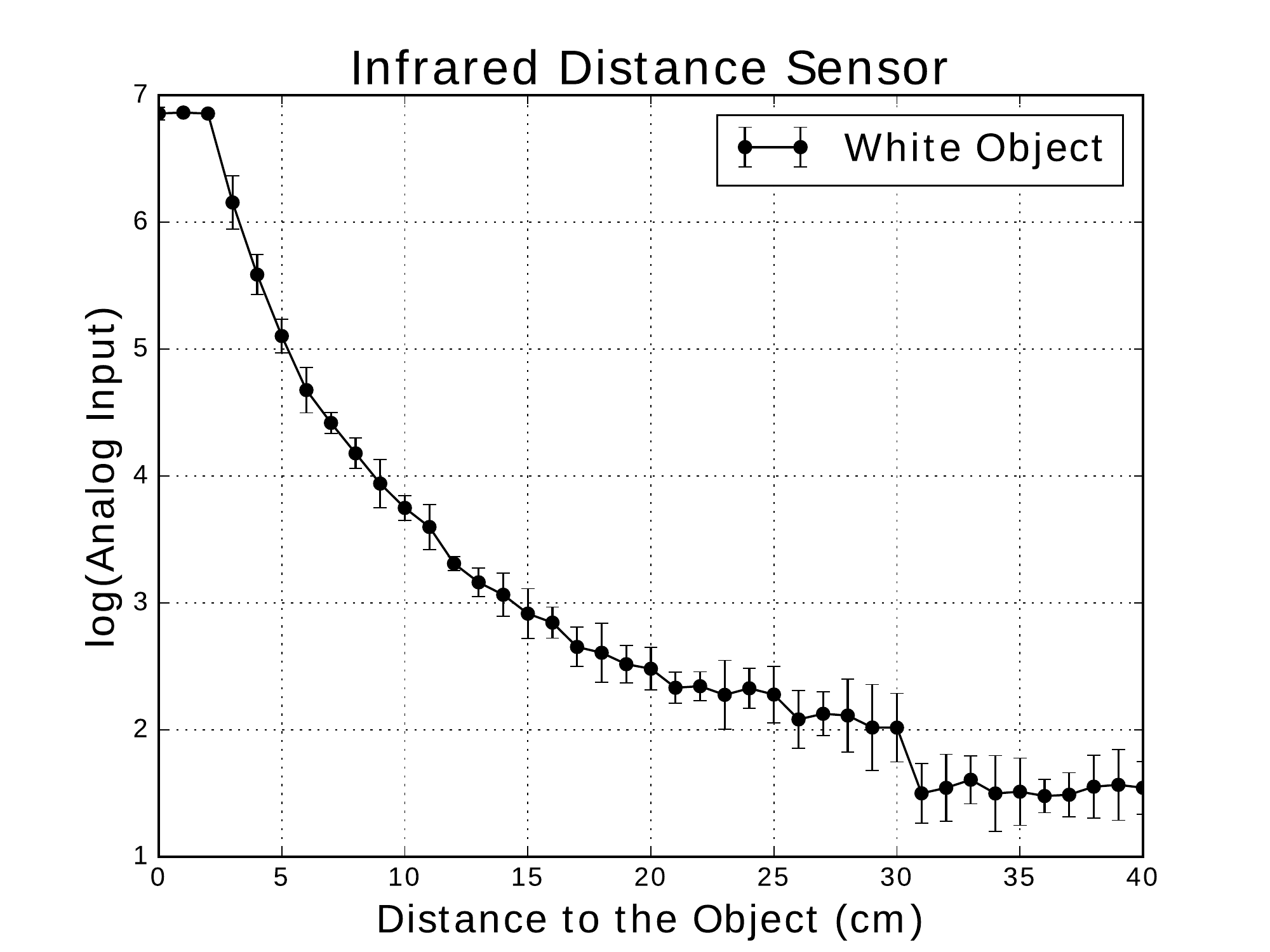}
    \caption{}
    \label{fig:infrared_distance_adc}
 \end{subfigure}
 \begin{subfigure}{.5\textwidth}
    \centering
    \includegraphics[width=.68\linewidth]{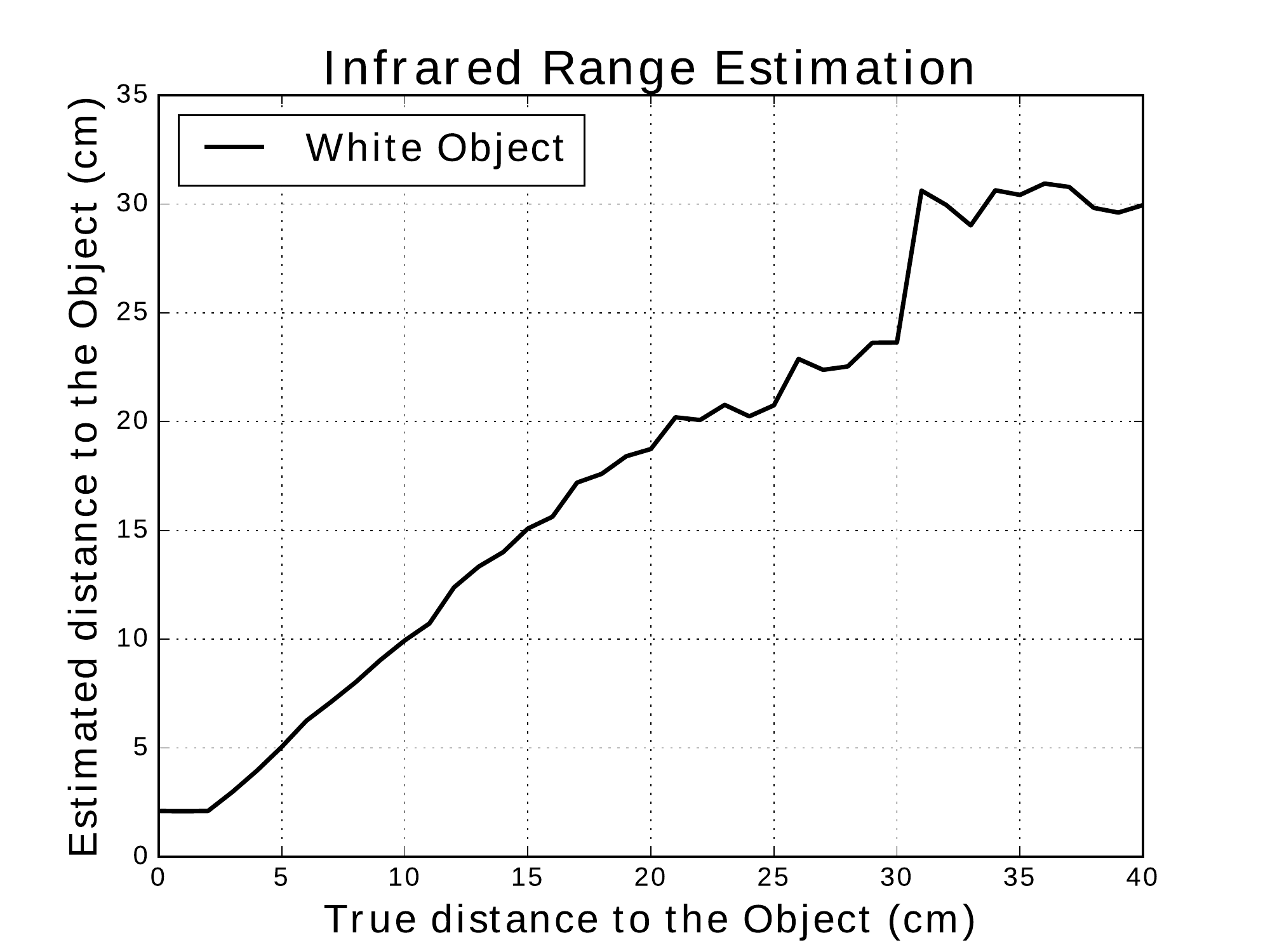}
  \caption{}
  \label{fig:infrared_distance_dist}
 \end{subfigure}
   \caption{Analysis of the infrared distance sensor. (a) Shows the readings obtained from a single IR sensor as a function of the distance to the white target; and (b) shows the estimated distance after calibrating the sensor for a maximum range of $20$~cm.}
  \label{fig:infrared_distance}
\end{figure}

Fig.~\ref{fig:infrared_distance_adc} shows these measurements assuming log-scale for y-axis. As observed, it seems possible to detect objects within the range of $30$~cm, but to better estimate the distance, we decided to limit such a range to $20$~cm. After collecting these measurements, we perform the distance sensor calibration solving the Equation~\ref{eq:infrared}. Fig.~\ref{fig:infrared_distance_dist} shows the distance estimates for the object after the calibration process.

%% Communication
\subsection{Communication}
%One of the essential capabilities a swarm robot requires is the scalability of the communication architecture. 
Communication mechanisms is swarm robots must be scalable to accommodate a large number of robots.  
This experiment evaluates the communication scalability regarding the bandwidth (the maximum amount of data that can travel through a channel) when using ROS to program the robots. We estimate the number of robots supported in the network by computing the total bandwidth used by one robot and the maximum bandwidth supported in the network.

Table~\ref{tab:communication_bandwidth} shows the measured bandwidth of each topic (communication channel) between the robot and a server executing ROS. We assume that these topics are publishing or subscribing to messages at $20$~Hz, the default frequency rate of HeRo processing. We captured these measurements using the \textit{rostopic tool} which provides the packet size of a single message considering: an overhead of $20$~Bytes for the TCP packet (for the Wi-Fi data connection) and $8$~Bytes for \textit{rosserial} serialization; and the message data size, depending on each type of topic. In addition to packet size, the \textit{rostopic tool} also provides the actual bandwidth for each topic, allowing us to compute the total bandwidth used by one robot ($24$~KBps), assuming communication at $20$~Hz.

Assuming the Wi-Fi module used in our robot can handle at least $1$~MBps\footnote{Datasheet: \url{www.espressif.com/sites/default/files/documentation/0a-esp8266ex\_datasheet\_en.pdf}}, which is the same to communicate all these topics at $800$~Hz meaning that we are only using $2.5$\% of the maximal capacity. Moreover, the robot connects to a consumer-grade wireless network route in infrastructure mode that provides a maximum bandwidth of $150$~Mbps (or $18$~MBps). Considering that one HeRo uses only $23.44$~KBps to communicate at $20$~Hz, theoretically, we estimate that almost $820$ robots are supported in this network. Despite typical Dynamic Host Configuration Protocol (DHCP) can not address all these IPs, one may use other ways to avoid this limitation, such as subnetwork or using multiple routers.

\begin{table}[t]
  \centering
  \caption{Maximum amount of data that can travel through a ROS topic. These topics are operating at $20$~Hz, which is the default frequency rate of HeRo.}
  \label{tab:communication_bandwidth}
  \scalebox{0.42}{ \resizebox{\textwidth}{!}{
  \begin{tabular}{lll}
    \toprule
    \textbf{ROS Topics} & \textbf{Packet Size (KB)} & \textbf{Bandwidth (KBps)} \\
    \midrule
    /imu & 0.320 & 5.45 \\
    /laser & 0.130 & 2.53 \\
    /odom & 0.730 & 14.18 \\
    /led & 0.016 & 0.32 \\
    /cmd\_vel & 0.048 & 0.96 \\
    \midrule
    \textbf{Total} &  & \textbf{23.44 KBps} \\
    \bottomrule
\end{tabular}
}}
\end{table}

% Power Consumption
\subsection{Power Consumption}
Another critical concern is the robot's power autonomy, which defines its operating time. This experiment analyzes the power consumption of the components, establishing the power autonomy of our robot. To better understand the consumption of the robot, we measure the current (mA) used by the robot in three typical situations: \hec{(i) when sensors and communication are active, (ii) with the indicator LEDs turned on, and (iii) with the motors active.} Table~\ref{tab:power_consumption} shows the consumption (in mA) of the robot for these combinations.

For the first case, we observe the effect of the frequency of publication in the robot's power consumption. Thus, we measured the current for three different frequency rates ($5$, $20$, and $40$~Hz) and noticed that these rates have a minimal impact on consumption. For the second case, we kept the communication frequency at $20$~Hz, and turned on the two indicator LEDs, and changed its white light intensity from half to full. We noticed that the light intensity used by the LEDs significantly impacts the consumption (almost $50$~mA). In the last case, we kept the communication frequency rate and turned on the two motors. To observe the impact of the robot's velocity on the power consumption, we measured the current for two different velocities. As expected, the motors are the most power-consumer components in the robot, and their velocity proportionally affects the power consumption.

In order to analyze the power consumption in a general way, we assumed a typical use for these three combinations, in which we retained the sensors and communication at $20$~Hz; indicator LEDs with half intensity; and motors reaching a speed of $10$~cm/s. Thus, we observed the consumption reaching $550$~mA with a slight deviation of $25$~mA. Suppling the robot with a $3.7$~V $1800$~mAh Li-Po battery, we estimate the minimum and maximum autonomy of $3$~h and $9$~h, respectively.
% {
% \newcommand*\rot[1]{\hbox to1em{\hss\rotatebox[origin=br]{-50}{#1}}}
% \begin{table}[htbp]
% \centering
% \caption{Power consumption of HeRo considering a voltage of 3.7 V. Typical power consumption is the average of 30 samples.}
% \label{tab:power_consumption}
% \resizebox{0.7\columnwidth}{!}{
% \begin{tabular}{lllllll}
% \toprule
% \multicolumn{1}{c}{\rot{\textbf{Sensing (Hz)}}} & \multicolumn{1}{c}{\rot{\textbf{Communication (Hz)}}} & \multicolumn{1}{c}{\rot{\textbf{LEDs brightness (\%)}}} & \multicolumn{1}{c}{\rot{\textbf{Motors (cm/s)}}} & \rot{\textbf{Min (mA)}} & \rot{\textbf{Typical (mA)}} & \rot{\textbf{Max (mA)}} \\ \midrule
% \multicolumn{1}{c}{5} & \multicolumn{1}{c}{5} & \multicolumn{1}{c}{0} & \multicolumn{1}{c}{0} & 152 & 161 \textpm 9 & 180 \\
% 20 & 20 & 0 & 0 & 153 & 175 \textpm 16 & 205 \\
% 40 & 40 & 0 & 0 & 170 & 183 \textpm 9 & 205 \\
% 20 & 20 & 50 & 0 & 187 & 205 \textpm 14 & 245 \\
% 20 & 20 & 100 & 0 & 225 & 247 \textpm 2 & 292 \\
% 20 & 20 & 0 & 10 & 455 & 512 \textpm 30 & 584 \\
% 20 & 20 & 0 & 25 & 613 & 660 \textpm 25 & 717 \\ \midrule
% 20 & 20 & 50 & 10 & 396 & 550 \textpm 47 & 628 \\ \bottomrule
% \end{tabular}
% }
% \end{table}
% }

\begin{table}[htbp]
\centering
\caption{Power consumption of HeRo considering a voltage of 3.7 V. Typical power consumption is the average of 30 samples.}
\label{tab:power_consumption}
\resizebox{0.96\columnwidth}{!}{
\begin{tabular}{l@{\hskip -1in}ccc}
\toprule
\textbf{Mode} & \textbf{Min} & \textbf{Typical (mA)} & \textbf{Max} \\
\midrule
Sensing \& Communication at 5 Hz & 152 & 161 \textpm 9\ \ & 180 \\
Sensing \& Communication at 20 Hz$^{1}$ & 153 & 175 \textpm 16 & 205 \\
Sensing \& Communication at 40 Hz & 170 & 183 \textpm 9\ \ & 205 \\
& & & \\
Sensing \& Communication$^{1}$ \& LEDs (50\%)$^{2}$  & 187 & 205 \textpm 14 & 245 \\
Sensing \& Communication$^{1}$ \& LEDs (100\%) & 225 & 247 \textpm 2\ \ & 292  \\
& & & \\
Sensing \& Communication$^{1}$ \& Motors (10 cm/s)$^{3}$ & 455 & 512 \textpm 30 & 584 \\
Sensing \& Communication$^{1}$ \& Motors (25 cm/s) & 613 & 660 \textpm 25 & 717 \\
& & & \\
\midrule
Typical Use$^{123}$ & 396 & 550 \textpm 47 & 628\\
\bottomrule \\
$^{1}$ Typical frequency rate used for sensing and communication. \\
$^{2}$ Common brightness used in the LEDs indicators (White color). \\
$^{3}$ Common velocity performed by the robot during the experiments.
\end{tabular}}
\end{table}

% Odometry comparison
% IR sensor
% Power Consumption
% Communication (scalability)

\section{Applications}
\label{sec:casestudy}
% Application
In addition to evaluating the robot's performance, we also demonstrate its capability executing a set of different applications. 

\subsection{Mapping}
\begin{figure}[!t]
      \centering
      \includegraphics[width=0.80\linewidth]{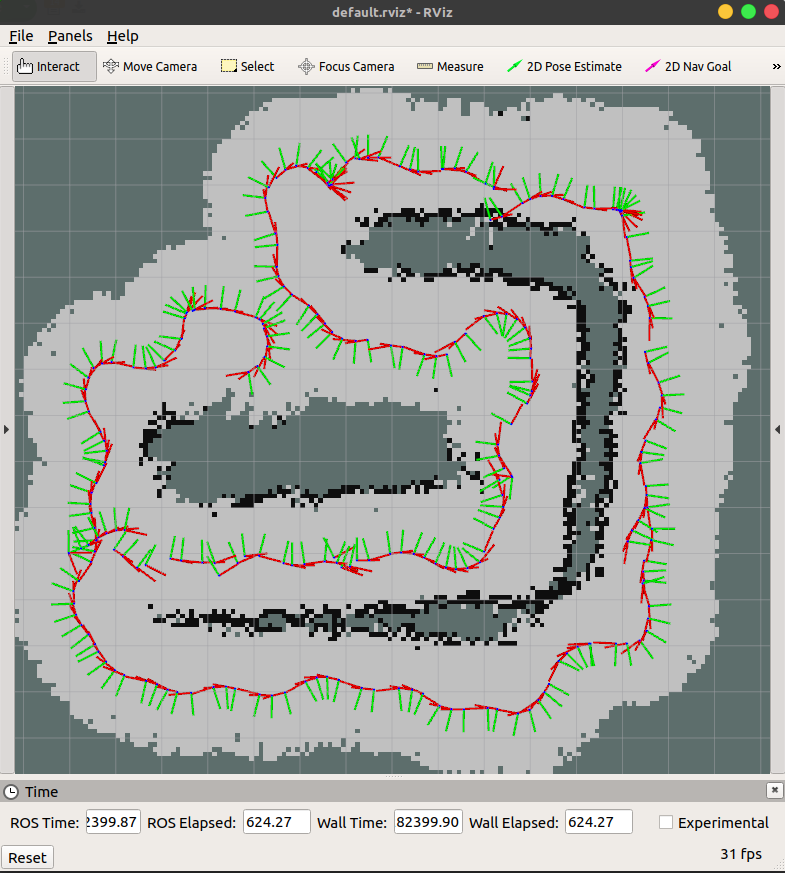}
      \caption{A map produced by the robot using only eight IR sensors and odometry.}
      \label{fig:herogmapping}
\end{figure}

This experiment shows the capacity of a single real robot to perform a mapping task. We set up an environment composed of white cardboards forming a scene similar to a hall contained in an area of $1.20\times1.20$~m. Using the odometry computed by the robot, and configuring the ROS Gmapping\footnote{GMapping: \url{http://wiki.ros.org/gmapping}} package, we get some interesting results. Fig.~\ref{fig:herogmapping} shows the map produced by the robot. \rez{The sequence of axes (x-red, y-green, z-blue coming out of the figure) projected onto the map represents the temporal pose of the robot computed from the robot's odometry.}
%The axes projected onto the map are the robot's odometry. 
A video showing the execution of this experiment is available on Youtube\footnote{Mapping Performance: \url{https://youtu.be/pD-tutQLuDo}}.

% Random Walk
\subsection{Decentralized coverage} 
\begin{figure*}[t]
    \centering
    % trim={<left> <lower> <right> <upper>}
    
	\begin{subfigure}{.280\textwidth}
		\centering
		\includegraphics[trim={10cm 5cm 7cm 2cm},clip,width=\linewidth]{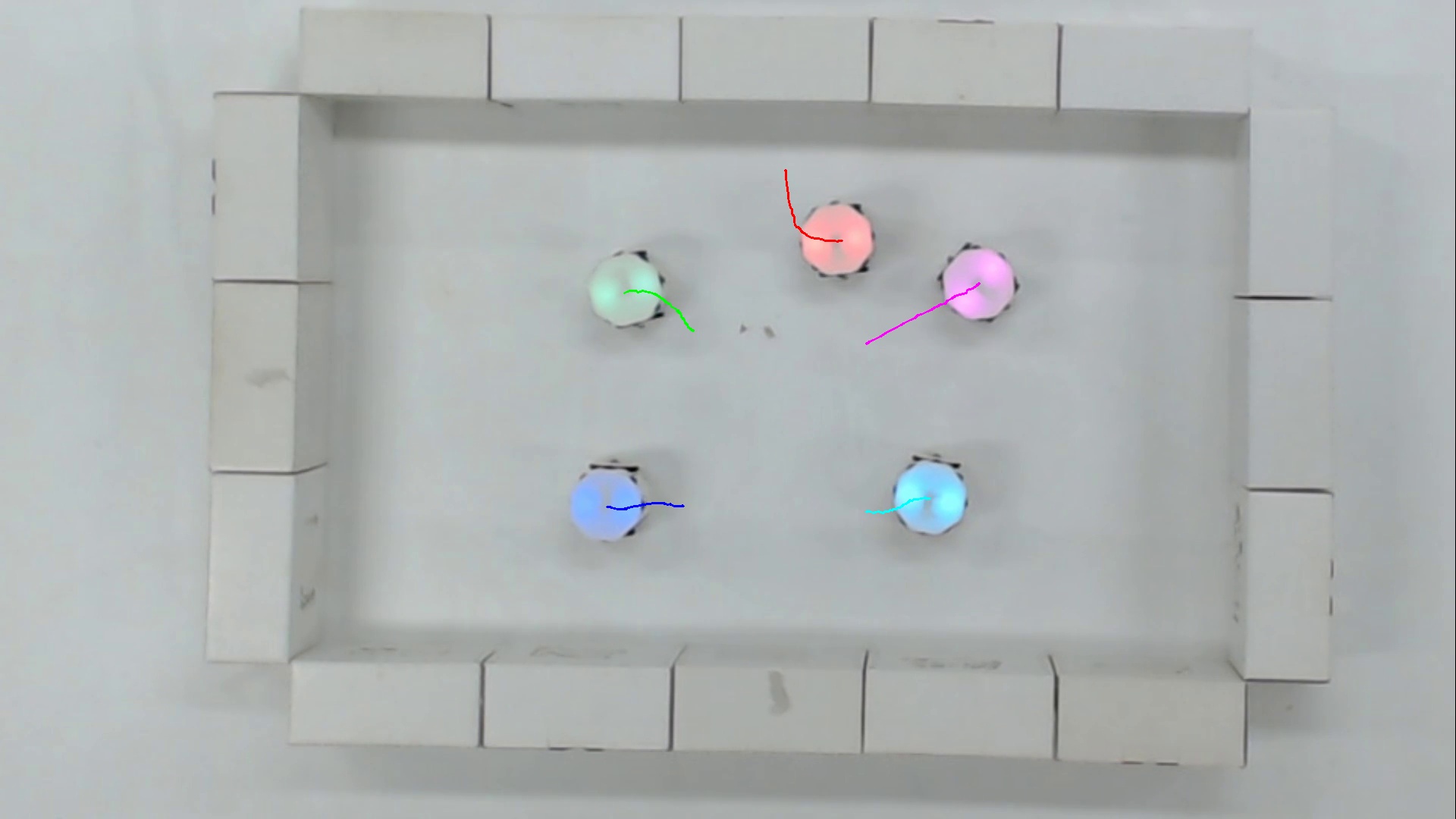}
		\caption{$t=2$~s.}
	\end{subfigure}%
	\hfil
% 	\hspace*{-1.2em}
	\begin{subfigure}{.280\textwidth}
		\centering
		\includegraphics[trim={10cm 5cm 7cm 2cm},clip,width=\linewidth]{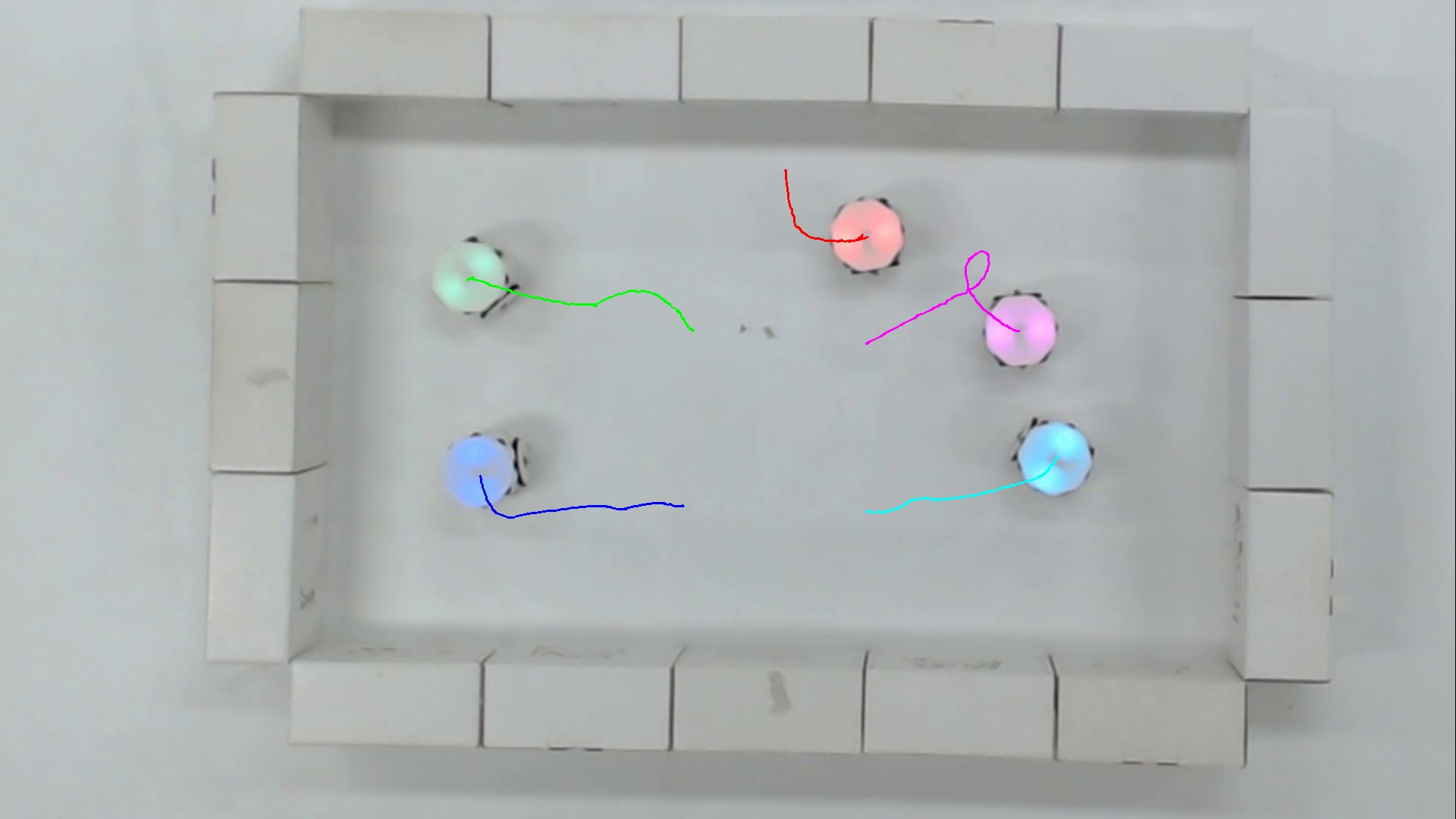}
		\caption{$t=4$~s.}
	\end{subfigure}
	\hfil
% 	\hspace*{-1.6em}
	\begin{subfigure}{.280\textwidth}
		\centering
		\includegraphics[trim={10cm 5cm 7cm 2cm},clip,width=\linewidth]{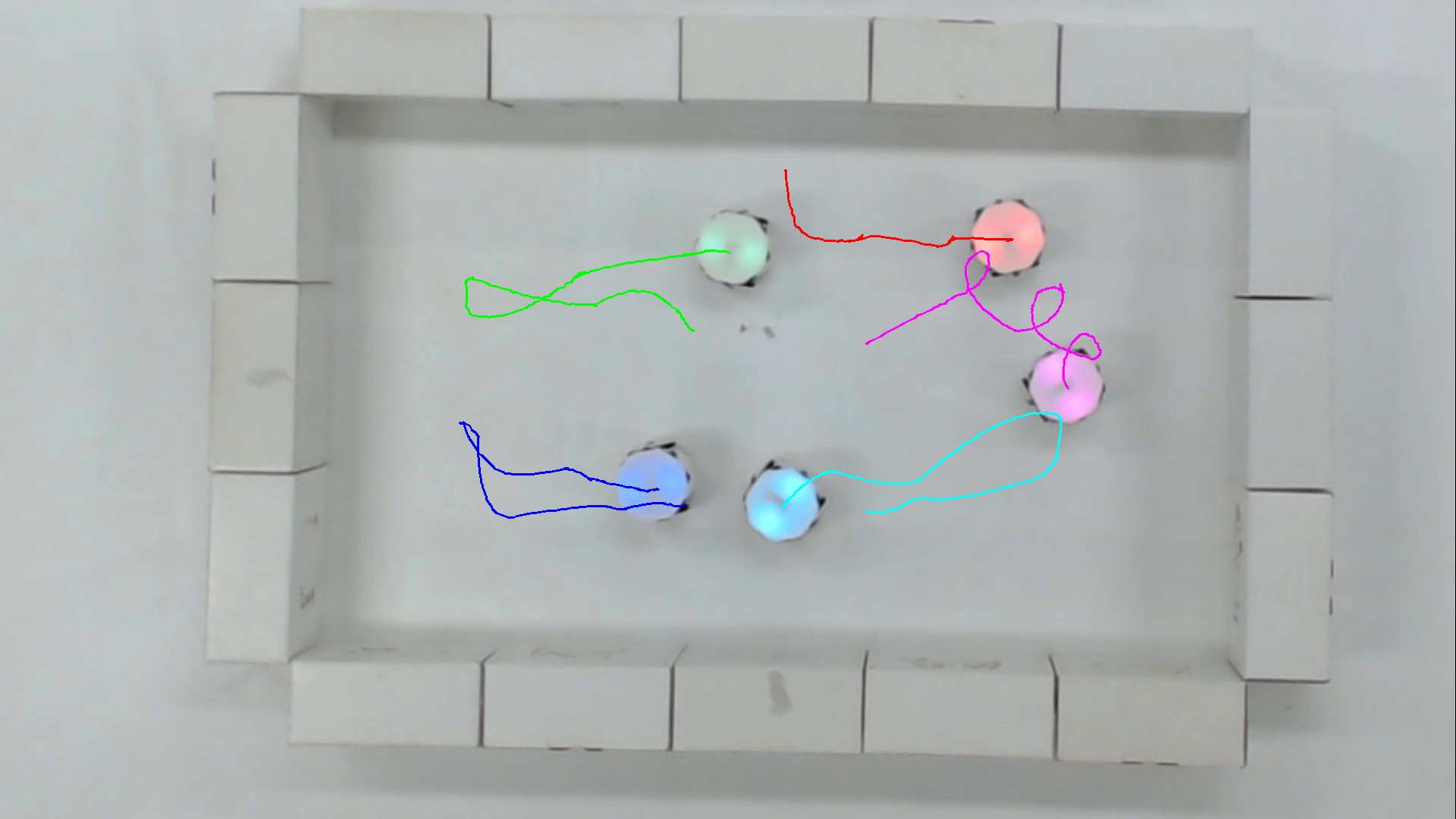}
		\caption{$t=10$~s.}
	\end{subfigure}
	\medskip
% 	\hspace*{-1.6em}
	\begin{subfigure}{.280\textwidth}
		\centering
		\includegraphics[trim={10cm 5cm 7cm 2cm},clip,width=\linewidth]{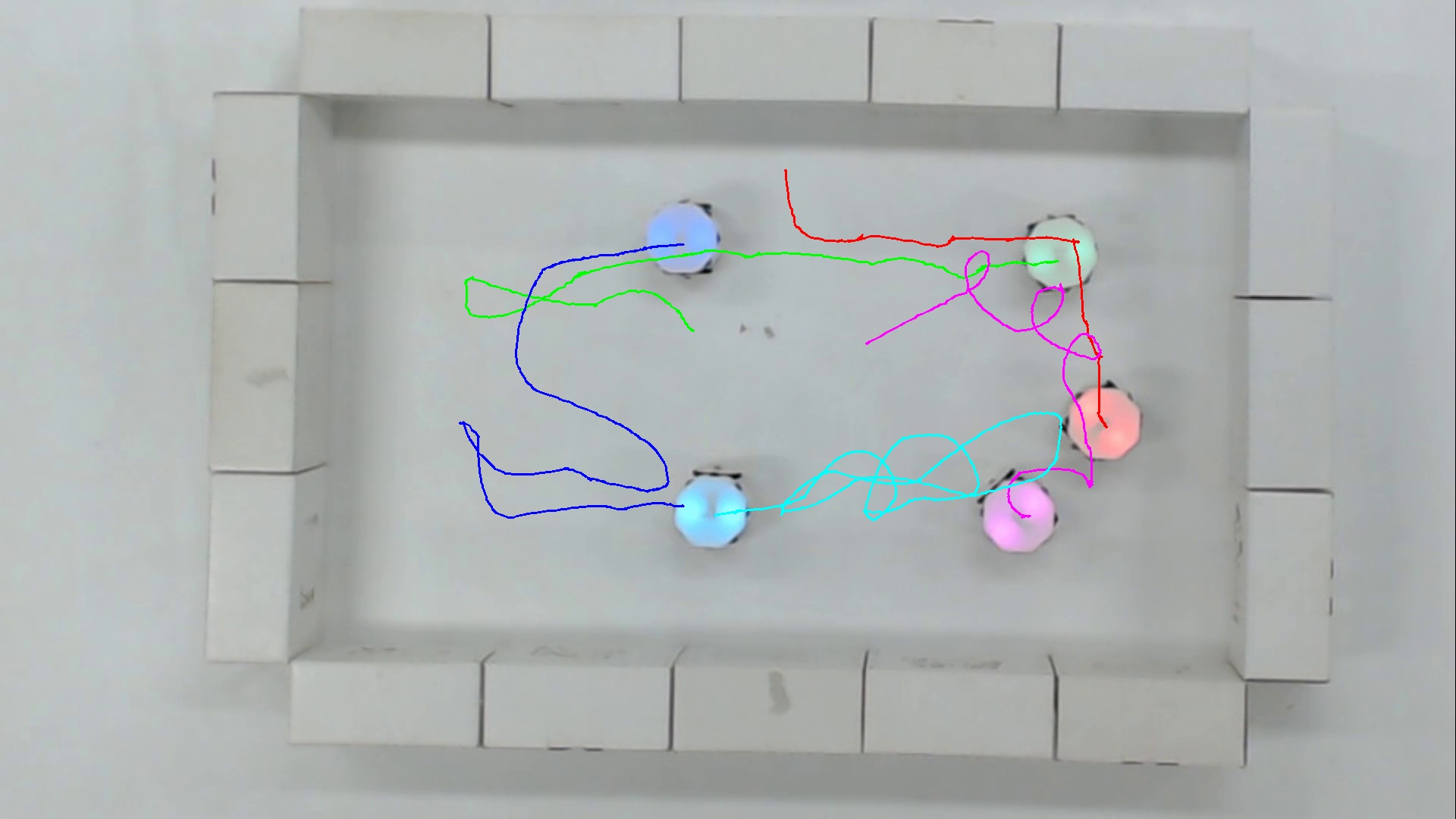}
		\caption{$t=20$~s.}
	\end{subfigure}
	\hfil
% 	\hspace*{-1.6em}
	\begin{subfigure}{.280\textwidth}
		\centering
		\includegraphics[trim={10cm 5cm 7cm 2cm},clip,width=\linewidth]{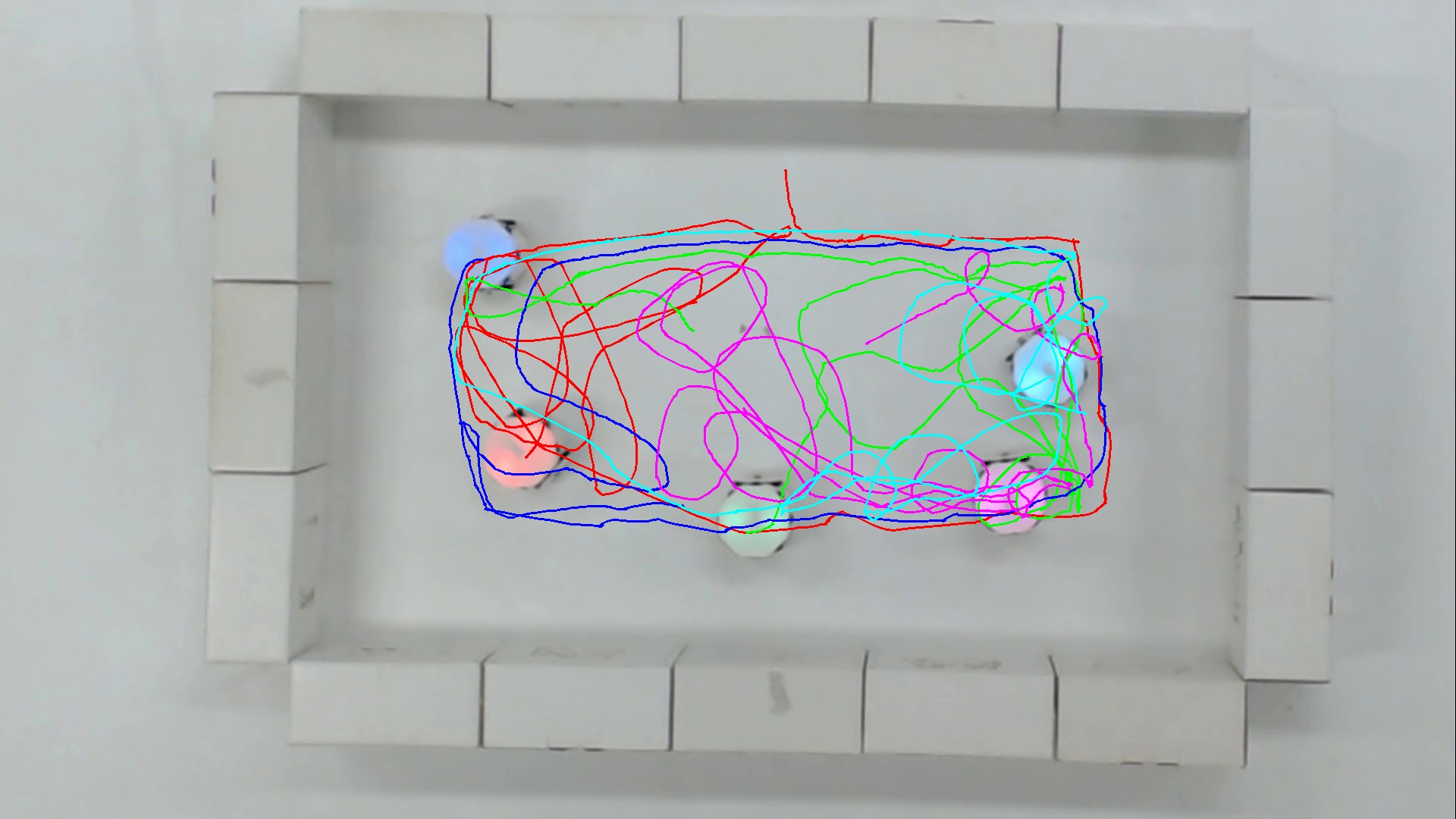}
		\caption{$t=80$~s.}
	\end{subfigure}
	\hfil
% 	\hspace*{-1.6em}
	\begin{subfigure}{.280\textwidth}
		\centering
		\includegraphics[trim={10cm 5cm 7cm 2cm},clip,width=\linewidth]{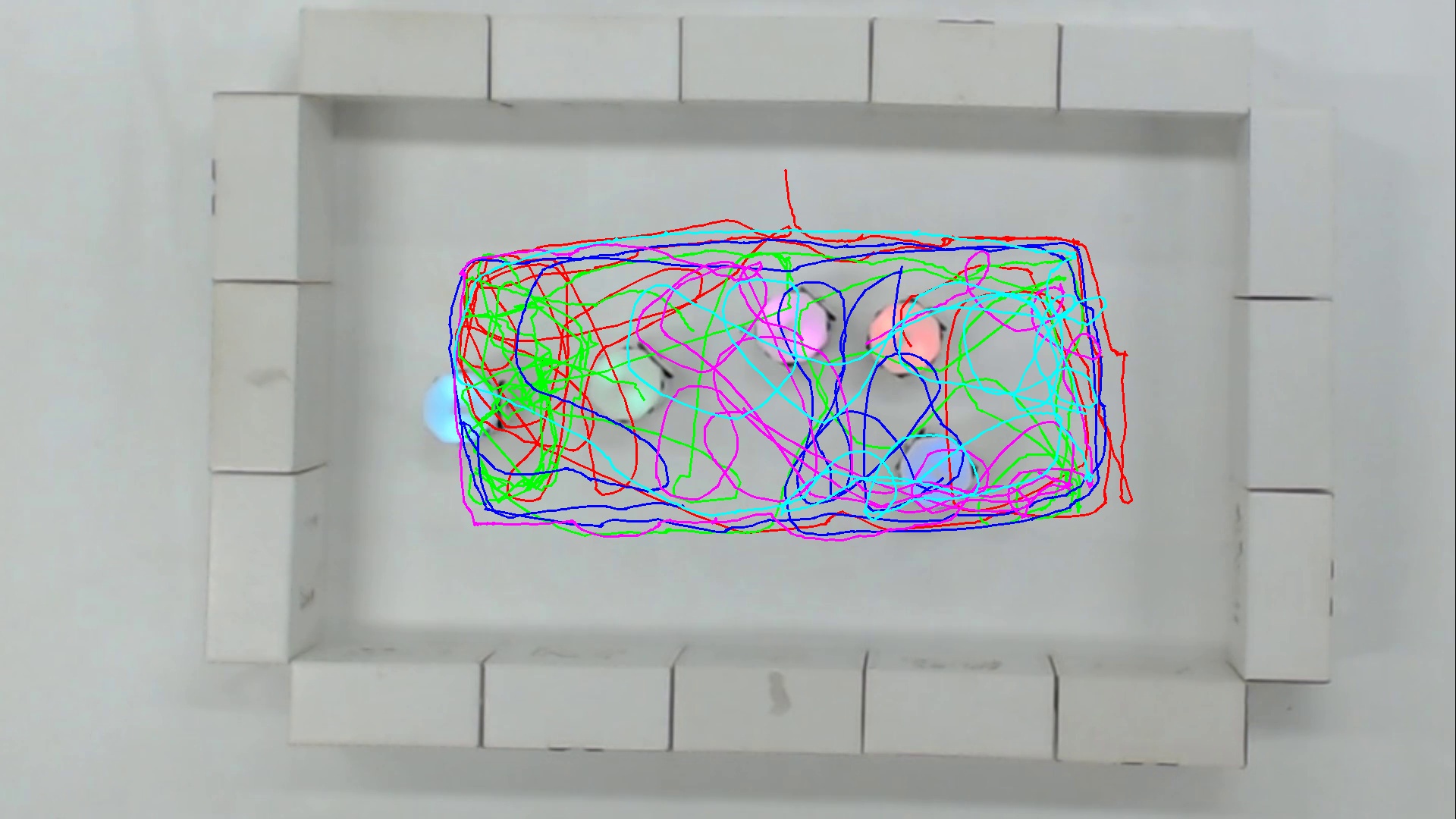}
		\caption{$t=120$~s.}
	\end{subfigure}
	\caption{Snapshots of an experiment showing five HeRo robots performing decentralized coverage.}
	\label{fig:deszentralizedcoverage}
\end{figure*}

In this experiment, five robots perform a covering task in a $0.8\times1.20$ m bounded environment. The coverage method implemented consists of randomly navigating the robots through the environment and avoiding obstacles and collisions with other robots using only local sensing. Fig.~\ref{fig:deszentralizedcoverage} shows a sequence of images captured by an overhead camera. Each of the five robots is programmed to emit a different color, making them easier to see and identify. The lines in the images represent the path performed by each robot. A video of this experiment is available on Youtube\footnote{Decentralized Coverage: \url{https://youtu.be/KmQXBcXKBtE}}.

\subsection{Flocking Behavior}
\begin{figure*}[t]
    \centering
    % trim={<left> <lower> <right> <upper>}
    
	\begin{subfigure}{.280\textwidth}
		\centering
		\includegraphics[trim={9cm 0cm 8cm 2cm},clip,width=\linewidth]{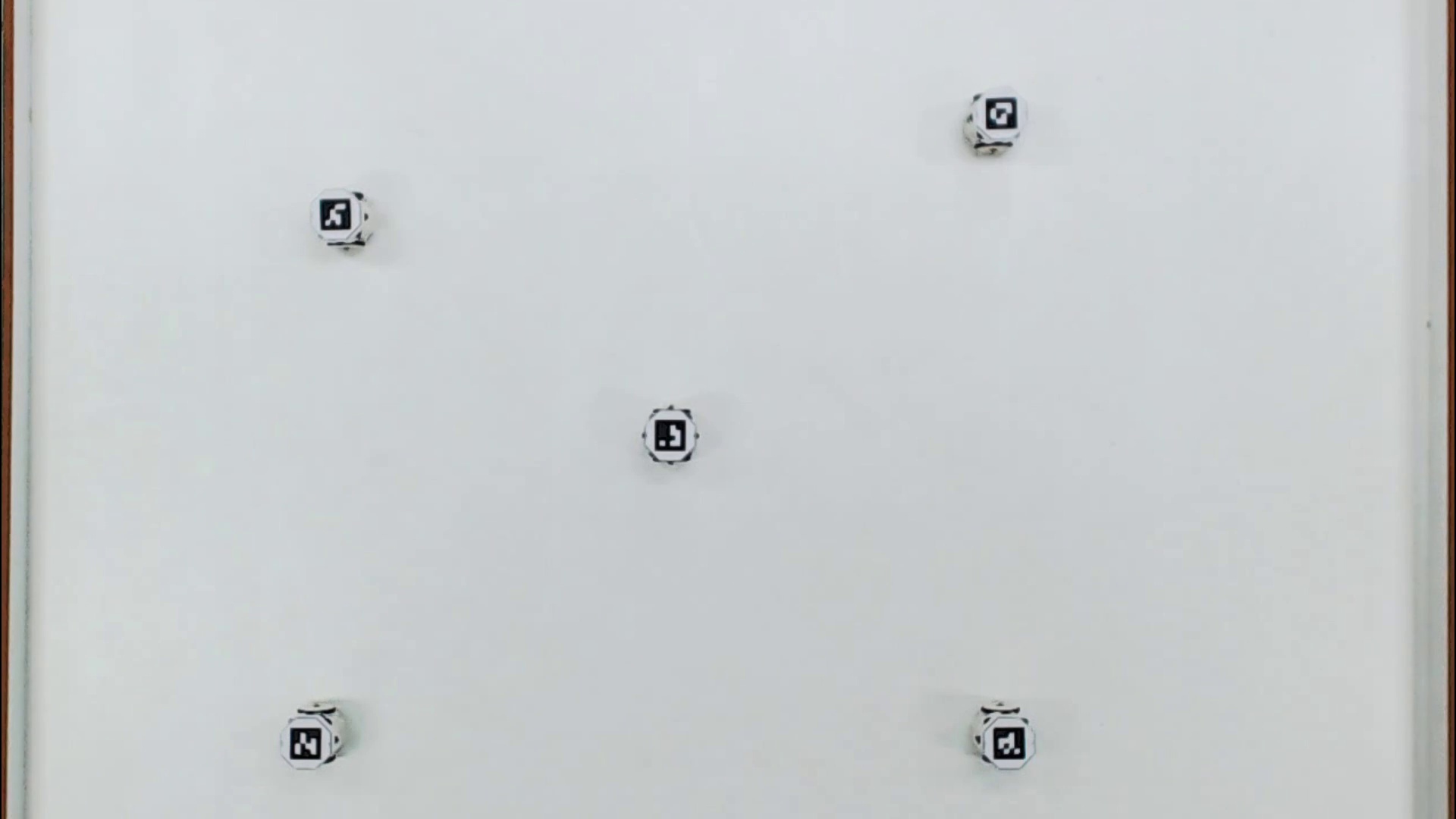}
		\caption{$t=0$~s.}
	\end{subfigure}%
	\hfil
% 	\hspace*{-1.2em}
	\begin{subfigure}{.280\textwidth}
		\centering
		\includegraphics[trim={9cm 0cm 8cm 2cm},clip,width=\linewidth]{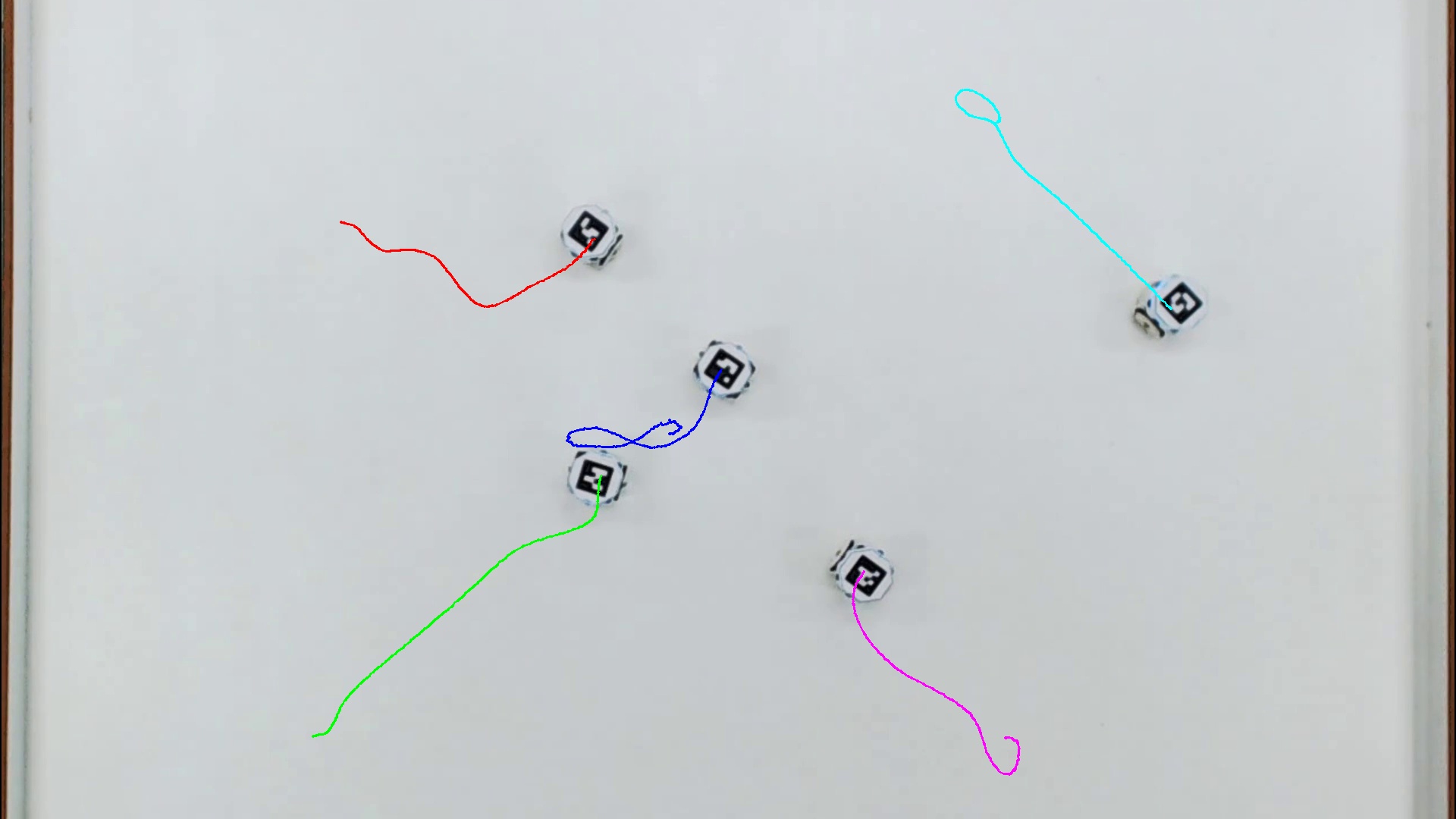}
		\caption{$t=5$~s.}
	\end{subfigure}
	\hfil
% 	\hspace*{-1.6em}
	\begin{subfigure}{.280\textwidth}
		\centering
		\includegraphics[trim={9cm 0cm 8cm 2cm},clip,width=\linewidth]{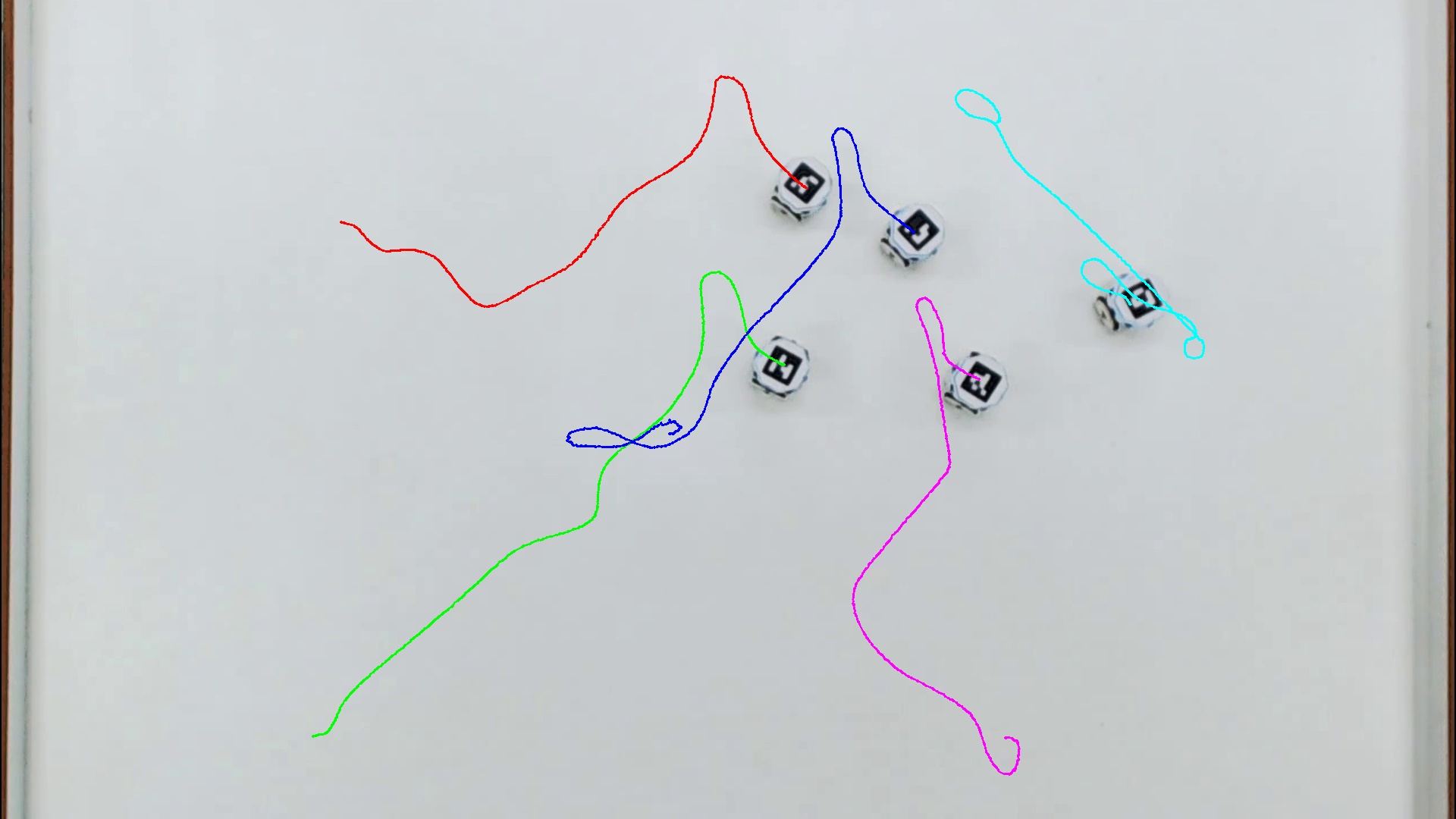}
		\caption{$t=10$~s.}
	\end{subfigure}
	\medskip
% 	\hspace*{-1.6em}
	\begin{subfigure}{.280\textwidth}
		\centering
		\includegraphics[trim={9cm 0cm 8cm 2cm},clip,width=\linewidth]{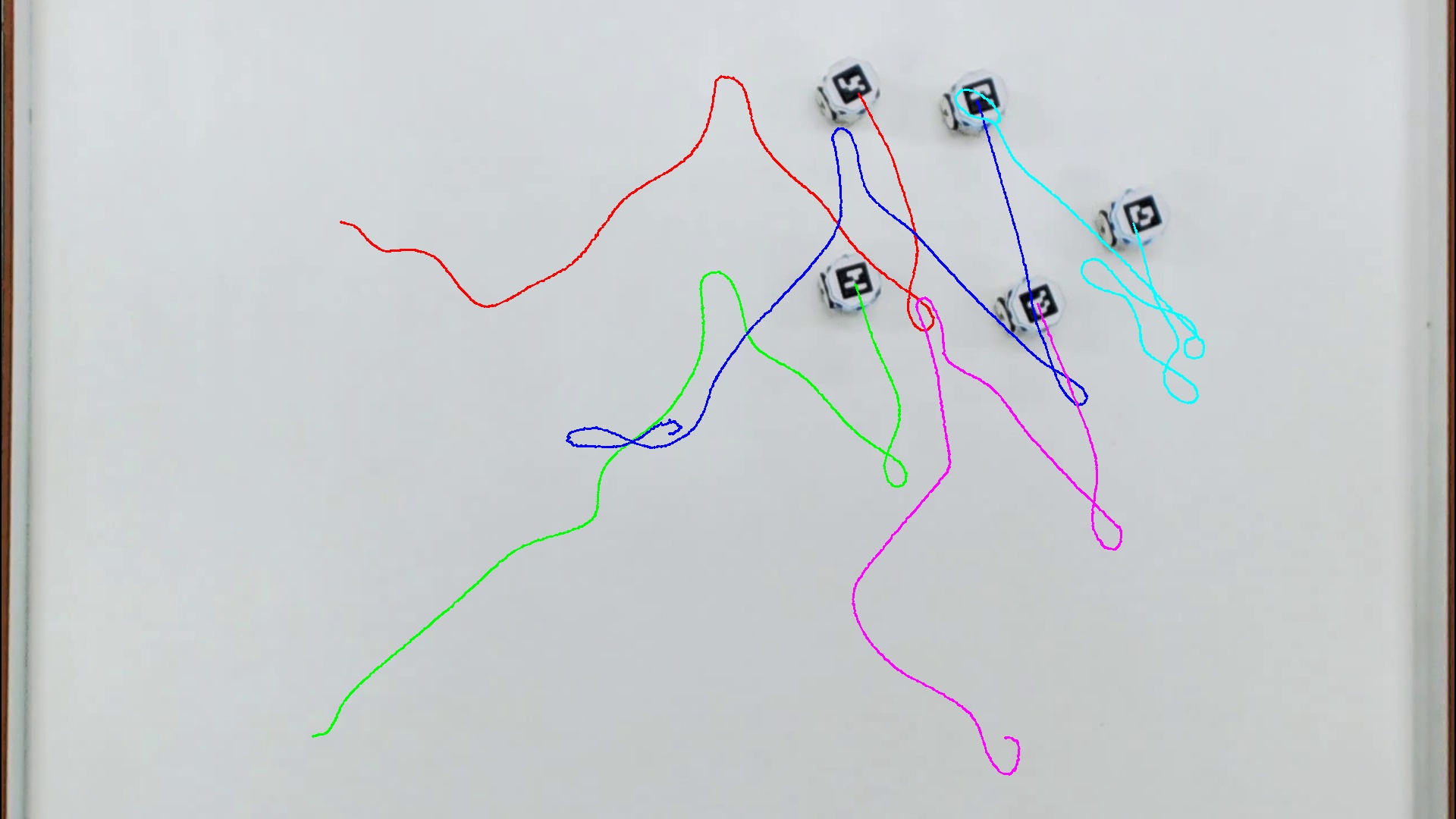}
		\caption{$t=15$~s.}
	\end{subfigure}
	\hfil
% 	\hspace*{-1.6em}
	\begin{subfigure}{.280\textwidth}
		\centering
		\includegraphics[trim={9cm 0cm 8cm 2cm},clip,width=\linewidth]{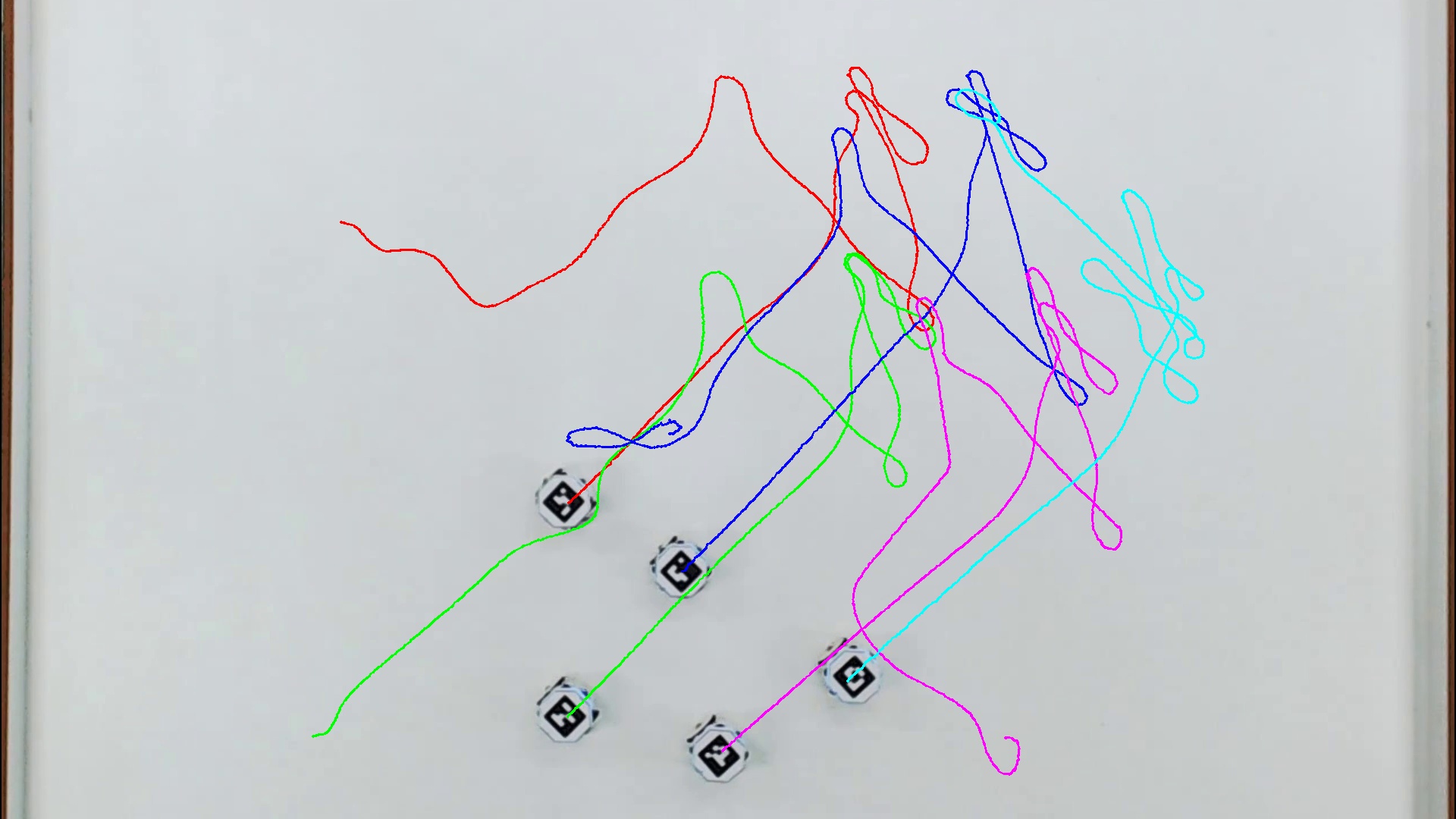}
		\caption{$t=25$~s.}
	\end{subfigure}
	\hfil
% 	\hspace*{-1.6em}
	\begin{subfigure}{.280\textwidth}
		\centering
		\includegraphics[trim={9cm 0cm 8cm 2cm},clip,width=\linewidth]{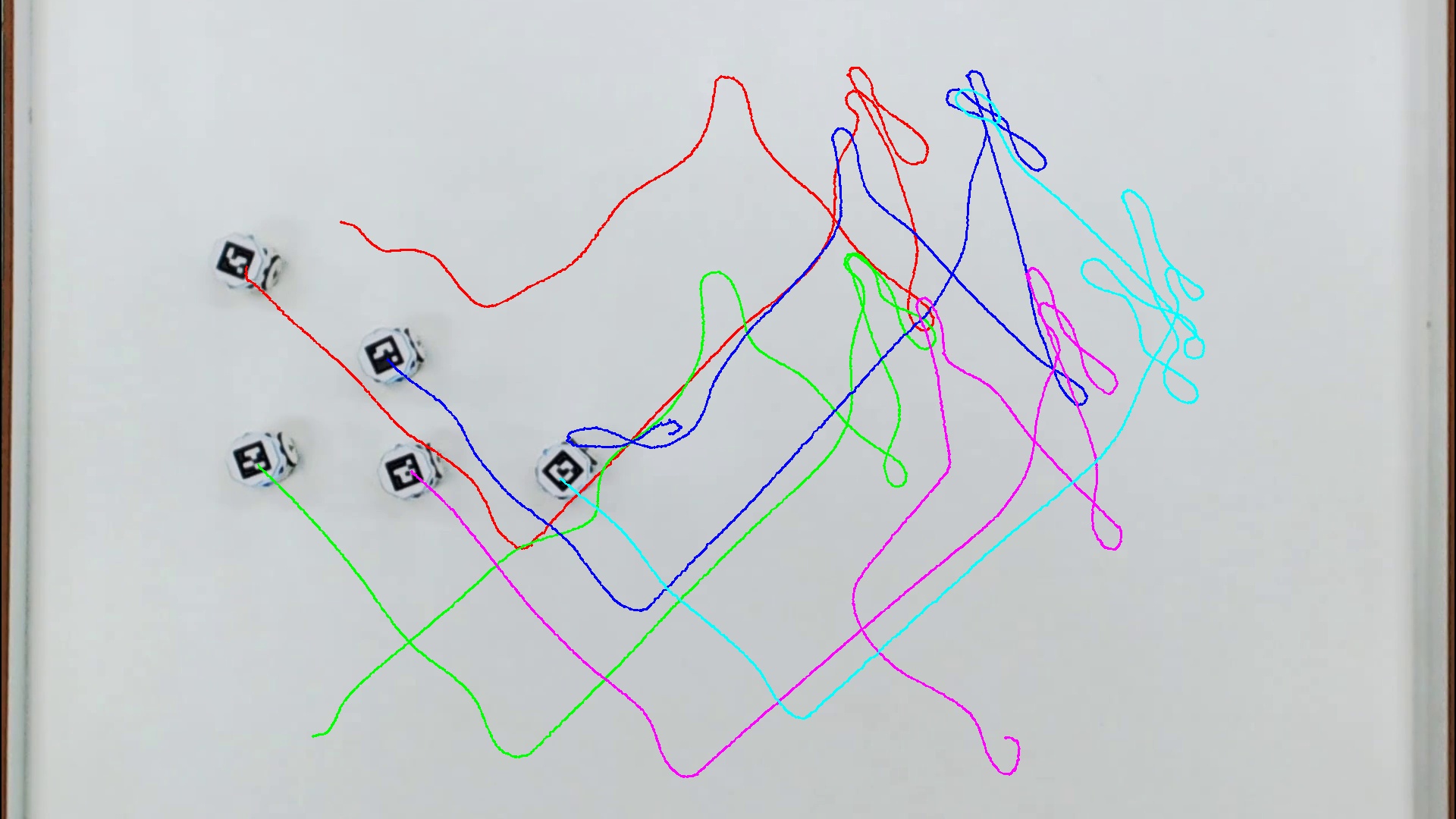}
		\caption{$t=30$~s.}
	\end{subfigure}
	\caption{Snapshots of an experiment showing five HeRo robots performing flocking behavior.}
	\label{fig:flocking}
\end{figure*}

Another experiment demonstrates the robot's capacity to perform a flocking behavior. We used five robots initially randomly distributed in an environment in this experiment. The flocking algorithm implemented in this experiment is decentralized and only requires neighbors' position and relative speed (see~\cite{rezeck2021flocking} for more details). Since we do not yet have a set of sensors to estimate such information, we use the overhead camera and the Apriltag tracker algorithm~\cite{wang2016apriltag,malyuta2018guidance} to locate the robots in the scene. From this information, we could emulate a sensor in the robot that captures its neighbors' position and relative velocity. Fig.~\ref{fig:flocking} shows a sequence of images captured by the overhead camera. The figure shows the initial configuration of the robots. After a runtime, the robots manage themself to aggregate and navigate the environment as a group. A video of this experiment is available on Youtube\footnote{Flocking Behavior: \url{https://youtu.be/u7iioSKtHU8}}.

%%% Cooperative Transportation
\subsection{Cooperative Transportation}
\begin{figure*}[t]
    \centering
    % trim={<left> <lower> <right> <upper>}
    
	\begin{subfigure}{.280\textwidth}
		\centering
		\includegraphics[trim={5cm 2cm 5cm 1cm},clip,width=\linewidth]{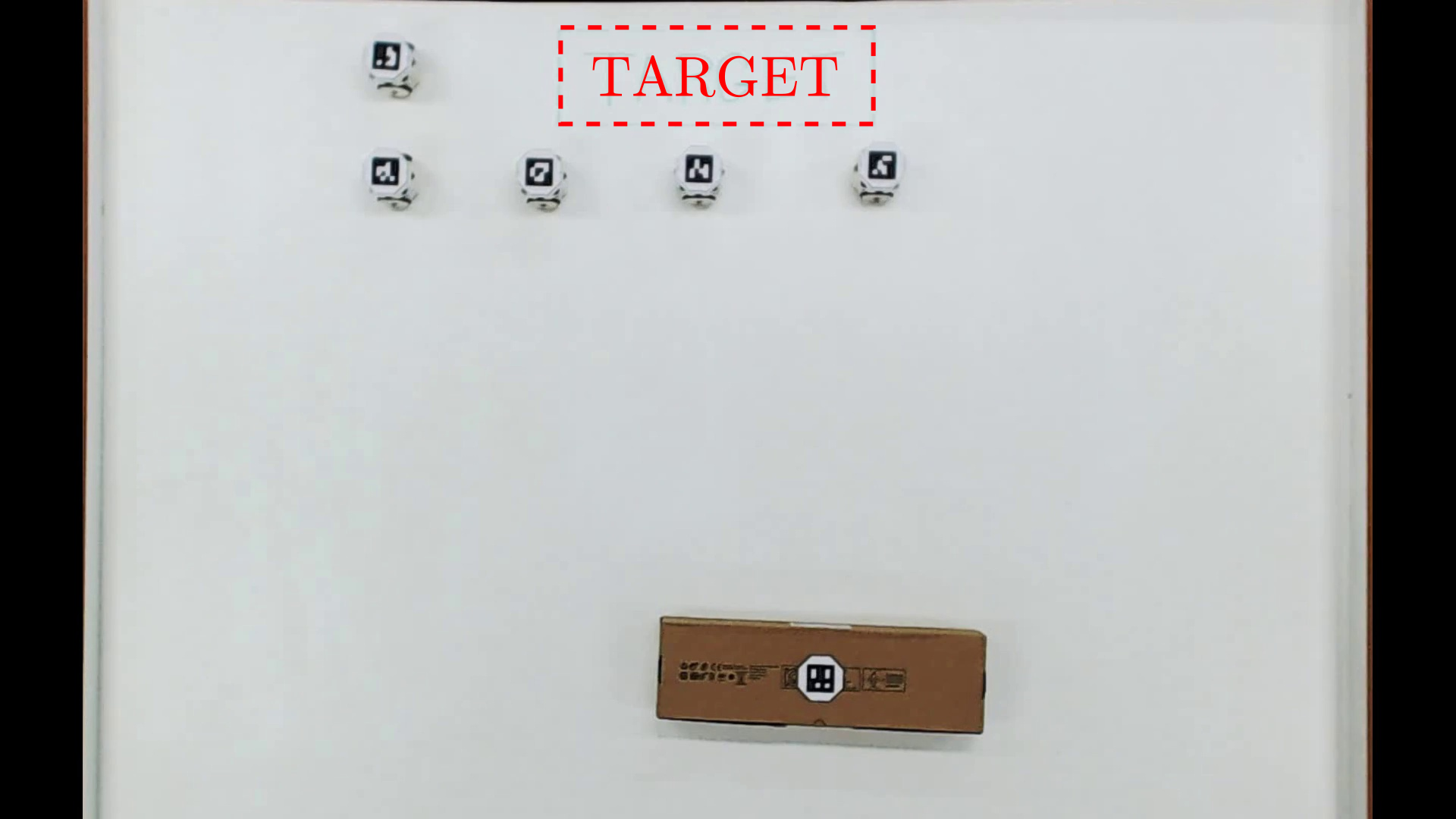}
		\caption{$t=0$~s.}
	\end{subfigure}%
	\hfil
% 	\hspace*{-1.2em}
	\begin{subfigure}{.280\textwidth}
		\centering
		\includegraphics[trim={5cm 2cm 5cm 1cm},clip,width=\linewidth]{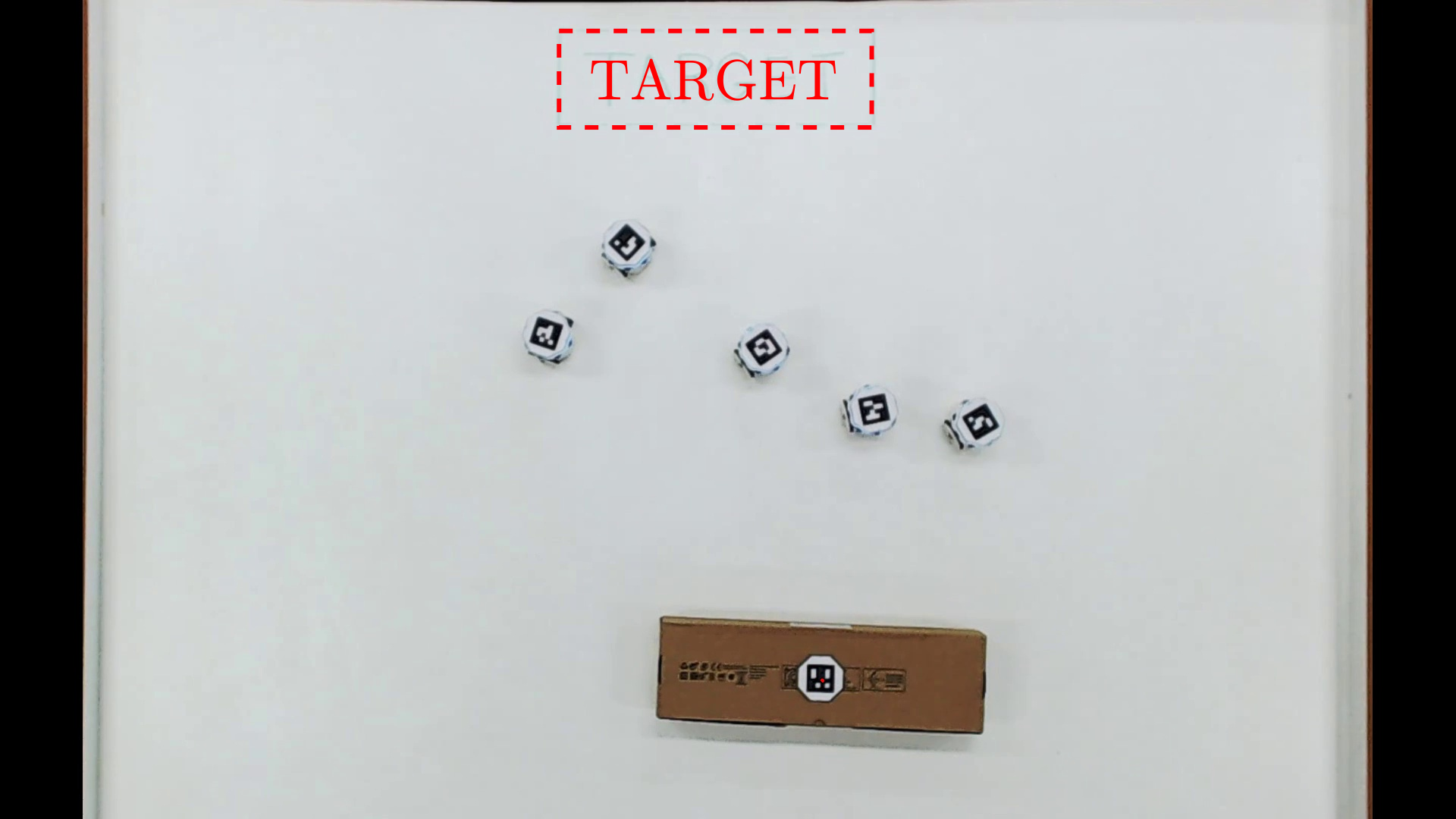}
		\caption{$t=3$~s.}
	\end{subfigure}
	\hfil
% 	\hspace*{-1.6em}
	\begin{subfigure}{.280\textwidth}
		\centering
		\includegraphics[trim={5cm 2cm 5cm 1cm},clip,width=\linewidth]{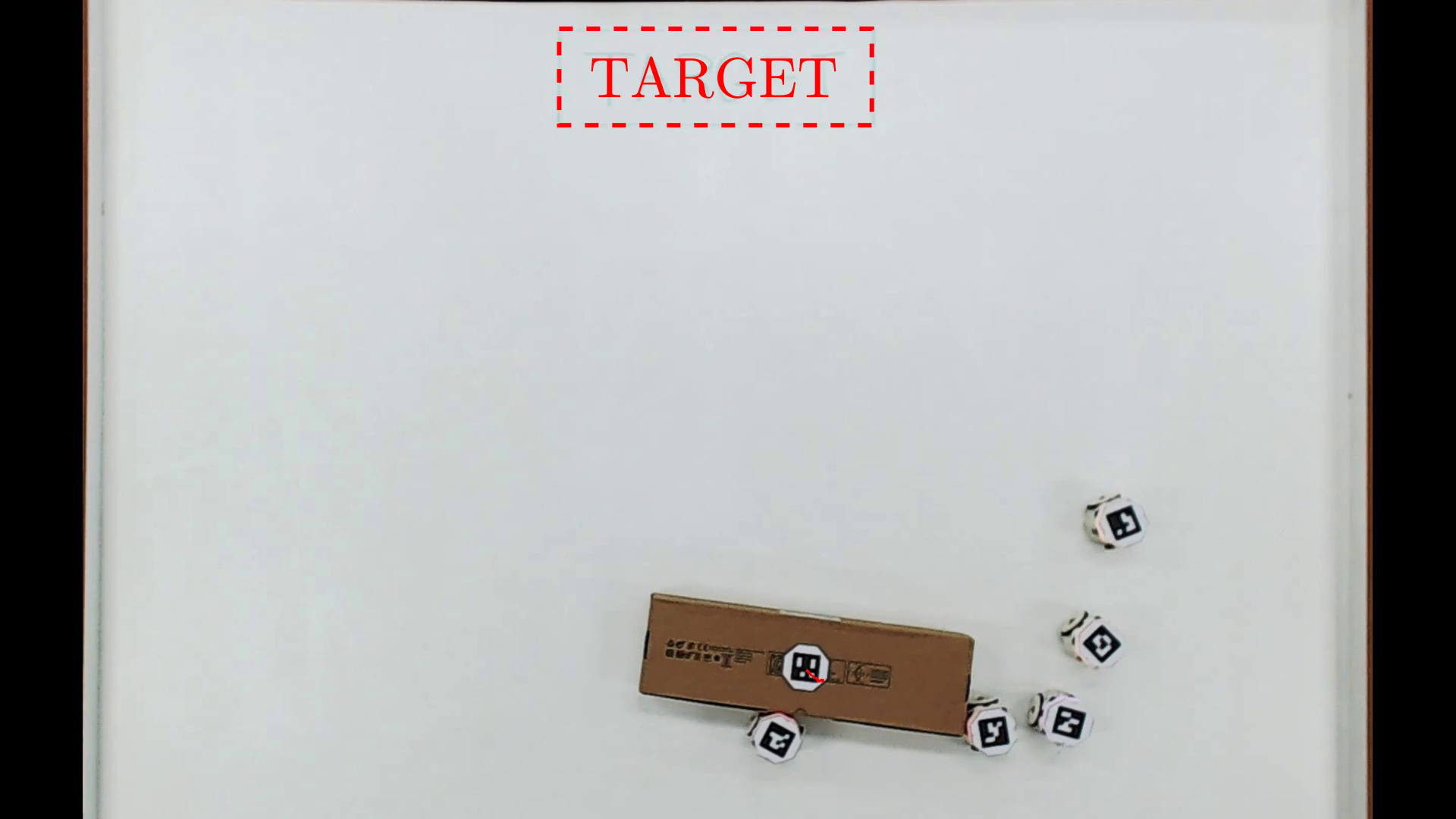}
		\caption{$t=10$~s.}
	\end{subfigure}
	\medskip
% 	\hspace*{-1.6em}
	\begin{subfigure}{.280\textwidth}
		\centering
		\includegraphics[trim={5cm 2cm 5cm 1cm},clip,width=\linewidth]{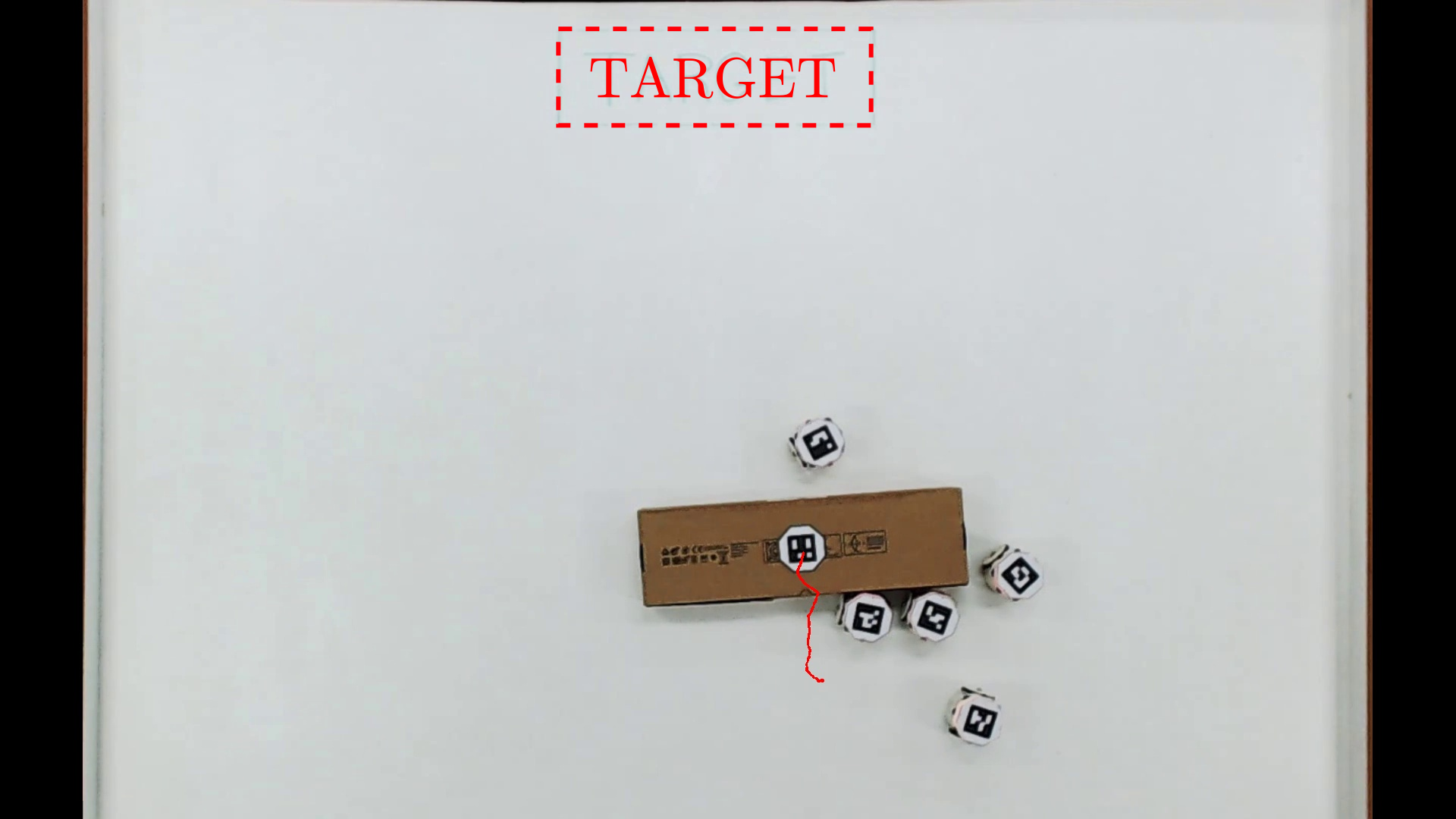}
		\caption{$t=15$~s.}
	\end{subfigure}
	\hfil
% 	\hspace*{-1.6em}
	\begin{subfigure}{.280\textwidth}
		\centering
		\includegraphics[trim={5cm 2cm 5cm 1cm},clip,width=\linewidth]{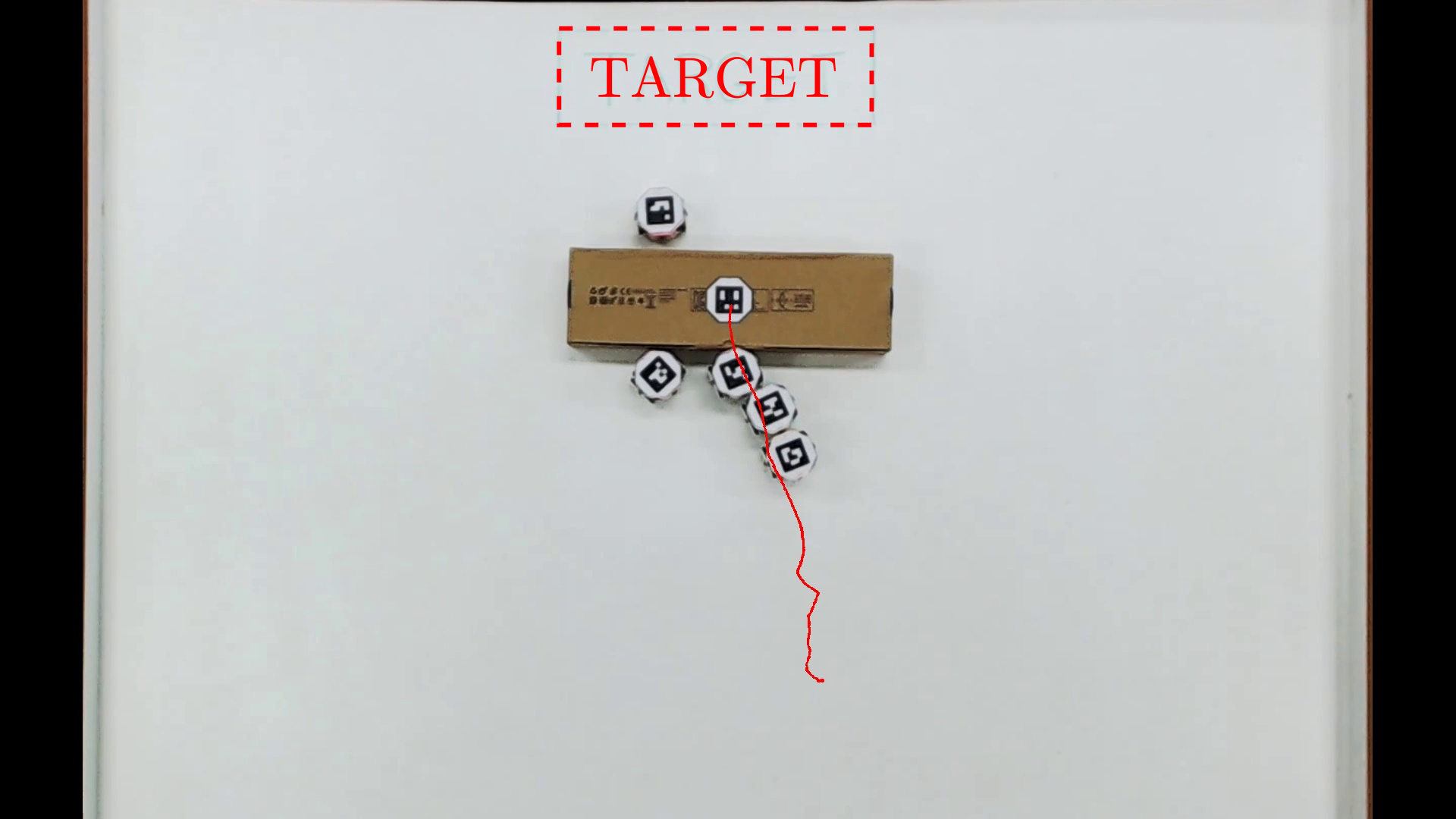}
		\caption{$t=20$~s.}
	\end{subfigure}
	\hfil
% 	\hspace*{-1.6em}
	\begin{subfigure}{.280\textwidth}
		\centering
		\includegraphics[trim={5cm 2cm 5cm 1cm},clip,width=\linewidth]{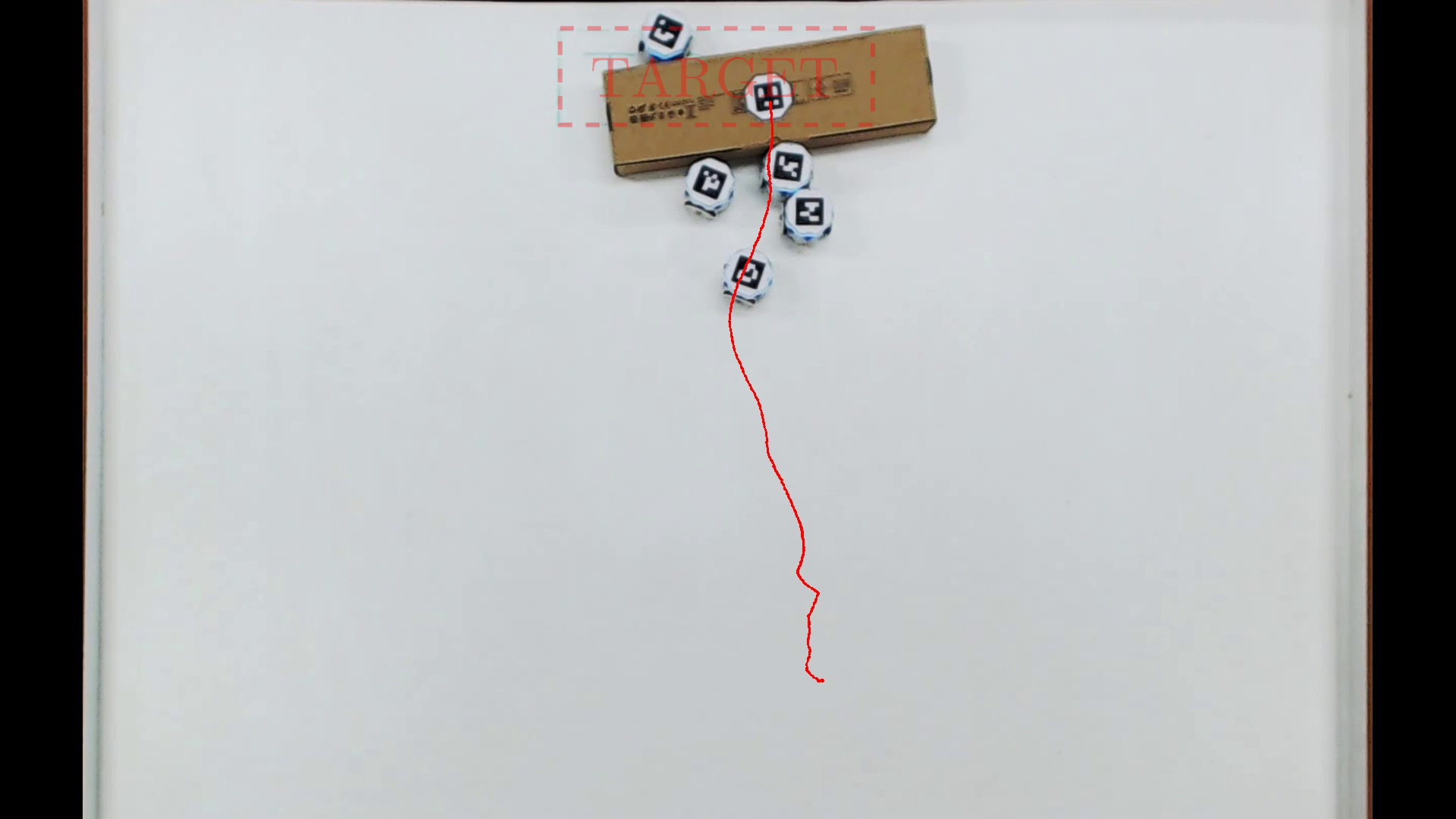}
		\caption{$t=25$~s.}
	\end{subfigure}
	\caption{Snapshots of an experiment showing five real HeRo robots transporting an object toward its goal location.}
	\label{fig:transportation}
\end{figure*}

Finally, we conducted experiments evaluating the robot's performance in a cooperative transportation task. In this task, the swarm must coordinate to push an object toward its goal location, taking advantage of multiple robots' forces applied to the object. The strategy implemented in these experiments is described in~\cite{rezeck2021cooperative} and does not require prior knowledge of the shape and location of the object, only its target. So, the robots can navigate through the environment, form groups, and when they detect the object, they can move around it looking for contact positions that allow the object to be pushed towards its objective. Despite being a decentralized strategy and not requiring global information on the swarm or the object, robots must estimate their neighbors' relative position and velocity and distinguish between object and obstacles detection. As in the previous task, we used the overhead camera location system to emulate such sensors. Fig.~\ref{fig:transportation} shows a sequence of snapshots taken by the overhead camera. In the sequence, the robots group and coordinate to transport the object to its goal. A video of this experiment is available on Youtube~\footnote{Cooperative Transport: \url{https://youtu.be/hAS7FKYkKWQ}}.

% \begin{figure*}[t]
% \centering
% 	\begin{subfigure}{.280\textwidth}
% 		\centering
% 		\includegraphics[width=\linewidth]{images/experiments/transportation/image00000001.png}
% 		\caption{$t=0$ s}
% 	\end{subfigure}%
% 	\hfil
% % 	\hspace*{-1.2em}
% 	\begin{subfigure}{.280\textwidth}
% 		\centering
% 		\includegraphics[width=\linewidth]{images/experiments/transportation/image00008458.png}
% 		\caption{$t=338$ s}
% 	\end{subfigure}
% 	\hfil
% % 	\hspace*{-1.6em}
% 	\begin{subfigure}{.280\textwidth}
% 		\centering
% 		\includegraphics[width=\linewidth]{images/experiments/transportation/image00012136.png}
% 		\caption{$t=485$ s}
% 	\end{subfigure}
% 	\medskip
% % 	\hspace*{-1.6em}
% 	\begin{subfigure}{.280\textwidth}
% 		\centering
% 		\includegraphics[width=\linewidth]{images/experiments/transportation/image00016071.png}
% 		\caption{$t=642$ s}
% 	\end{subfigure}
% 	\hfil
% % 	\hspace*{-1.6em}
% 	\begin{subfigure}{.280\textwidth}
% 		\centering
% 		\includegraphics[width=\linewidth]{images/experiments/transportation/image00020148.png}
% 		\caption{$t=805$ s}
% 	\end{subfigure}
% 	\hfil
% % 	\hspace*{-1.6em}
% 	\begin{subfigure}{.280\textwidth}
% 		\centering
% 		\includegraphics[width=\linewidth]{images/experiments/transportation/image00022986.png}
% 		\caption{$t=919$ s}
% 	\end{subfigure}
% 	\caption{Snapshots of an experiment showing $10$ simulated robots transporting an object toward a sequence of goal locations in a complex environment.}
% 	\label{fig:navigation}
% \end{figure*}

\section{Conclusion}
\label{sec:conclusion}
% 6) Conclusion and futures works
% - summup

This work presented HeRo, a novel open-source swarm robotic platform that is cost-effective, capable, and scalable, developed using off-the-shelf components and additive manufacturing. This robot was specially designed for swarm applications, given its small size, sensing, and networking capabilities. The proposed robot has WiFi communication, over-the-air firmware upgrades and is fully compatible with ROS, facilitating the development of newer functionalities. We also presented a simulation environment for the HeRo platform using Gazebo, a popular simulation engine for continuous testing and quick prototyping. 
%The platform was validated using real-world experiments  the highly popular e-Puck platform, showing similar to better odometry performance when using the external HeRo e-hat. 
Experiments evaluating the sensor accuracy, odometry, autonomy, swarm communication, and control show a performance in par or superior to other commercial or more expensive platforms. Mapping, decentralized coverage, flocking behavior and transportation tasks performed with a group of HeRo robots validate the robot's capacities for real-world swarm applications and educational use. 

\subsection{Limitations}
% - limitation
Despite the robot's remarkable performance, there are still some issues that impact its use and maintenance. Below we list and discuss some of these points.

\begin{itemize}
    \item \textbf{Reproduction/Assembly}: although the robot has a simple mechanical design, the use of additive manufacturing technologies such as conventional 3D printers does not always allow a proper fit of the parts. Then, it requires manual adjustments or finishing during assembly, demanding time and effort. This process is essential for the robot's transmission mechanisms, impacting wheel movement and encoder readings if left unattended.
    
   \item \textbf{Robot calibration}: the robot is designed to use affordable parts and components that have been available for several years. As expected, low-cost components also impact robot performance, requiring the user to calibrate the IR sensors and motors occasionally. As each component has different characteristics, the calibration process is required for each of the eight IR sensors and the two servo motors.
    
    \item \textbf{Wireless recharge}: another point is the lack of convenience in charging the battery of each robot by plugging in a cable. Recently, wireless charging modules have become commonplace, but they still come at a high cost compared to the robot's cost. Although the idea of having an automatic recharge system is interesting, due to the cost and size, we decided to wait and deal with the manual recharge of the robots.
    
    \item \textbf{Mechanical wear}: finally, another point impacted by the use of low-cost components is their durability. Although the rotary encoder is an interesting solution, some low-cost models has a short lifespan for our application, requiring replacement after months of use. In addition to the encoder, we also have to check the gear mechanism since we use ABS/PLA material that wears out with use and storage.
    
\end{itemize}

% - montagem/reproducao: etapas de construcao mecanica demanda atencao e ate mesmo ajustes na pecas impressas
% - calibrar motores e tunar pid nao e um processo escalavel
% - sem recarregamento sem fio, embora carregamento usando usb
% - desgaste a longo prazo das pecas impressas
% - ?

\subsection{Future Work}
% - future works
% ROS2 integration

We are continuously working on improving HeRo capabilities. In a near future, we intend to develop newer expansions to the platform in the form of e-Hats, improve the internal filters for localization and improve the assembly process of the platform. We will also validate newer forms of localization using UWB or other indirect wireless methods. A migration to ROS2 will also be performed to maintain the software stack of the robot up to date with current robotics advancements.

% %\begin{acknowledgements}
% %If you'd like to thank anyone, place your comments here
% %and remove the percent signs.
% %\end{acknowledgements}

% % Authors must disclose all relationships or interests that 
% % could have direct or potential influence or impart bias on 
% % the work: 


\clearpage
\bibliographystyle{spmpsci}      % mathematics and physical sciences
\bibliography{sections/bibfile.bib}   % name your BibTeX data base

\begin{thebibliography}{10}
\providecommand{\url}[1]{{#1}}
\providecommand{\urlprefix}{URL }
\expandafter\ifx\csname urlstyle\endcsname\relax
  \providecommand{\doi}[1]{DOI~\discretionary{}{}{}#1}\else
  \providecommand{\doi}{DOI~\discretionary{}{}{}\begingroup
  \urlstyle{rm}\Url}\fi

\bibitem{arvin2018mona}
Arvin, F., Espinosa, J., Bird, B., West, A., Watson, S., Lennox, B.: Mona: an
  affordable open-source mobile robot for education and research.
\newblock Journal of Intelligent \& Robotic Systems pp. 1--15 (2018)

\bibitem{arvin2011imitation}
Arvin, F., Samsudin, K., Ramli, A.R., Bekravi, M.: Imitation of honeybee
  aggregation with collective behavior of swarm robots.
\newblock International Journal of Computational Intelligence Systems
  \textbf{4}(4), 739--748 (2011)

\bibitem{arvin2009development}
Arvin, F., Samsudin, K., Ramli, A.R., et~al.: Development of a miniature robot
  for swarm robotic application.
\newblock International Journal of Computer and Electrical Engineering
  \textbf{1}(4), 436--442 (2009)

\bibitem{arvin2014comparison}
Arvin, F., Turgut, A.E., Bellotto, N., Yue, S.: Comparison of different
  cue-based swarm aggregation strategies.
\newblock In: International Conference in Swarm Intelligence, pp. 1--8.
  Springer (2014)

\bibitem{arvin2018perpetual}
Arvin, F., Watson, S., Turgut, A.E., Espinosa, J., Krajn{\'\i}k, T., Lennox,
  B.: Perpetual robot swarm: long-term autonomy of mobile robots using
  on-the-fly inductive charging.
\newblock Journal of Intelligent \& Robotic Systems \textbf{92}(3-4), 395--412
  (2018)

\bibitem{benet2002using}
Benet, G., Blanes, F., Sim{\'o}, J.E., P{\'e}rez, P.: Using infrared sensors
  for distance measurement in mobile robots.
\newblock Robotics and autonomous systems \textbf{40}(4), 255--266 (2002)

\bibitem{caprari2003design}
Caprari, G., Siegwart, R.: Design and control of the mobile micro robot alice.
\newblock In: Proceedings of the 2nd International Symposium on Autonomous
  Minirobots for Research and Edutainment, AMiRE 2003: 18-20 February 2003,
  Brisbane, Australia, pp. 23--32. CITI (2003)

\bibitem{eshaghi2020mroberto}
Eshaghi, K., Li, Y., Kashino, Z., Nejat, G., Benhabib, B.: mroberto 2.0--an
  autonomous millirobot with enhanced locomotion for swarm robotics.
\newblock IEEE Robotics and Automation Letters \textbf{5}(2), 962--969 (2020)

\bibitem{farrow2014miniature}
Farrow, N., Klingner, J., Reishus, D., Correll, N.: Miniature six-channel range
  and bearing system: algorithm, analysis and experimental validation.
\newblock In: 2014 IEEE International Conference on Robotics and Automation
  (ICRA), pp. 6180--6185. IEEE (2014)

\bibitem{hostettler2016real}
Hostettler, L., {\"O}zg{\"u}r, A., Lemaignan, S., Dillenbourg, P., Mondada, F.:
  Real-time high-accuracy 2d localization with structured patterns.
\newblock In: Robotics and Automation (ICRA), 2016 IEEE International
  Conference on, pp. 4536--4543. IEEE (2016)

\bibitem{hu2018colias}
Hu, C., Fu, Q., Yue, S.: Colias iv: The affordable micro robot platform with
  bio-inspired vision.
\newblock In: Annual Conference Towards Autonomous Robotic Systems, pp.
  197--208. Springer (2018)

\bibitem{kernbach2011swarmrobot}
Kernbach, S.: Swarmrobot. org-open-hardware microrobotic project for
  large-scale artificial swarms.
\newblock arXiv preprint arXiv:1110.5762  (2011)

\bibitem{kim2016mroberto}
Kim, J.Y., Colaco, T., Kashino, Z., Nejat, G., Benhabib, B.: mroberto: A
  modular millirobot for swarm-behavior studies.
\newblock In: 2016 IEEE/RSJ International Conference on Intelligent Robots and
  Systems (IROS), pp. 2109--2114. IEEE (2016)

\bibitem{klingner2014stick}
Klingner, J., Kanakia, A., Farrow, N., Reishus, D., Correll, N.: A stick-slip
  omnidirectional powertrain for low-cost swarm robotics: Mechanism,
  calibration, and control.
\newblock In: Intelligent Robots and Systems (IROS 2014), 2014 IEEE/RSJ
  International Conference on, pp. 846--851. IEEE (2014)

\bibitem{koenig2004design}
Koenig, N.P., Howard, A.: Design and use paradigms for gazebo, an open-source
  multi-robot simulator.
\newblock In: IROS, vol.~4, pp. 2149--2154. Citeseer (2004)

\bibitem{le2016zooids}
Le~Goc, M., Kim, L.H., Parsaei, A., Fekete, J.D., Dragicevic, P., Follmer, S.:
  Zooids: Building blocks for swarm user interfaces.
\newblock In: Proceedings of the 29th Annual Symposium on User Interface
  Software and Technology, pp. 97--109. ACM (2016)

\bibitem{limeira2019wsbot}
Limeira, M.A., Piardi, L., Kalempa, V.C., de~Oliveira, A.S., Leit{\~a}o, P.:
  Wsbot: A tiny, low-cost swarm robot for experimentation on industry 4.0.
\newblock In: 2019 Latin American Robotics Symposium (LARS), 2019 Brazilian
  Symposium on Robotics (SBR) and 2019 Workshop on Robotics in Education (WRE),
  pp. 293--298. IEEE (2019)

\bibitem{malyuta2018guidance}
Malyuta, D.: Guidance, navigation, control and mission logic for quadrotor
  full-cycle autonomy.
\newblock Master's thesis, ETH Zurich (2018)

\bibitem{mondada2009puck}
Mondada, F., Bonani, M., Raemy, X., Pugh, J., Cianci, C., Klaptocz, A.,
  Magnenat, S., Zufferey, J.C., Floreano, D., Martinoli, A.: The e-puck, a
  robot designed for education in engineering.
\newblock In: Proceedings of the 9th conference on autonomous robot systems and
  competitions, vol.~1, pp. 59--65. IPCB: Instituto Polit{\'e}cnico de Castelo
  Branco (2009)

\bibitem{mondada1999development}
Mondada, F., Franzi, E., Guignard, A.: The development of khepera.
\newblock In: Experiments with the Mini-Robot Khepera, Proceedings of the First
  International Khepera Workshop, CONF, pp. 7--14 (1999)

\bibitem{mondada1994mobile}
Mondada, F., Franzi, E., Ienne, P.: Mobile robot miniaturisation: A tool for
  investigation in control algorithms.
\newblock In: Experimental robotics III, pp. 501--513. Springer (1994)

\bibitem{olaronke2020systematic}
Olaronke, I., Rhoda, I., Gambo, I., Oluwaseun, O., Janet, O.: A systematic
  review of swarm robots.
\newblock Curr. J. Appl. Sci. Technol. \textbf{39}, 79--97 (2020)

\bibitem{ozgur2017cellulo}
{\"O}zg{\"u}r, A., Lemaignan, S., Johal, W., Beltran, M., Briod, M., Pereyre,
  L., Mondada, F., Dillenbourg, P.: Cellulo: Versatile handheld robots for
  education.
\newblock In: Proceedings of the 2017 ACM/IEEE International Conference on
  Human-Robot Interaction, pp. 119--127. ACM (2017)

\bibitem{pickem2017robotarium}
Pickem, D., Glotfelter, P., Wang, L., Mote, M., Ames, A., Feron, E., Egerstedt,
  M.: The robotarium: A remotely accessible swarm robotics research testbed.
\newblock In: Robotics and Automation (ICRA), 2017 IEEE International
  Conference on, pp. 1699--1706. IEEE (2017)

\bibitem{pickem2015gritsbot}
Pickem, D., Lee, M., Egerstedt, M.: The gritsbot in its natural habitat-a
  multi-robot testbed.
\newblock In: Robotics and Automation (ICRA), 2015 IEEE International
  Conference on, pp. 4062--4067. IEEE (2015)

\bibitem{quigley2009ros}
Quigley, M., Conley, K., Gerkey, B., Faust, J., Foote, T., Leibs, J., Wheeler,
  R., Ng, A.Y., et~al.: Ros: an open-source robot operating system.
\newblock In: ICRA workshop on open source software, vol.~3, p.~5. Kobe, Japan
  (2009)

\bibitem{rezeck2021cooperative}
Rezeck, P., Assunção, R.M., Chaimowicz, L.: Cooperative object transportation
  using gibbs random fields.
\newblock In: 2021 IEEE/RSJ International Conference on Intelligent Robots and
  Systems (IROS), pp. 9131--9138 (2021).
\newblock \doi{10.1109/IROS51168.2021.9635928}

\bibitem{rezeck2021flocking}
Rezeck, P., Assunção, R.M., Chaimowicz, L.: Flocking-segregative swarming
  behaviors using gibbs random fields.
\newblock In: 2021 IEEE International Conference on Robotics and Automation
  (ICRA), pp. 8757--8763 (2021).
\newblock \doi{10.1109/ICRA48506.2021.9561412}

\bibitem{rezeck2017hero}
Rezeck, P.A.F., Azpurua, H., Chaimowicz, L.: Hero: An open platform for
  robotics research and education.
\newblock In: 2017 Latin American Robotics Symposium (LARS) and 2017 Brazilian
  Symposium on Robotics (SBR), pp. 1--6 (2017).
\newblock \doi{10.1109/SBR-LARS-R.2017.8215317}

\bibitem{rubenstein2014kilobot}
Rubenstein, M., Ahler, C., Hoff, N., Cabrera, A., Nagpal, R.: Kilobot: A low
  cost robot with scalable operations designed for collective behaviors.
\newblock Robotics and Autonomous Systems \textbf{62}(7), 966--975 (2014)

\bibitem{slavkov2018morphogenesis}
Slavkov, I., Carrillo-Zapata, D., Carranza, N., Diego, X., Jansson, F.,
  Kaandorp, J., Hauert, S., Sharpe, J.: Morphogenesis in robot swarms.
\newblock Science Robotics \textbf{3}(25) (2018)

\bibitem{soares2016khepera}
Soares, J.M., Navarro, I., Martinoli, A.: The khepera iv mobile robot:
  performance evaluation, sensory data and software toolbox.
\newblock In: Robot 2015: second Iberian robotics conference, pp. 767--781.
  Springer (2016)

\bibitem{wang2016apriltag}
Wang, J., Olson, E.: Apriltag 2: Efficient and robust fiducial detection.
\newblock In: 2016 IEEE/RSJ International Conference on Intelligent Robots and
  Systems (IROS), pp. 4193--4198. IEEE (2016)

\bibitem{west2018ros}
West, A., Arvin, F., Martin, H., Watson, S., Lennox, B.: Ros integration for
  miniature mobile robots.
\newblock In: Annual Conference Towards Autonomous Robotic Systems, pp.
  345--356. Springer (2018)

\bibitem{wilson2020robotarium}
Wilson, S., Glotfelter, P., Wang, L., Mayya, S., Notomista, G., Mote, M.,
  Egerstedt, M.: The robotarium: Globally impactful opportunities, challenges,
  and lessons learned in remote-access, distributed control of multirobot
  systems.
\newblock IEEE Control Systems Magazine \textbf{40}(1), 26--44 (2020).
\newblock \doi{10.1109/MCS.2019.2949973}

\bibitem{yu2017portable}
Yu, J., Han, S.D., Tang, W.N., Rus, D.: A portable, 3d-printing enabled
  multi-vehicle platform for robotics research and education.
\newblock In: Robotics and Automation (ICRA), 2017 IEEE International
  Conference on, pp. 1475--1480. IEEE (2017)

\end{thebibliography}

% \begin{thebibliography}{}
% %
% % and use \bibitem to create references. Consult the Instructions
% % for authors for reference list style.
% % Format for Journal Reference
% Author, Article title, Journal, Volume, page numbers (year)
% Author, Book title, page numbers. Publisher, place (year)
% \end{thebibliography}

\end{document}